\documentclass[acmsmall]{acmart}

\AtBeginDocument{%
  \providecommand\BibTeX{{%
    \normalfont B\kern-0.5em{\scshape i\kern-0.25em b}\kern-0.8em\TeX}}}

\copyrightyear{2021} 
\acmYear{2021} 
\setcopyright{acmlicensed}\acmConference[CHI '21]{CHI Conference on Human Factors in Computing Systems}{May 8--13, 2021}{Yokohama, Japan}
\acmBooktitle{CHI Conference on Human Factors in Computing Systems (CHI '21), May 8--13, 2021, Yokohama, Japan}
\acmPrice{15.00}
\acmDOI{10.1145/3411764.3445315}
\acmISBN{978-1-4503-8096-6/21/05}

\acmConference[CHI '21]{CHI 2021: ACM CHI Conference on Human Factors in Computing Systems }{May 8-13, 2021}{Yokohama, Japan}

\usepackage[title]{appendix}
\usepackage{subcaption} 
\usepackage{framed} 
\usepackage{enumitem} 
\usepackage{pdfpages}
\usepackage{pbox} 
\usepackage{graphics}
\usepackage{longtable}
\usepackage[export]{adjustbox} 
\usepackage{soul} 

\soulregister\citep7 
\soulregister\abr7
\soulregister\ref7

\newcommand{\ignore}[1]{}

\newcommand{\abr}[1]{\textsc{#1}}
\newcommand{\NA}{---}
\newcommand{\edit}[1]{#1}
\newcommand{\medit}[1]{#1}
\renewcommand{\comment}[1]{}

\def\imagebox#1#2{\vtop to #1{\null\hbox{#2}\vfill}}

\begin{document}

\title{Manipulating and Measuring Model Interpretability}

\author{Forough Poursabzi-Sangdeh}
\email{fpoursabzi@microsoft.com}
\affiliation{%
 \institution{Microsoft Research}
 \streetaddress{300 Lafayette Street}
 \city{New York}
 \state{NY}
 \postcode{10012}
}
\author{Daniel G. Goldstein}
\email{dgg@microsoft.com}
\orcid{0000-0002-0970-5598}
\affiliation{%
 \institution{Microsoft Research}
 \streetaddress{300 Lafayette Street}
 \city{New York}
 \state{NY}
 \postcode{10012}
}
\author{Jake M. Hofman}
\email{jmh@microsoft.com}
\affiliation{%
 \institution{Microsoft Research}
 \streetaddress{300 Lafayette Street}
 \city{New York}
 \state{NY}
 \postcode{10012}
}
\author{Jennifer Wortman Vaughan}
\email{jenn@microsoft.com}
\affiliation{%
 \institution{Microsoft Research}
 \streetaddress{300 Lafayette Street}
 \city{New York}
 \state{NY}
 \postcode{10012}
}
\author{Hanna Wallach}
\email{wallach@microsoft.com}
\affiliation{%
 \institution{Microsoft Research}
 \streetaddress{300 Lafayette Street}
 \city{New York}
 \state{NY}
 \postcode{10012}
}

\renewcommand{\shortauthors}{Poursabzi-Sangdeh, et al.}

\begin{abstract}
  \medit{With machine learning models being increasingly used to aid
    decision making even in high-stakes domains, there has been a
    growing interest in developing interpretable models. Although many
    supposedly interpretable models have been proposed, there have
    been relatively few experimental studies investigating whether
    these models achieve their intended effects, such as making people
    more closely follow a model's predictions when it is beneficial
    for them to do so or enabling them to detect when a model has made
    a mistake. We present a sequence of pre-registered experiments
    ($N=3,800$) in which we showed participants functionally identical
    models that varied only in two factors commonly thought to make
    machine learning models more or less interpretable: the number of
    features and the transparency of the model (i.e., whether the model internals are clear or black
    box). Predictably, participants who saw a clear model with few
    features could better simulate the model's predictions. However,
    we did not find that participants more closely followed its predictions. Furthermore, showing participants a clear model meant that
    they were \emph{less} able to detect and correct for the model's
    sizable mistakes, seemingly due to information overload. These
    counterintuitive findings emphasize the importance of testing over
    intuition when developing interpretable models.}
\end{abstract}




\maketitle

\section{Introduction}
\label{sec:introduction}
Machine learning models are \medit{increasingly used to aid decision making in high-stakes domains, such as} medical diagnosis~\citep{kononenko2001}, credit risk assessment~\citep{hand1997}, judicial sentencing and bail~\citep{kleinberg2018human,angwin2016,chouldechova2017,jung2020simple}, and hiring~\citep{liem_2018}. \medit{Machine learning models also} influence \medit{people's} decisions about what news \medit{articles to read}~\citep{bucher2017,alvarado2018,rader2015,liu2018,vaccaro2018illusion}, what movies to watch~\citep{bennett_netflix}\medit{,} what music to listen to~\citep{mehrotra_2018}, \medit{what clothes to buy}~\citep{colson_stitchfix}, and \medit{even} who to date~\citep{rudder_dataclysm}. \medit{In all of these settings, decision making is a collaboration between people and models, where models make predictions and people can choose whether to follow these predictions or to override them.}

\edit{There are many reasons why following a machine learning model's predictions may be advantageous, chief among them being improved accuracy. Indeed, there have been many studies showing that models are often more accurate than people.} A meta-analysis from \medit{twenty} years ago, reviewing work from some \medit{seventy} years \medit{ago}, found that 
models were more accurate than people in a variety of domains~\citep{dawes1989clinical,grove2000clinical}, and the gap has widened since then as models have \medit{become more accurate}~\citep{kleinberg2018human}. \edit{Although following a model's predictions should enable people to make faster and more consistent decisions, there are also scenarios in which following a model's predictions can be disadvantageous, most notably when those predictions are incorrect.}

\edit{That said, people are resistant to using models to aid their decision making~\citep{dawes1979,bazerman1985,promberger2006patients,DSM15}.
There are many reasons for this: First, people may feel that they do not understand models, including what information they rely on and how this information is being used. For instance, in a study of machine learning use in the public sector~\citep{veale2018fairness}, several practitioners noted challenges in getting organizational buy-in for the use of machine learning-based systems without the ability to explain those systems' internals.}
\edit{Second, people may feel that models do not rely on the right
information}~\cite[p.~151]{dawes1989clinical} \edit{or that they do not use
information in the right ways~\citep{veale2018fairness}. Third, people may be
worried that models might behave in ways that are
unfair~\citep{veale2018fairness,kleinberg2019simplicity,rudin2019stop,
fastcompany_hiring2018,time_hiring2017,noble2018,amazon_recruiting_2018}. Concerns about fairness are often exacerbated by the first two
reasons.}

\edit{In response to these concerns,} a prolific line of research has emerged
that focuses on \edit{the \emph{interpretability} of machine learning
models. There are two main approaches to developing supposedly
interpretable models. First, because there is evidence that simple
models with clear internals} can be as accurate as more complex,
black-box models in some
domains~\citep{rudin2018,dawes1979,schoemaker1982,gigerenzer1996reasoning,jung2020simple,aastebro2006},
one approach
is to create simple\edit{, clear} models such as point systems that
can be memorized~\citep{jung2020simple,UR16} or \edit{generalized} additive models \edit{that facilitate visualizing} the impact of each feature on the model's prediction\edit{s}~\citep{lou2012intelligible,lou2013ga2m,caruana2015intelligible}. \edit{The hope is that these models will be easier for people to understand and use.}

The second approach is to provide post-hoc explanations for
potentially complex\edit{, black-box} models. Threads of research \edit{that focus on this approach} look at how to explain individual predictions by learning simple local
approximations of a model around particular data
points~\citep{ribeiro2016should,LL17,LKCL17}, training simple models
to mimic \edit{more} complex ones~\citep{lakkaraju2019faithful,tan2018distill},
estimating the influence of training \edit{data points}~\citep{KL17}, describing
the change to an input data point that would change a model's
prediction for it~\citep{ustun2019actionable, russell2019efficient,
wachter2017counterfactual, weld2019challenge}, and visualizing model
\edit{predictions} or properties~\citep{WVH16,KD+17}.

But despite this progress, there is still no
consensus about how to define, quantify, or measure the
interpretability of a machine learning model~\citep{doshi2017towards}, raising the following question: What is interpretability and how can \edit{we determine}
\edit{whether} one model is more interpretable than another? Different notions of interpretability, such as \edit{simplicity,
transparency,} simulatability, \edit{and} trustworthiness, are often
conflated~\citep{lipton2016mythos}. This problem is exacerbated by the
fact that \edit{machine learning models have many different types of stakeholders} and these \edit{stakeholders} may have different needs in different
scenarios~\citep{TBH+18,tintarev2015explaining, hohman2019gamut}. The
approach that works best for a regulator who wants to understand why a
particular person was denied a loan may be different to the approach
that works best for a data scientist debugging a machine learning
model or for a CEO using a model to make a high-stakes decision. \edit{Moreover,
regardless of how interpretability is defined, quantified, or
measured, there is very little scientific evidence demonstrating that
a) people are better able to understand interpretable models, b)
people more closely follow the predictions of interpretable
models when it is beneficial for them to do so, and c) people are
better able to detect when an interpretable model has made a
mistake, enabling them to override its prediction.}

We take the perspective that the \edit{lack of consensus around defining,
quantifying, or measuring interpretability, as well as the lack of
scientific evidence for its benefits,} stem from the fact that
interpretability is not something that can be directly manipulated or
measured.  Rather, the interpretability of a model is a latent\edit{---and fundamentally
human---}property that can be influenced by different \emph{manipulable
factors} (such as the number of features, the complexity of the model,
the transparency of the model, or even the user interface) and that
impacts different \emph{measurable \edit{(human)} outcomes} (such as people's \edit{abilities to simulate the model's predictions, the extent to which people} follow the model's predictions \edit{when it is beneficial for them to do so, }or people's abilit\edit{ies} to \edit{detect when} the model\edit{ has made a mistake}). Different factors may
influence \edit{different} outcomes in different ways. As such, we argue that to
understand interpretability, it is necessary to directly manipulate different factors \edit{and measure their effects}. What is or is not interpretable must
be defined by \edit{people's} behavior, not by what appeals to intuition~\citep{miller_2017,
miller2019explanation,wang2019designing,liao2020questioning}.

\edit{Draw}ing on \edit{this perspective, we} present a sequence of pre-registered
experiments ($N=3,800$) in which we varied factors \edit{commonly} thought to
make \edit{machine learning} models more interpretable and measured
the\edit{ir effects on} people's \edit{behavior.} Based on a structured review of the
literature on \edit{interpretability}, we focused on two factors---the number
of features and the transparency \edit{of the model} (i.e., whether the model
internals are clear or black box)---and investigated how these
factors affected three \edit{measurable} outcomes:
\begin{enumerate}
  \item {\bf \edit{How well can people simulate a model's predictions?}}
  \item {\bf \edit{To what extent do people follow a model's predictions when it is beneficial for them to do so?}}
  \item {\bf \edit{How well can people detect when a model has made a mistake and correct for it?}}
\end{enumerate}
\edit{We found} that people can better simulate \edit{the predictions of} a clear model with few features compared to \edit {the predictions of} \edit{a clear model with more features or the predictions of a black-box model. However, contrary to our expectations, we did} not find a significant improvement in the \edit{extent} to which people follow the predictions of a clear model with few features when it is beneficial for them to do so compared to the predictions of a black-box model with more features. \edit{We also found that using a clear model hampers people's abilities to} detect when the model had made a mistake. \edit{All three of these findings are based on highly-powered, pre-registered experiments with multiple, representative stimuli. 
Our latter two findings are notable and surprising because they contradict common intuition about interpretability.}


\subsection{Domain}

\edit{We focused} our experiments \edit{on the domain of} real-estate valuation\edit{,} in which \edit{machine learning models are used to} predict \edit{the} selling prices of properties.\footnote{The Zestimate prices on the website Zillow may be a familiar example of these predictions.} In each experiment, participants were asked to predict the prices of apartments in a single neighborhood in New York City with the help of a machine learning model.
We conducted our experiments on laypeople, as \edit{laypeople represent one type of stakeholder that might potentially use or be affected} by machine learning models. \edit{We chose the domain of real-estate valuation} because many people have considered purchasing a \edit{property}, making the setting both familiar and potentially interesting to participants.
Each apartment was represented in terms of eight features: number of bedrooms, number of bathrooms, square footage, total rooms, days on the market, maintenance fee, distance from the subway, and distance from a school. All participants saw the same set of apartments (i.e., the same feature values) and, crucially, the same predict\edit{ed selling price} for each apartment, which came from \edit{either a two-feature or an eight-feature} linear regression model. \edit{To achieve this, the models were constrained to make the same predictions for the apartments we used, as we describe below in Section}~\ref{subsec:exp1_design}. What varied between the experimental conditions was therefore \emph{only the presentation of the model}\edit{: whether it was presented as using
two- or eight-features and whether the model internals were clear or black box.} As a result, any observed differences in participants' behavior \edit{could} be attributed entirely to the presentation\edit{ of the model---}a key feature of our experimental design.

\edit{Because of our decision to vary only the presentation of the model,
each participant had access to all eight feature values for each
apartment, regardless of the experimental condition to which they were
assigned. This meant that some participants (those who were shown an eight-feature model) had access to the same information as the
model, while others (those who were shown a two-feature
model) had access to more information than the model.
This scenario has a rich history in the decision-making literature, where it
is called the ``broken leg problem''}~\cite[p.~151]{dawes1989clinical}\edit{,
based on an anecdote in which a model that is very good at predicting
weekly attendance at the movies should be ignored if it is known that
someone has a broken femur with a full length
cast.} \edit{This scenario is also often encountered in practice:} Table~\ref{tab:decision_aid_table} in
appendix~\ref{appndx:decision_aid_table} contains a number of \edit{instances
from the literature in which people have access to more information
than a model, meaning that they can use their knowledge of this
additional information as a reason to override the model's
predictions.}

\subsection{Overview of experiments}

In our first experiment,
we \edit{showed participants} a sequence of twelve apartments.
The first ten apartments had typical configurations \edit{(i.e., typical combinations of feature values),} whereas the last two had unusual configurations (\edit{such as three bathrooms squeezed into 726 square feet}).
For each apartment, participants \edit{were} first \edit{shown} its configuration (i.e., feature \edit{values}) alongside the model \edit{(whose internals were either clear or black box)} and were asked to \edit{guess} what the model would predict for the apartment's selling price.
They were then shown the model's prediction and asked for their own \edit{prediction} of the apartment\edit{'s selling price.}

We hypothesized that participants who were shown \edit{the clear, two-}feature \edit{model} would better simulate the model's predictions~\citep{lage2019human} and \edit{would} more \edit{closely} follow \edit{its} predictions \edit{when it was beneficial for them to do so}.
We also hypothesized that \edit{participants} assigned \edit{to different experimental} conditions would \edit{be differently able to detect and} correct \edit{for} the model's \edit{sizable mistakes on} the \edit{apartments with unusual configurations}. \edit{We note that here and throughout the rest of paper, when we refer to detecting and correcting a model's mistakes, we are specifically referring to whether participants notice that the model has made an inaccurate prediction and provide a more accurate prediction themselves; doing so does not necessarily imply that they understand why the model made the mistake.}

\edit{As expected, we found that} participants who \edit{saw the} clear model with \edit{two} features \edit{could} better simulate the model's predictions. However, we did not find that \edit{participants more closely followed its predictions when it was beneficial for them to do so. Moreover, participants' predictions were generally} less accurate than the models'\edit{---}a familiar finding in the literature on \edit{the predictions of people versus computational systems}~\citep{grove2000clinical,dawes1989clinical,gummadi2019}.
Furthermore, \edit{and contrary to our intuition when designing the experiment}, participants who were shown a clear model were \textit{less} \edit{able} to \edit{detect and} correct \edit{for} the model's \edit{sizable mistakes on} the apartments \edit{with unusual configurations} \edit{compared to participants who were shown a black-box model.} To account for these unexpected \edit{findings, we designed and} ran three additional experiments.

In our second experiment, we
scaled down the apartment\edit{s' selling} prices and maintenance fees to match median \edit{prices} in the U.S. in order to determine whether the findings from our first experiment were merely an artifact of New York City's high prices. \edit{Reassuringly,} with scaled-down \edit{selling prices and maintenance fees}, the findings from our first experiment replicated quite closely.

Our third experiment \edit{used} \emph{weight of advice}---\edit{a measure} commonly used in the literature on advice-taking~\citep{Y04,GM07} and subsequently used in the context of computational \edit{decision making} by Logg~\citep{L17,logg2019algorithm}---\edit{as an alternative way to measure the extent to which people follow a model's predictions. Here too, we found no significant differences in the extent to which participants followed the predictions of the model when it was beneficial for them to do so between the experimental conditions.
Surprisingly, and contrary to our findings from the previous two experiments, we did \emph{not} find that participants who were shown a clear model were less able to detect and correct for the model's sizable mistakes.}


\edit{We conjectured two possible reasons for this finding. First, in all three experiments, participants may have anchored on the prediction visible to them when making their own final prediction of an apartment's selling price. However, the possible anchor values in the first two experiments were different to that in the third: in the third experiment, participants saw their own initial prediction of each apartment's selling price when making their final prediction, whereas in the first two experiments, participants saw their simulation of the model's prediction. In the first two experiments, participants who were shown the clear, two-feature model could better simulate the model's predictions compared to participants assigned to the other experimental conditions, and might therefore have anchored on higher selling prices for the apartments with unusual configurations, in line with the model's predictions. Additionally, participants who were shown a clear model may have been overwhelmed by the amount of detail in front of them, causing them to be less likely to notice the unusual apartment configurations when making their own predictions. This effect may have been less prominent in our third experiment because participants made their initial predictions before being shown the model.}


\edit{This motivated our fourth and final experiment. For this experiment, we returned to the design of our first experiment, but removed the simulation step and varied whether or not participants were shown an  ``outlier focus'' message highlighting the apartments with unusual configurations as possible outliers. We found that participants who were shown a clear model and no outlier focus message were less able to detect and correct for the model's sizable mistakes, as in our first two experiments. In contrast, this difference disappeared for participants who were shown an outlier focus message, in line with the findings from our third experiment. The findings from our fourth experiment are therefore consistent with the possible explanation outlined above.}

\edit{In light of this, we then conducted some additional post-hoc analyses of the data from our first two experiments, finding that participants who were shown a clear, eight feature model (i.e., the model presentation with the most information) were worse at simulating the model's predictions~\citep{lage2019human} and followed its predictions less closely compared to participants assigned to the other experimental conditions. We also found that these participants' predictions of the apartments' selling prices were less accurate. These findings, which we present along with our findings from the fourth experiment, are also consistent with the explanation above.}

\edit{To summarize, via a sequence of pre-registered} experiments involving several thousand participants, we \edit{found} that \edit{two} factors \edit{commonly} thought to \edit{make machine learning models more interpretable often have} negligible effects \edit{on people's behavior and, in some cases,} even have detrimental effects. \edit{Contrary to the intuition that models with clear internals can only improve people's} decisions, our \edit{findings} suggest otherwise. Taken together, \edit{our findings} emphasize the importance of testing over intuition \edit{when developing} interpretable models.

\edit{In the next section, we further situate our experiments in the literature from the machine learning, human--computer interaction, and decision-making communities.}  In the \edit{subsequent four sections, we describe our} experiments and present \edit{our findings in detail. We then} conclude by discussing limitations of and possible extensions to our work, as well as implications for design\edit{ing user} interfaces \edit{that facilitate effective collaborations between people and models.}

\section{Related Work}
\label{sec:related}
\edit{Although there has been a recent surge of research in the machine learning community on techniques for achieving interpretability~\citep{ribeiro2016should,LL17,LKCL17,jung2020simple,UR16,caruana2015intelligible,lakkaraju2019faithful,tan2018distill,KL17,ustun2019actionable,russell2019efficient, wachter2017counterfactual,isaac_2018}, there have been relatively fewer studies of how factors commonly thought to make machine learning models more interpretable affect people's behavior.
Perhaps most closely related and contemporaneous to our work, Lage et al.~\citep{lage2019human} used controlled experiments involving laypeople to investigate how the complexity of a model affects its simulatability, focusing on decision sets. They found that the number of cognitive chunks and the model size both affect people's abilities to simulate a model's predictions. Other researchers have conducted user studies in order to understand people's use of specific tools or methods.
For example, Huysmans et al.~\citep{huysmans2011empirical} studied the effects of presenting people with models that are traditionally thought to be more interpretable (such as decision tables and binary decision trees) on people's accuracies and their stated confidences in completing a task;
Lim et al.~\citep{lim2009and} studied the effects of different types of explanations (such as probing a machine learning model about why it made a particular prediction or why it did not make a different prediction) on laypeople's understandings of and trust in a model; Rader et al.~\citep{rader2018explanations} studied the effects of different ways of explaining Facebook's News Feed algorithm on people's understandings of how the algorithm works and their ability to evaluate the correctness of the algorithm's output; Cheng et al.~\citep{cheng19explaining} studied the effects of different design and interface choices for presenting explanations on people's understandings of and trust in computational decisions; Binns et al.~\citep{binns2018s} and Dodge et al.~\citep{dodge2019explaining} studied the effects of different types of explanations on people's perceptions of a model's fairness; and
Kaur et al.~\citep{kaur2020interpreting} studied data scientists' use of two specific interpretability tools (the InterpretML~\citep{nori2019interpretml} implementation of generalized additive models and the SHAP Python package), finding that data scientists over-trust and misuse these tools.
More commonly, machine learning researchers include small-scale user studies to evaluate their own proposed techniques. For example, Lakkaraju et al.
~\citep{lakkaraju16interpretable} ran a user study comparing 47 students' understandings of decision boundaries corresponding to interpretable decision sets versus Bayesian decision lists, while Ribeiro et al.~\citep{ribeiro2016should} ran experiments to investigate whether laypeople are able to use local interpretable model-agnostic explanations to choose which of two classifiers is better, to perform feature engineering, and to identify classifier irregularities.}

\edit{Within the human--computer interaction community, there is a longstanding practice of taking a user-centered perspective and acknowledging that people are active participants who form their own mental models of how computational systems work~\citep{norman87human-computer,J83,GS83}.  Bellotti and Edwards~\citep{BE01} argued for design principles that support intelligibility, so that systems ``represent to their users what they know, how they know it, and what they are doing about it.'' Similarly, Glass et al.~\citep{glass2008toward} demonstrated empirically that transparency around how complex adaptive agents work improves people's trust in those agents.
Stumpf et al.~\citep{stumpf2009interacting} were among the first researchers to address the role of mental models when people interact with machine learning-based systems. They conducted a series of experiments to study the benefits of allowing rich interactions between people and systems, assessing whether different types of explanations would better enable people to form useful mental models.
They and others~\citep{kocielnik2019will}
also found that people become more willing to use computational
systems when they are given the opportunity to review and potentially modify the systems, even when the modifications have no
effects~\citep{vaccaro2018illusion}.
Kulesza et al.~\citep{KS+13} studied several ways in which intelligent agents might explain themselves to stakeholders. They showed that completeness of explanations is more important than soundness in accurately shaping mental models, but that people lose trust in a system when soundness is too low.
}

\edit{Another line of human--computer interaction research that relates to forming mental models focuses on sensemaking~\citep{russell1993sensemaking,pirolli2011sensemaking}. Sensemaking refers to the process by which people collect and organize information and acquire ``situation awareness'' (i.e., build a mental model of the knowledge and data at hand). In the context of our work, sensemaking relates to people's understandings of machine learning models, while situation awareness facilitates insight and enables people to make intelligent decisions. Sensemaking research often involves designing tools to support rich interactions among people~\citep{goyal2016sensemaking} or between people and computational systems. Sensemaking processes are likely operating in our experiments when participants examine the model to simulate its predictions. Sensemaking processes may also be at play when participants detect and correct for the model's sizable mistakes, though we do not collect cognitive process measures to investigate their reasoning directly. We do, however, test a hypothesis about information overload that rests on an assumption about interference in the information-intake process.
}


\edit{Finally, there  is considerable research related to our experiments in the decision-making literature.
To date, much of this work has focused on people's aversion \citep{dawes1979,bazerman1985,DSM15} or proclivity \citep{logg2019algorithm} to trust computational decision-making aids, and ways to increase this trust \citep{dietvorst2018overcoming}. Other relevant decision-making work has endorsed the creation of simple or ``improper'' linear models that bear a strong resemblance to the models that we used in our experiments~\citep{dawes1979, jung2020simple, gigerenzer1996reasoning, goldstein2009fast}. Although decision-making researchers have tested the accuracies of these models in simulations, there have been far fewer tests of these models when used by people to aid their decision making. In our experiments, which we describe starting in the next section, we extend this line of research by taking a slightly different approach and asking how presentation differences in functionally identical models---specifically, differences in two factors thought to make machine learning models more interpretable---affect people's behavior.}






\section{Experiment 1: Predicting Apartment Selling Prices}
\label{sec:exp1}
Our first experiment was designed to measure the \medit{effects} of the
number of features and \medit{the} transparency \medit{of the model (i.e., whether the model internals are} clear or black box) on three \medit{measurable outcomes} that our literature \medit{review} revealed to be often associated with interpretability: laypeople's abilities to simulate \medit{a} model's predictions, \medit{the extent to which laypeople} follow a model's predictions when it \medit{is} beneficial \medit{for them} to do so, and \medit{laypeople's abilities to} detect when a model \medit{has made a} mistake. Before running
the experiment, we posited and pre-registered three
hypotheses, stated informally below:\edit{\footnote{Pre-registered hypotheses for this experiment are available at \url{https://aspredicted.org/xy5s6.pdf}.}}
\begin{itemize}[nosep]
\item[H1.] \textbf{Simulation.} \medit{Participants will better simulate the predictions of a clear model with few features.}
\item[H2.] \textbf{Deviation.} \medit{For} typical \medit{data points}, participants will more \medit{closely} follow (\medit{i.e., deviate less from}) the predictions of a clear model with \medit{few} features when it is beneficial for them to do so \medit{compared to} the predictions of a black-box model with \medit{more} features.
\item[H3.] \textbf{Detection of mistakes.} Participants \medit{assigned to} different \medit{experimental} conditions will \medit{be differently able} to \medit{detect and} correct \medit{for} the model's \medit{sizable mistakes on} unusual \medit{data points}.
\end{itemize}

We test\medit{ed} the first hypothesis by showing \medit{each participant} an apartment\medit{'s configuration (i.e., feature values)}, asking them to \medit{guess} what the model \medit{would} predict for \medit{the apartment'}s selling price, and \medit{then} comparing this \medit{prediction to} the model's prediction.
A small difference between these two quantities\medit{, which we refer to as the participant's simulation error,} indicate\medit{s} that \medit{the participant could better simulate the model's prediction}.

For the second hypothesis, we \medit{measured} the \medit{extent} to which \medit{each} participant \medit{deviated from the model's predictions} by\medit{, for each of the apartments with typical configurations,} showing them the model's prediction\medit{, asking them} for \medit{their own prediction of the apartment's selling price, and measuring the difference between these two quantities.}

We use\medit{d} the same measure for the third hypothesis, but \medit{focused on only the apartments with} unusual \medit{configurations}. \edit{Specifically, we said that a participant was able to detect and correct for the model's sizable mistakes if we saw large deviations between the model's predictions and their own predictions for the apartments with unusual configurations. We note that this does not necessarily imply that they understood why the model made the mistakes.}
We did not \medit{pre-register} any directional \medit{hypotheses} about which \medit{experimental} conditions would \medit{result in} participants \medit{being} more or less able to \medit{detect} and correct \medit{for the model's sizable mistakes}. On the one hand, if a participant better understands the model, \medit{they} may be better equipped to \medit{detect and} correct \medit{for its overly high predictions}. On the other hand, a participant may \medit{more closely} follow \medit{the model's predictions} if \medit{they} better understand \medit{it}. \edit{For this reason, we pre-registered our third hypothesis to be bi-directional, but we note that our intuition at the time was that participants who were shown a clear model would be better able to detect and correct for its sizable mistakes, compared to participants who were shown a black-box model.}

We additionally pre-registered our intent to analyze participants' prediction error\medit{s (i.e., how far their own predictions of the apartments' selling prices were from the actual selling prices)}, but again refrained from \medit{pre-registering} any directional \medit{hypotheses.}

\begin{figure*}[t]
  \captionsetup[subfigure]{aboveskip=-0.25pt,belowskip=-0.25pt}
  \centering
\begin{subfigure}[b]{0.5\textwidth}
                \includegraphics[width=\linewidth]{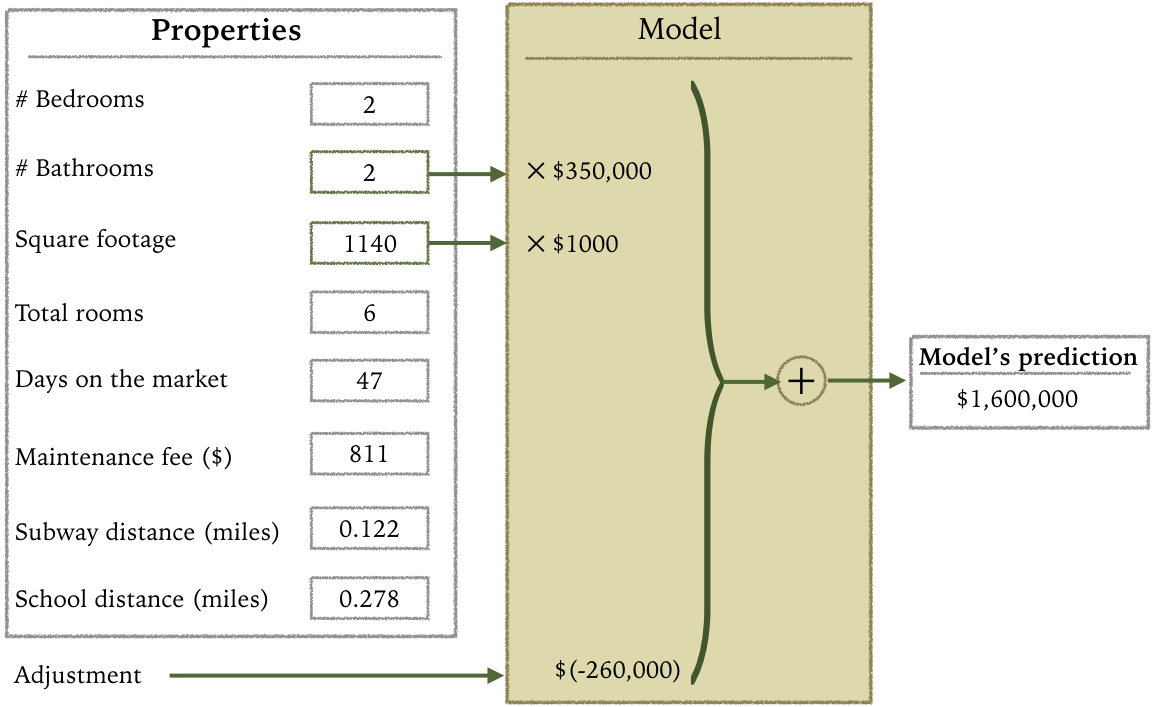}
				\caption{Clear, two-feature condition (\abr{clear-2}).}                \label{fig:screenshot_clear2}
        \end{subfigure}%
 \begin{subfigure}[b]{0.5\textwidth}
                \includegraphics[width=\linewidth]{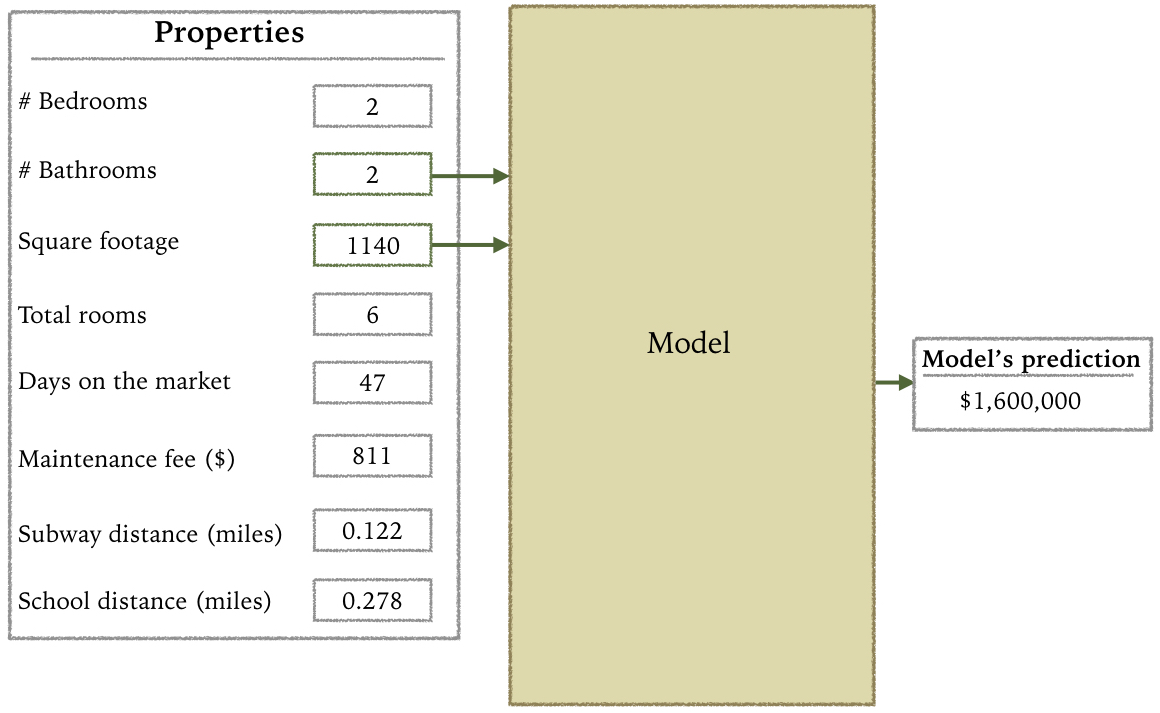}
                \caption{Black-box, two-feature condition (\abr{bb-2}).}
                \label{fig:screenshot_bb2}
        \end{subfigure}%
\\
~\\
        \begin{subfigure}[b]{0.5\textwidth}
                \includegraphics[width=\linewidth]{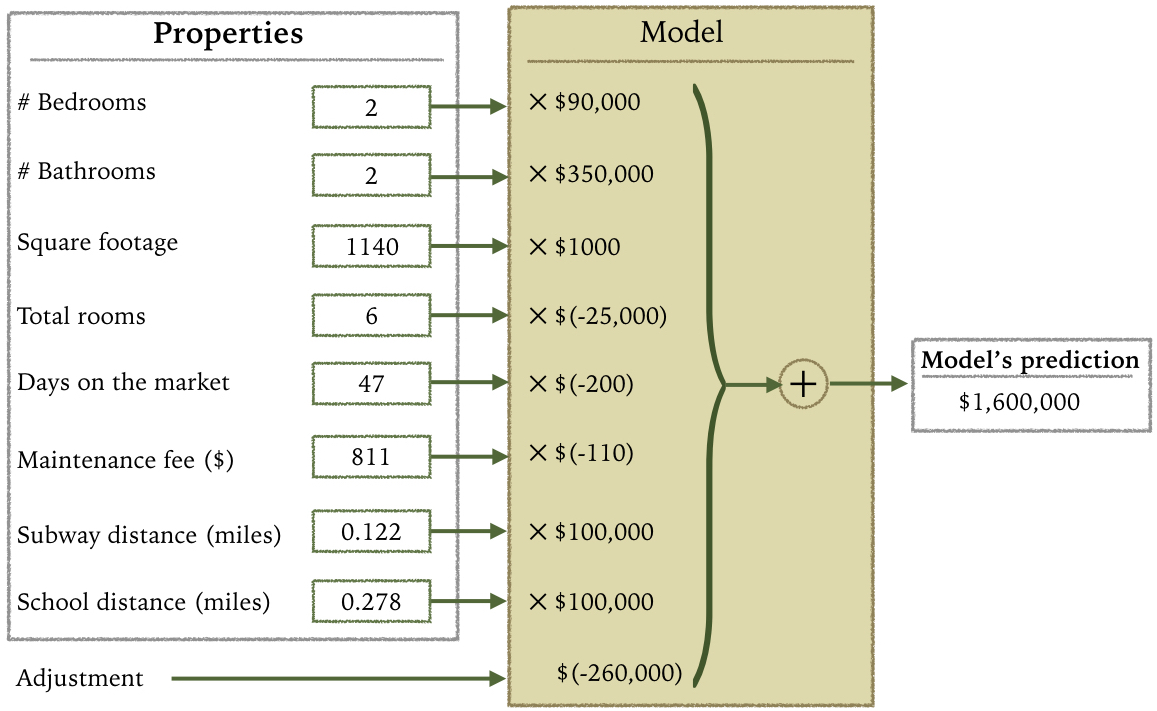}
				\caption{Clear, eight-feature condition (\abr{clear-8}).}                \label{fig:screenshot_clear8}
        \end{subfigure}%
                                \begin{subfigure}[b]{0.5\textwidth}
                \includegraphics[width=\linewidth]{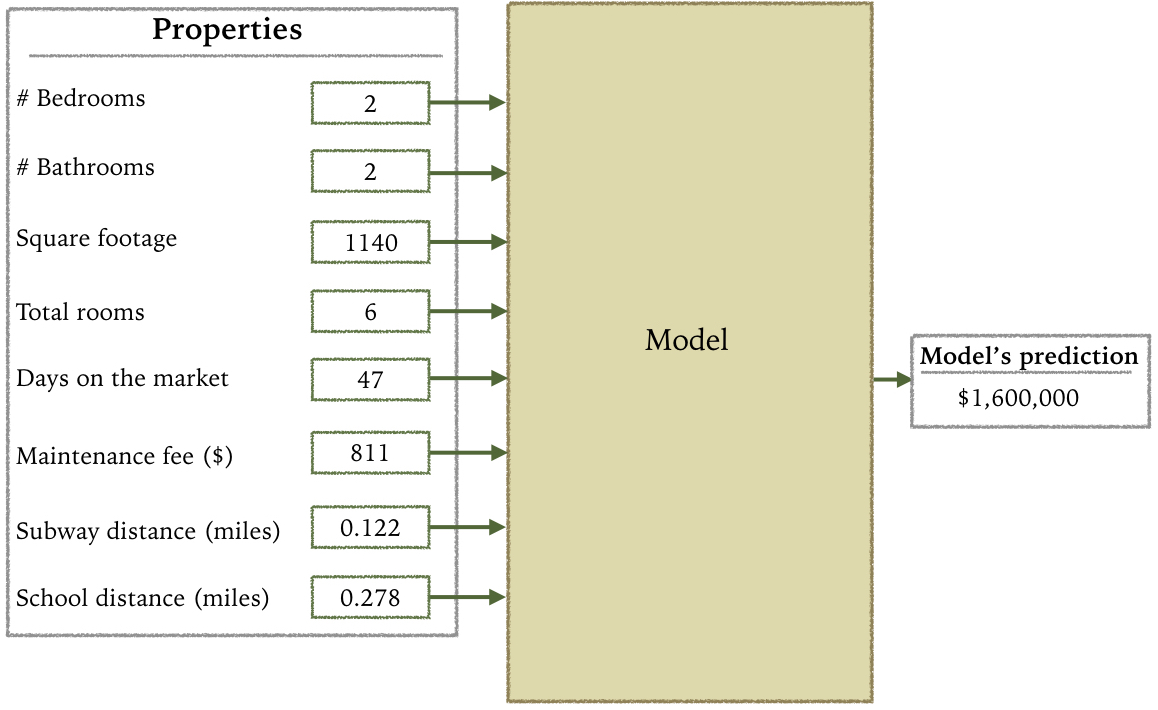}
			\caption{Black-box, eight-feature condition (\abr{bb-8}).}                \label{fig:screenshot_bb8}
        \end{subfigure}%
\caption{The four primary experimental conditions. In the conditions in the top row, the model used two features; in the conditions in the bottom row, the model used eight.  In the conditions on the left, the model internals were clear; in the conditions on the right, the model internals were black box.}
\label{fig:conditions}
\Description[]{Four images of how models were shown to participants in different conditions. The same apartment configuration (i.e., feature values) are shown in each image. The names of the features are provided in the main text. In the conditions involving clear models, the model coefficients are shown, while in the conditions involving black-box models, they are not. Arrows from feature values into a box labeled ``model'' show which features the model uses.}
\end{figure*}
\subsection{Experimental design}
\label{subsec:exp1_design}

\begin{figure*}
  \captionsetup[subfigure]{aboveskip=-1pt,belowskip=-1pt}
\centering
	\begin{framed}
        \begin{subfigure}[t!]{\textwidth}
                \centering
                \imagebox{40mm}{\includegraphics[width=\linewidth]{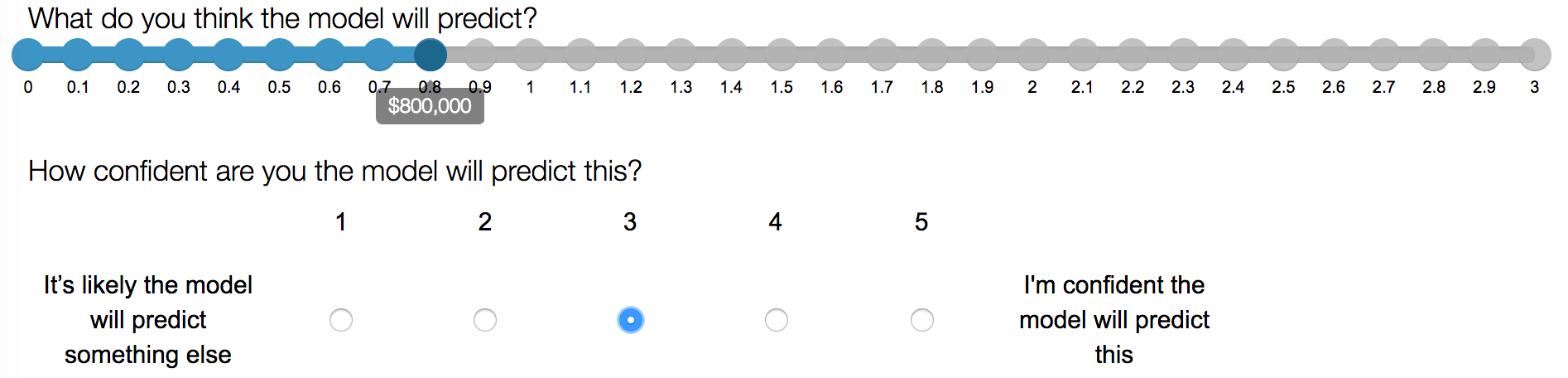}}
                \caption{Step 1: Participants were asked to guess \medit{what} the model \medit{would} predict and state their confidence\medit{ in this guess}.}
                \label{fig:exp1_step1}
                \Description[]{Image of a slider running from 0 to 3 million (dollars) under the
                question ``What do you think the model will predict?'' Beneath this
                is a question ``How confident are you the model will predict this?'' with
                a five-point Likert scale with endpoints ``It's likely the model will predict something else'' to ``I'm confident the model will predict this''.}
        \end{subfigure}%
	\end{framed}

        \begin{framed}
        \begin{subfigure}[t!]{\textwidth}
                \centering
                \imagebox{55mm}{\includegraphics[width=\linewidth]{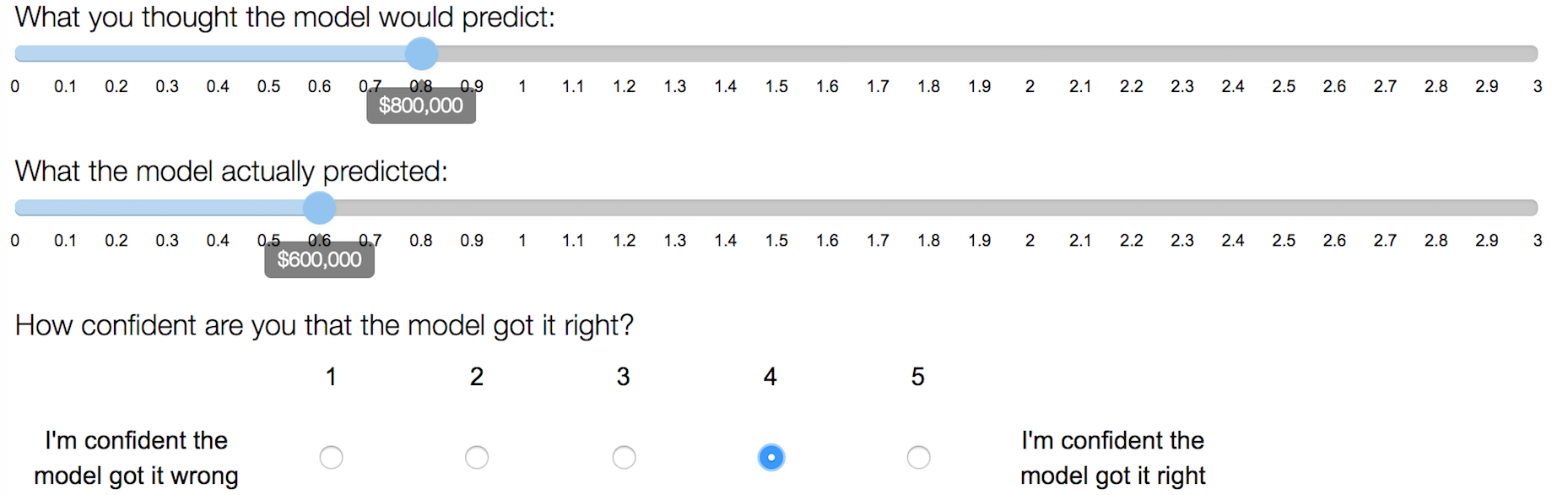}}
                \caption{Step 2: Participants were asked to state their confidence in the model's prediction.}
                \label{fig:exp1_step2}
                \Description[]{Image of a slider showing the participant's prediction (from step 1) under the words ``What you thought the model would predict''
                and an additional slider showing ``What the model actually predicted''.
                Beneath is is a question ``How confident are you that the model got it right"''
                with a five-point Likert scale with endpoints ``I'm confident the model got it wrong'' to ``I'm confident the model got it right''.
                }
        \end{subfigure}
        \end{framed}

        \begin{framed}
        \begin{subfigure}[t!]{\textwidth}
                \centering
                \imagebox{70mm}{\includegraphics[width=\linewidth]{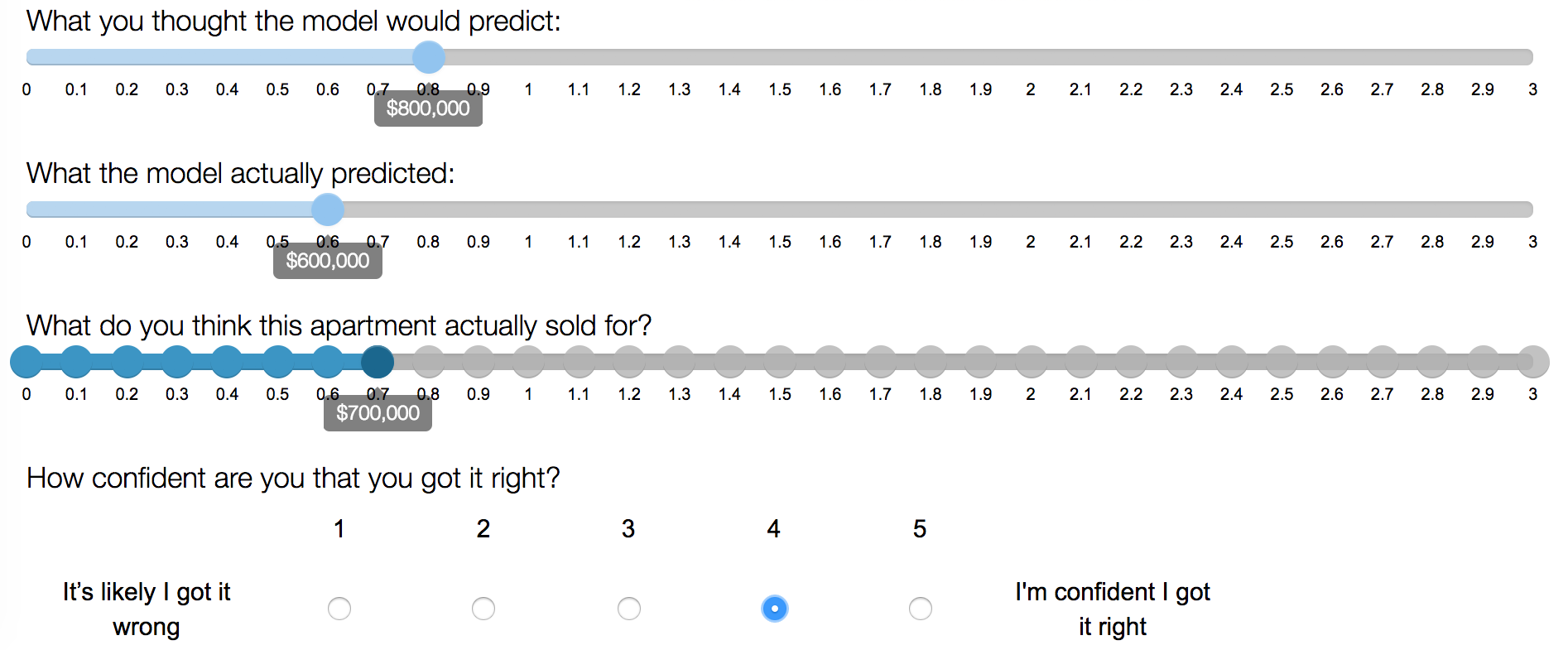}}
                \caption{Step 3: Participants were asked \medit{for} their own prediction and \medit{to} state their confidence \medit{in this prediction}.}
                \label{fig:exp1_step3}
                \Description[]{Image with same two sliders from step 2, with an additional
                slider underneath asking ``What do you think this apartment actually sold for?''. Beneath this is a question ``How confident are you that you got it right?''
                with a five-point Likert scale with endpoints ``It's likely I got it wrong'' to ``I'm confident I got it right''.}
              \end{subfigure}%
		\end{framed}
        \caption{Part of the testing phase from our first experiment.}\label{fig:UI_testing_phase}
\end{figure*}
As \medit{we} explained in \medit{Section~\ref{sec:introduction}}, we asked participants to predict \medit{the selling} prices \medit{of apartments in a single neighborhood in New York City} with the help of a machine learning model. \medit{To do this, we used} a $2 \times 2$ design:
\begin{itemize}[nosep]
\item Participants were randomly assigned to see either a \medit{two-feature model} (number of bathrooms and square footage---the two most predictive features) or \medit{an eight-feature model}.
\item Participants were randomly assigned to either see \medit{a clear} model (i.e., a  linear regression model with visible coefficients) or a \medit{black-box model}.
\end{itemize}
We additionally included a baseline condition in which there was no model available \medit{to participants}.

\medit{A}ll participants \medit{saw} the same set of apartments (i.e., the same feature values). \medit{The models were constrained to make the same predictions for these apartments, so participants saw} the same model prediction\medit{s} regardless of the experimental condition \medit{to which they were assigned (see Appendix~\ref{appndx:apartment_selection_details}). Furthermore, the accuracies of the models were nearly identical, as described below. What} varied between the \medit{experimental} conditions was therefore \medit{\emph{only the presentation of the model}.} \medit{This was a} key feature of our experimental design \medit{t}hat enabled us to run tightly controlled experiments.

Screenshots from each of the four primary experimental conditions \medit{(i.e., each experimental condition in our $2 \times 2$ design, but not the baseline condition)} are shown in Figure~\ref{fig:conditions}. \medit{We} note that \medit{each participant had access to} all eight feature values \medit{for each apartment, regardless of the experimental condition to which they were assigned}.

We ran the experiment on Amazon Mechanical Turk using psiTurk~\citep{gureckis2016psiturk}, an open-source platform for designing online experiments.
\edit{Multiple studies have shown that data from high-reputation Turkers is comparable to data from commercial panels and university pools when assessing outcomes such as attentiveness, honesty, and effort~\citep{coppock2019generalizing,paolacci2014,casler2013,mason2012,buhrmester2011}.}\edit{\footnote{\edit{Although recent research has shown that Turkers may manipulate their demographic information so as to be included in studies~\citep{sharpe2017mturk}, we only screened Turkers using two criteria, both of which are enforced by the platform and would require some effort to manipulate: country and approval rating. Moreover, even if some participants had manipulated their information, it would have had only a minor effect on our findings because we were not trying to estimate quantities relating to the entire population of participants, but were instead trying to detect differences between the experimental conditions. That is, the proportion of Turkers with manipulated information would be, on average, the same for each condition, thereby permitting valid measurement of randomly assigned treatment effects.}}}

We recruited 1,250 participants, all located in the U.S., with approval ratings greater than $97\%$. \medit{We} randomly assigned \medit{participants} to the \medit{experimental} conditions (\abr{clear-2}, $N=248$; \abr{clear-8}, $N=247$; \abr{bb-2}, $N=247$; \abr{bb-8}, $N=256$; and \abr{no-model}, $N=252$). Each participant received a flat payment of \$2.50. The experiment was approved by our institutional review board.

Participants were first shown detailed instructions, including, in the conditions \medit{involving clear models}, a simple English description of the corresponding two- or eight-feature model (\medit{see} Appendix~\ref{appndx:instructions_exp1}).
To ensure \medit{that participants} understood these instructions, \medit{each} participant \medit{was} required to answer a multiple choice question \medit{about} the number of features used by the model before proceeding with the experiment, \medit{which consisted of two phases}.

The \emph{training phase} familiarized participants with both the domain \medit{(i.e., real-estate valuation)} and the model's predictions. Participants were shown ten apartments in a random order.
In the four primary experimental conditions, participants were shown the model's prediction of each apartment's \medit{selling} price, asked \medit{for} their own prediction \medit{of the apartment's selling price}, and then shown the apartment's actual \medit{selling} price.
 In the baseline condition \medit{(i.e., no model)}, participants were asked to predict the \medit{selling} price \medit{for} each apartment and then shown \medit{its} actual \medit{selling} price.

In the \emph{testing phase}, participants were shown twelve apartments \medit{that} they had not previously seen.
The order of the first ten \medit{apartments} was randomized, while the remaining two \medit{apartments} always appeared last, for {the} reasons described below.
In the four primary experimental conditions, participants were asked to guess what the model would predict for each apartment\medit{'s selling price} (i.e., simulate the model\medit{'s prediction}) and to \medit{state their confidence} in this guess on a five-point scale (\medit{see} Figure~\ref{fig:exp1_step1}).
They were then shown the model's prediction and asked to \medit{state their confidence in that prediction (see} Figure~\ref{fig:exp1_step2}).
Finally, they were asked \medit{for} their own prediction of the apartment's \medit{selling} price and to \medit{state their confidence} in this prediction (Figure~\ref{fig:exp1_step3}).
 In the baseline condition, participants \medit{just} were asked to predict the \medit{selling} price \medit{of} each apartment and to \medit{state} their confidence \medit{in this prediction}.

\medit{We selected} the apartments from a data set of apartments sold between 2013 and 2015 on the Upper West Side of New York City, taken from StreetEasy.com, a popular real\medit{-}estate website.
To create the models, we first fit a two-feature linear regression model (i.e., estimated the model's coefficients) \medit{using} this data set \medit{and} round\medit{ed the} coefficients for readability.\footnote{For each coefficient, we found a round number that was within one quarter of a standard error of the estimated coefficient.}
   To \medit{ensure that} the models \medit{were} as similar as possible, we fixed the \medit{intercept and the} coefficients for number of bathrooms and square footage in an eight-feature model to match those of the two-feature model, and then fit \medit{the model (i.e., estimated the model's coefficients for} the remaining six features\medit{) and followed} the same rounding procedure \medit{as with the two-feature model. The rounded coefficients for both models are shown in Figure~\ref{fig:conditions}}. The models explain $82\%$ of the variance in the apartments' \medit{selling} prices. When presenting the models' predictions to participants, we rounded \medit{each} prediction to the nearest \$100,000.

\medit{All participants saw} the same set of apartments \medit{(i.e., the same feature values)} because randomizing the selection would \medit{have} introduce\medit{d} additional noise and reduce\medit{d} the power of the experiment, making it harder to spot differences between \medit{the experimental} conditions.
To enable comparisons \medit{between the} experimental conditions, the ten apartments in the training phase and the first ten apartments in the testing phase were selected from the apartments in our data set for which the rounded predictions of the two- and eight-feature models \medit{were the same}. We selected the \medit{apartments} to cover a representative range of \medit{model} prediction errors \medit{(i.e., how far the models' predictions were from the apartments' actual selling prices). We provide d}etails \medit{of} the apartment\medit{s'} configurations \medit{(i.e., feature values)} and \medit{our selection procedure} in Appendix~\ref{appndx:apartment_selection_details}.

\medit{We used t}he last two apartments in the testing phase to test our third hypothesis\medit{---}namely, that participants \medit{assigned to different experimental conditions will be differently able to detect and correct for the model's sizable mistakes on unusual data points}.
\medit{Ideally}, we would have used two apartment\medit{s} with \medit{unusual configurations for which both} models \medit{made} the same \medit{sizable mistakes}.
Unfortunately, there were no such apartments in our data set, so we \medit{selected (in one case) and synthetically generated (in the other) two apartments} to test different aspects of our hypothesis.
These apartments' \medit{configurations} exploited the models' large coefficient (\$350,000) for number of bathrooms.
The first \medit{apartment (``apartment 11'')} was a one-bedroom, two-bathroom apartment (selected from our data set) for which both models made overly high, but different, predictions.
\medit{As a result, c}omparisons between the \medit{conditions involving the two-feature model and the} conditions \medit{involving the eight-feature model were therefore impossible, although we were able to analyze participants' prediction errors because these did not rely on the models' predictions.}
The second \medit{apartment} (``apartment 12'') was a synthetically generated one-bedroom, three-bathroom\medit{, 726-square-foot} apartment for which both models made the same overly high prediction, allowing \medit{us to make} comparisons between all \medit{four primary experimental} conditions, but ruling out \medit{analyses of participants'} prediction error\medit{s because the apartment did not have an actual selling price.} \medit{We emphasize that even though it did not have an actual selling price, we are confident that it would have been overpriced by the models because of its three bathrooms squeezed into only 726 square feet.} Apartments 11 and 12 were always shown last to avoid the phenomenon in which people trust a model less after seeing it make a mistake~\citep{DSM15}.

\subsection{Findings}

\begin{figure*}[t!]
  \captionsetup[subfigure]{aboveskip=-2pt,belowskip=-2pt}
  \centering
  \textbf{\fontfamily{phv}\selectfont Experiment 1: New York City prices}\par\medskip
        \begin{subfigure}[b]{0.5\textwidth}
                \includegraphics[width=\linewidth]{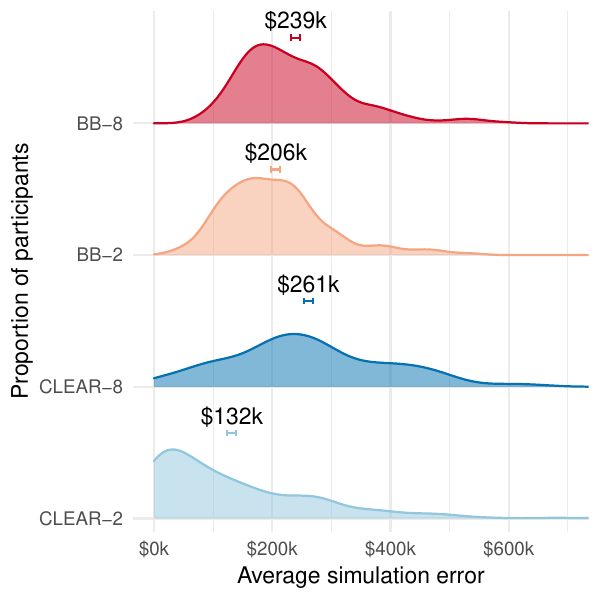}
                \caption{}
                \label{fig:exp1_simulation_error}
                \Description[]{Four density plots showing the distributions of simulation errors by condition. Error in the condition involving the clear-2 model is markedly lower than in the other conditions}
        \end{subfigure}%
        \begin{subfigure}[b]{0.5\textwidth}
                \includegraphics[width=\linewidth]{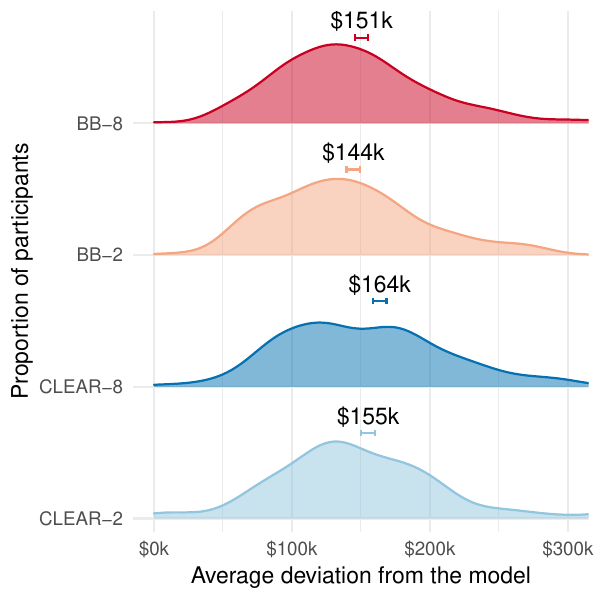}
               \caption{}
                \label{fig:exp1_dev_from_model}
                \Description[]{Four density plots showing the distribution of deviations by condition. There is a somewhat similar pattern across the four conditions.}
        \end{subfigure}
        \caption{Results from our first experiment: density plots for participants' (a) mean simulation errors and (b) mean deviations from the model's predictions. Numbers in each subplot indicate average values over all participants in the corresponding condition, while error bars indicate one standard error.}\label{fig:exp1_main_results}

  \captionsetup[subfigure]{aboveskip=-2pt,belowskip=-2pt}
  \centering
  \textbf{\fontfamily{phv}\selectfont Experiment 2: Representative U.S. prices}\par\medskip
        \begin{subfigure}[b]{0.5\textwidth}
                \includegraphics[width=\linewidth]{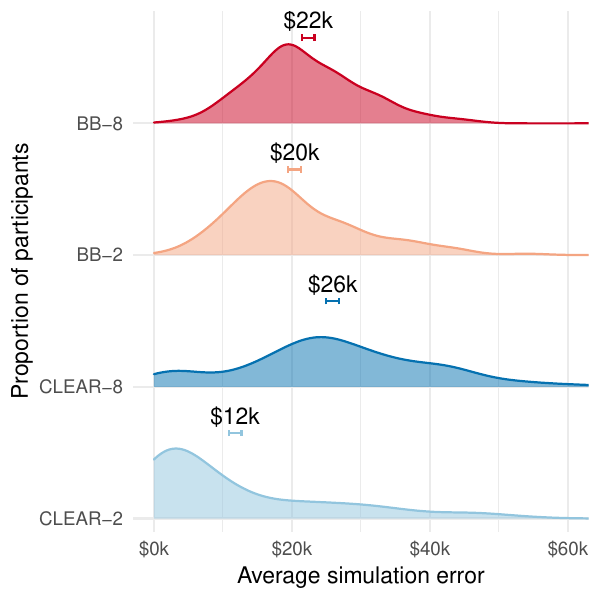}
                \caption{}
                \label{fig:exp2_simulation_error}
                \Description[]{Four density plots showing the distributions of simulation errors by condition. Error in the condition involving the clear-2 model is markedly lower than in the other conditions}
        \end{subfigure}%
        \begin{subfigure}[b]{0.5\textwidth}
                \includegraphics[width=\linewidth]{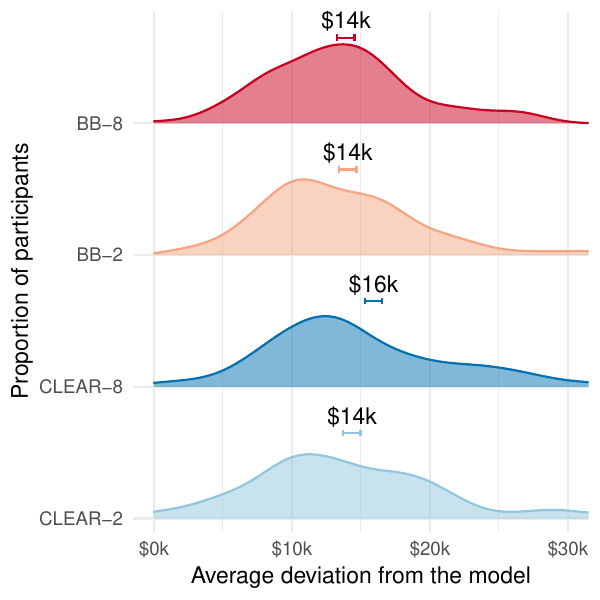}
               \caption{}
                \label{fig:exp2_dev_from_model}
                \Description[]{Four density plots showing the distribution of deviations by condition. There is a somewhat similar pattern across the four conditions.}
        \end{subfigure}

            \caption{Results from our second experiment,
              which replicate the findings from our first experiment.
            }\label{fig:exp2_main_results}
\clearpage
\end{figure*}

Having run \medit{the} experiment, we compared participants' behavior across the conditions.\footnote{For \medit{each} of our experiments, we report all sample sizes, conditions, data exclusions, and measures for the main analyses that were \medit{described} in our pre-registration documents.
  We determined the sample size for our first experiment based on estimates from a small pilot experiment\medit{, which} enable\medit{d us to} detect a difference of at least \$50,000 in deviation between the \medit{condition involving the clear, two-feature model and the condition involving the black-box, eight-feature model with 80\% power}. For \medit{our} subsequent experiments, \medit{however,} we adjusted the sample size to target a power of 80\% or more. \medit{We provide f}ull distributions of \medit{participants'} responses in Appendix~\ref{appndx:distributions} and details of \medit{our} statistical tests in Appendix~\ref{appndx:anova_tables}.}
Doing so required us to compare multiple responses \medit{(i.e., data about multiple apartments)} from multiple participants, which was complicated by possible correlations among \medit{each} participant's responses. For example, some \medit{participants} might \medit{have} consistently overestimate\medit{d} the apartments' \medit{selling prices} regardless of the condition \medit{to which they were} assigned, while others might \medit{have} consistently provide\medit{d} underestimates. We addressed this by fitting a mixed-effects model for each \medit{measurable outcome} of interest to capture differences \medit{between} conditions while controlling for participant-level effects---a standard approach for analyzing repeated measure experimental designs~\citep{B+15}.
We derived all statistical tests from these models. Bar plots and mean outcomes in the density plots \medit{correspond to} average \medit{values} ($\pm$ one standard error) by condition from the fitted \medit{mixed-effects} models.
To test our hypotheses, we ran contrasts and \medit{calculated} degrees of freedom, test statistics, and $p$-values under these models.
Unless otherwise noted, all plots and statistical tests correspond to just the first ten apartments from the testing phase.\edit{\footnote{All the data and code needed to reproduce our results are available at \url{https://github.com/Foroughp/Manipulating-and-Measuring-Model-Interpretability}.}}

\medit{Our fi}ndings are as follows\medit{:}

H1. \textbf{Simulation.} We defined \medit{each} participant's simulation error for each apartment to be $\lvert{m - u_m}\rvert$\medit{---i.e.}, the absolute \medit{difference} between the model's prediction of the apartment's selling price $m$ and the participant's guess for \medit{the model's} prediction $u_m$. Figure~\ref{fig:exp1_simulation_error} \medit{contains density plots for participants' mean} simulation errors.
Participants \medit{assigned to the condition involving} the \medit{clear, two-feature model} had, on average, lower simulation error\medit{s} compared to participants \medit{assigned to} the other primary experimental conditions ($t\left(994\right)=-12.06$, $p < 0.001$).
This \medit{means} that, as hypothesized, participants \medit{could} better \medit{simulate the predictions of the clear, two-feature model}.

H2. \textbf{Deviation.} We \medit{defined each} participant's deviation from the model\medit{'s prediction of each apartment's selling price to be} $\lvert{m - u_a}\rvert$\medit{---i.e.,} the absolute difference between the model's prediction of the apartment's selling price $m$ and the participant's \medit{own} prediction of the apartment's \medit{selling price $u_a$}. Figure~\ref{fig:exp1_dev_from_model} shows that contrary to our second hypothesis, we \medit{did not find a} significant difference in \medit{the extent to which} participants \medit{followed the predictions of the clear, two-feature model when it was beneficial for them to do so compared to the predictions of the black-box, eight-feature model} ($t\left(994\right)=0.67$, $p = 0.5$).

\begin{figure*}[t]
  \captionsetup[subfigure]{aboveskip=-2pt,belowskip=-2pt}
  \centering
             \begin{subfigure}[b]{0.5\linewidth}
                \includegraphics[width=\linewidth]{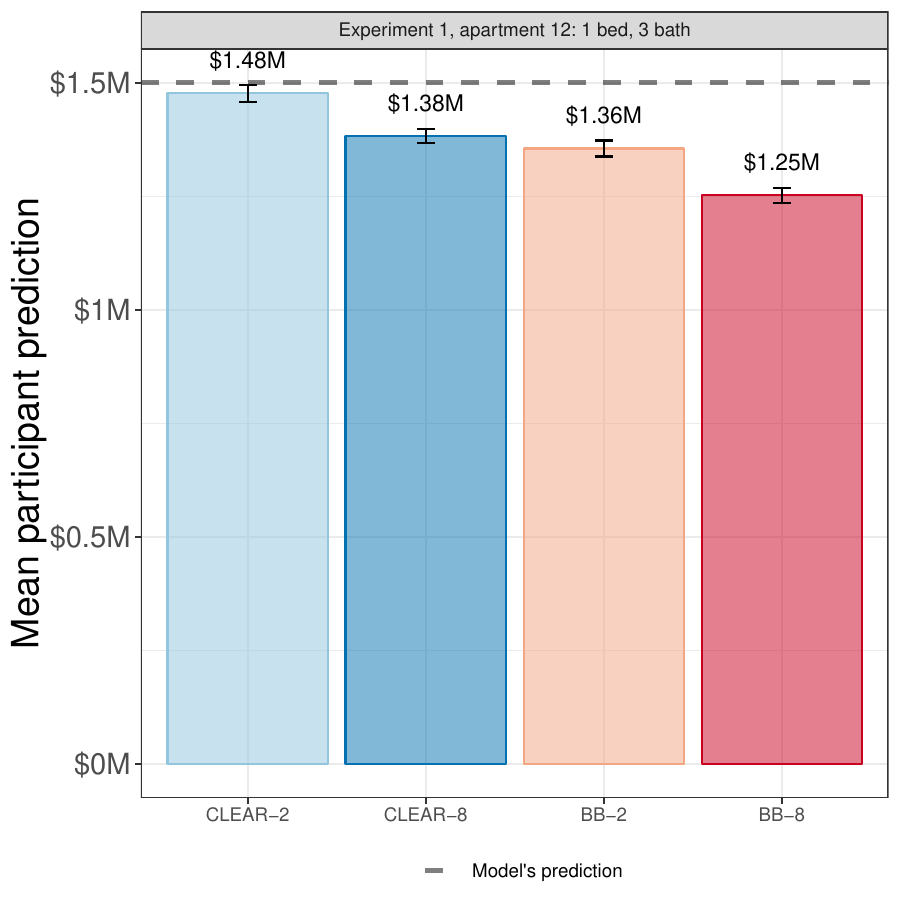}
                \caption{}
                \label{fig:exp1_q12_pred}
                \Description[]{Bar chart showing participants' mean predictions of apartment 12's selling price by condition for our first experiment. A dashed line shows the model's prediction of \$1.5 million. The mean for participants assigned to the condition involving the clear-2 is closest to the model's prediction at \$1.48 million .}
        \end{subfigure}%
        \begin{subfigure}[b]{0.5\linewidth}
		\includegraphics[width=\linewidth]{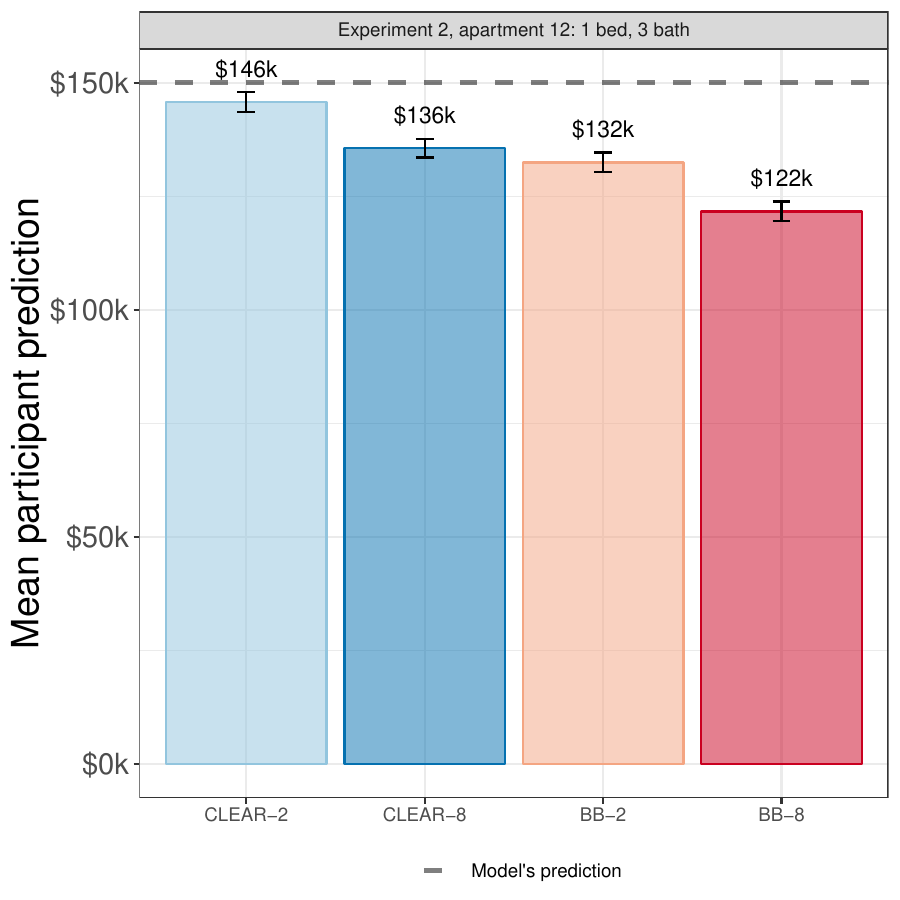}
                \caption{}
                \label{fig:exp2_q12_pred}
                \Description[]{Bar chart showing participants' mean predictions of apartment 12's selling price by condition for our second experiment. A dashed line shows the model's prediction of \$150,000. The mean for the condition involving the clear-2 model is closest to the model's prediction at \$146,000.}
              \end{subfigure}
\caption{Participants' mean predictions of apartment 12's selling price in (a) our first experiment and (b) our second experiment. Horizontal lines indicate the models' predictions and error bars indicate one standard error.}
\end{figure*}

H3. \textbf{Detection of mistakes.} \medit{As explained above, w}e used the last two apartments in the testing phase (apartments 11 and 12) to test our third hypothesis.
The models \medit{made} overly high predictions \medit{for} these \medit{apartments} because \medit{of} the\medit{ir unusual configurations}. For both apartments\medit{,} participants \medit{assigned to} the four primary \medit{experimental} conditions \medit{predicted higher selling} prices compared to participants \medit{assigned to} the baseline condition (i.e., no model). We suspect that this is because \medit{participants} anchored \medit{on} the models' predictions.
For apartment 11, we found no significant difference\medit{s} in participants' deviation\medit{s} from the model's prediction\medit{s} between the four primary experimental conditions ($F\left(3,994\right) = 1.03$, $p = 0.379$ under a one-way ANOVA). \medit{In other words, we found that participants assigned to different experimental conditions were similarly able to detect and correct for the model's overly high prediction for apartment 11.} For apartment 12, a one-way ANOVA revealed a significant difference \medit{in participants' deviations from the model's predictions} between the four primary experimental conditions ($F\left(3,994\right) = 4.42$, $p = 0.004$).
Participants assigned to the conditions \medit{involving clear models} deviated from the model\medit{s'} prediction, on average, less \medit{compared to} participants assigned to the \medit{conditions involving} black-box \medit{models} ($F\left(1,994\right) = 8.81$ , $p = 0.003$ for the main
effect of \medit{the} transparency \medit{of the model}, see Figure~\ref{fig:exp1_q12_pred}). \medit{This finding contradicts our intuition when designing the experiment, which was that participants who were shown a clear model would be better able to detect and correct for its sizable mistakes compared to participants who were shown a black-box model.} We explore this finding in more detail \medit{in Section~\ref{sec:exp4}}.

\begin{figure*}[t]
  \captionsetup[subfigure]{aboveskip=-2pt,belowskip=-2pt}
  \centering
        \begin{subfigure}[b]{0.5\textwidth}
             \hspace*{.5cm}\textbf{\fontfamily{phv}\selectfont Experiment 1:\\ New York City prices}\par\medskip
                \includegraphics[width=\linewidth]{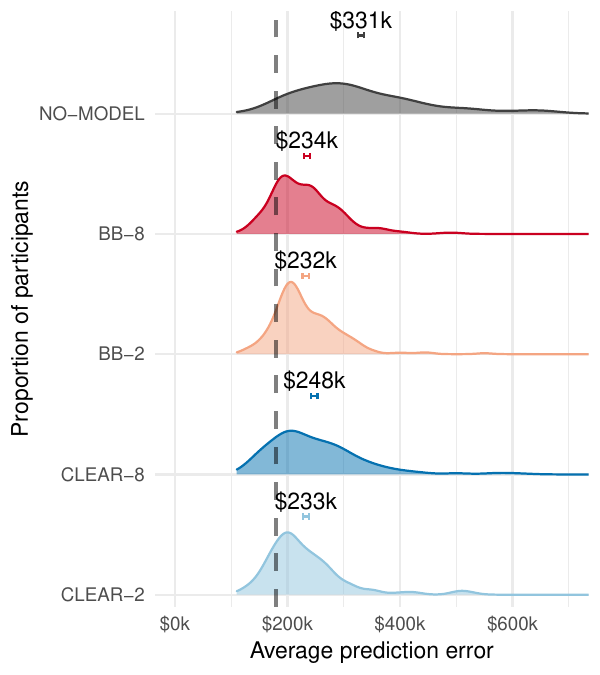}
                \caption{}
                \label{fig:exp1_prediction_error}
                \Description[]{Density plots for four primary experimental conditions are similar. Prediction errors in the baseline condition (i.e., no model) are considerably larger.}
        \end{subfigure}%
        \begin{subfigure}[b]{0.5\textwidth}
          \hspace*{.5cm}\textbf{\fontfamily{phv}\selectfont Experiment 2:\\ Representative U.S. prices}\par\medskip

                \includegraphics[width=\linewidth]{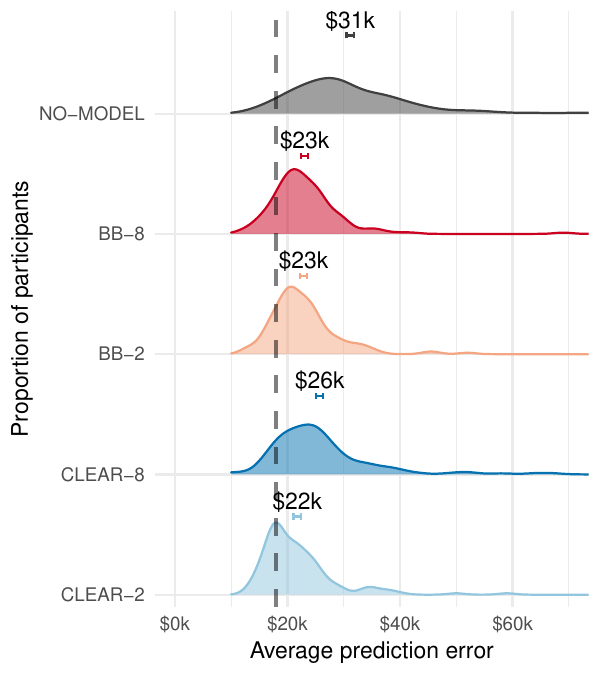}
               \caption{}
                \label{fig:exp2_prediction_error}
        \end{subfigure}
            \caption{Density plots for participants' mean prediction errors in our first experiment (left) and in our second experiment (right). Numbers in each subplot indicate average values over all participants in the corresponding condition, while error bars indicate one standard error. Vertical lines indicate the model's mean prediction error.}
        \label{fig:prediction_error}
        \Description[]{Density plots for the four primary experimental conditions are similar. Prediction errors in the baseline (no-model) condition are considerably larger.}
\end{figure*}

We \medit{also conducted} some post-hoc analyses. First, we \medit{analyzed} participants' \medit{stated} confidences in the models' predictions \medit{for each apartment}. \medit{Although} we \medit{did not} pre-register a hypothesis \medit{about this, we found} an interesting difference between participants' stated confidences and \medit{their} revealed behavior. Specifically, even though participants \medit{assigned to the condition involving the clear, two-feature model stated} that they were more confident in the \medit{model's} predictions\medit{ compared to participants assigned to} the condition \medit{involving} the \medit{black-box, eight-feature model (on average,} a difference of .25 on a five-point scale from ``I'm confident the model got it wrong'' to ``I'm confident the model got it right\medit{,''} ($t\left(994\right)=4.27$, $p < 0.001$)), their \medit{behavior} did not reflect this. \medit{We found no significant differences in the extent to which participants followed the model's predictions between the four primary experimental conditions.}

\medit{Our} second post-hoc analysis \medit{involved participants'} prediction error\medit{s}. We \medit{defined each participant's prediction error for each apartment to be} $\lvert{a - u_a}\rvert$\medit{---i.e.}, the absolute difference between the apartment's actual \medit{selling} price $a$, and the participant's \medit{own} prediction of the apartment's \medit{selling} price $u_a$. \medit{A one-way ANOVA did not reveal any significant differences in participants' prediction errors between the four primary experimental conditions}
($F\left(3,994\right) = 2.43$, $p = 0.06$).

As shown in Figure~\ref{fig:exp1_prediction_error}, we also found that \medit{for the} apartments with \medit{typical} configuration\medit{s}, participants \medit{assigned to} the four primary experimental conditions had\medit{, on average,} higher \medit{prediction} error\medit{s} than the model, but \medit{lower prediction errors} than \medit{participants assigned to} the baseline condition ($t\left(1245\right) = 15.28$, $p < 0.001$ for the comparison of the baseline \medit{condition} with the four primary \medit{experimental conditions)}, suggesting that \medit{using} a model was advantageous. \medit{In contrast,} participants' prediction error\medit{s for} apartment 11 revealed \medit{the opposite} pattern: participants \medit{assigned to} the four primary \medit{experimental} conditions had, \medit{on average}, lower prediction error\medit{s} than the model but higher prediction error\medit{s} than \medit{participants assigned to} the baseline condition ($t\left(1245\right) = -7.99$, $p < 0.001$). \medit{Using the} model helped \medit{participants more accurately predict the selling prices of the apartments with typical configurations, but hindered them when predicting the selling price of an apartment with an unusual configuration for which the model had made a sizable mistake.}

\medit{Additionally, we found that using a clear model further hampered participants when making their own predictions about apartment 11's selling price:}
participants who were shown a clear model made, on average, \medit{less accurate} predictions compared to participants who were shown a black-box model ($F\left(1,994\right) = 31.98$, $p < 0.001$ for the main effect of \medit{the} transparency \medit{of the model} under a two-way ANOVA).

\medit{To summarize, as hypothesized}, we found that participants who were shown \medit{the clear, two-}feature \medit{model could} better simulate the model's predictions. However, we did not find that they followed \medit{its} predictions more closely when it would have been beneficial \medit{for them} to do so. We also found that, contrary to our intuition, participants who were shown a clear model were less able to \medit{detect and} correct \medit{for} the model\medit{'s} sizable mistakes on unusual data points. Finally, we found no differences in participants' prediction errors between the four primary experimental conditions. We also found that \medit{that using a model is advantageous, but that participants would have been better off (i.e., had lower prediction errors) had they followed the model's predictions for the apartments with typical configurations.}


\section{Experiment 2: Representative U.S. Prices}
\label{sec:exp2}
One potential \medit{concern about} our first experiment is that participants' lack of familiarity with New York City's unusually high prices \medit{might have} influence\medit{d} the \medit{extent} to \medit{which they} follow\medit{ed} the model\medit{s or their} abilit\medit{ies} to detect \medit{and correct for} the model\medit{s'} \medit{sizable mistakes}.
Our second experiment was \medit{therefore} designed as a robustness check \medit{targeted at} this \medit{concern.

In this experiment, we} replicat\medit{ed} our first experiment \medit{but} with \medit{the} apartment\medit{s' selling} prices and maintenance fees scaled down to match median prices in the U.S. Before running this experiment we \medit{again posited and} pre-registered three hypotheses.\edit{\footnote{Pre-registered hypotheses for this experiment are available at \url{https://aspredicted.org/3bv8i.pdf}.}} The first two hypotheses (H4 and H5) were identical to \medit{the first two hypotheses from our} first experiment. \medit{However, w}e made the third hypothesis (H6) more precise than \medit{the third hypothesis from our} first experiment \medit{to reflect the findings from} \medit{that experiment, as well as our findings from a small pilot experiment} with scaled-down prices. This hypothesis is stated informally below:
\begin{itemize}[nosep]
\item[H6.] \textbf{Detection of mistakes.} Participants \medit{who are shown a clear model} will be less \medit{able} to \medit{detect and} correct \medit{for the model's sizable mistakes on} unusual \medit{data points}, and this effect will be more prominent \medit{for more unusual data points (i.e., for apartment 12 compared with apartment 11)}.
\end{itemize}

\subsection{Experimental design}
\label{subsec:exp2_design}

We first scaled down the apartment\medit{s' selling} prices and maintenance fees by a factor of ten. To account for this change, we also scaled down all coefficients (except for the coefficient for maintenance fee) by a factor of ten. Apart from the description of the neighborhood from which the apartments were selected, the experimental design was unchanged \medit{from our first experiment}. We again ran the experiment on Amazon Mechanical Turk \medit{using psiTurk}. We excluded \medit{Turkers} who had participated in our first experiment, and recruited 750 new participants\medit{,} all of whom satisfied the \medit{screening} criteria from \medit{our} first experiment. \medit{W}e randomly assigned \medit{participants} to the \medit{experimental} conditions (\abr{clear-2}, $N=150$; \abr{clear-8}, $N=150$; \abr{bb-2}, $N=147$; \abr{bb-8}, $N=151$; and \abr{no-model}, $N=152$). \medit{Again,} each participant received a flat payment of $\$2.50$.

\subsection{Findings}
\label{subsec:exp2_results}
The \medit{findings from our first} experiment replicated \medit{quite closely.}

H4. \textbf{Simulation.} As hypothesized, and \medit{as} shown in Figure~\ref{fig:exp2_simulation_error}, participants \medit{assigned to} the \medit{condition involving the clear, two-feature model} had, on average, lower simulation error\medit{s} \medit{compared to} participants \medit{assigned to} the other primary experimental conditions ($t\left(594\right)=-10.41$, $p < 0.001$). This is in line with the finding\medit{s} from \medit{our} first experiment.

H5. \textbf{Deviation.} Also in line with the findings from our first experiment, but contrary to our hypothesis, we found no significant difference in the extent to which participants followed the predictions of the clear two-feature model when it was beneficial for them to do so compared to the predictions of the black-box, eight-feature model ($t\left(594\right) = 0.49$, $p = 0.626$, see Figure~\ref{fig:exp2_dev_from_model}).

H6. \textbf{Detection of mistakes.}
For apartment 11, \medit{although} a one-way ANOVA revealed a significant difference \medit{between} the four primary \medit{experimental} conditions ($F\left(3,594\right) = 3.00$ , $p  = 0.03$), as \medit{was the case in our first e}xperiment, we found no significant difference\medit{s between the conditions involving clear models and the conditions involving black-box models} ($t\left(594\right) = -1.82$, $p = 0.069$), perhaps because apartment 11's configuration \medit{was} not sufficiently unusual. For apartment 12, in line with \medit{the} findings from \medit{our first e}xperiment, \medit{and as hypothesized}, a one-way ANOVA revealed a significant difference \medit{between} the four primary \medit{experimental} conditions ($F\left(3,594\right) = 7.96, p < 0.001$). In particular, participants \medit{assigned to} conditions \medit{involving clear models followed} the model's prediction, on average, \medit{ more closely} than participants \medit{assigned to conditions involving} black-box \medit{models, indicating that they were \emph{less} able to detect and correct} for \medit{the model's overly high prediction, thereby} resulting in an even worse final prediction \medit{for} the apartment's \medit{selling} price ($t\left(594\right) = -4.16$, $p < 0.001$, see Figure~\ref{fig:exp2_q12_pred}).

\medit{We again} conducted some post-hoc analyses. In contrast to the findings from our first experiment, we found no significant difference in participants' stated confidences between the condition involving the clear, two-feature model and the condition involving the black-box, eight-feature model ($t\left(594\right)=1.03$, $p = 0.303$). We note that the effect size of the difference in \medit{our first e}xperiment was small (Cohen's $d$ of 0.23) and even smaller in \medit{our second e}xperiment (Cohen's $d$ of 0.07), which was identical \medit{to the first}, except for prices. We also note that there was no significant difference in \medit{the extent to which participants followed the model's predictions when it was beneficial for them to do so} in either experiment.

\medit{We also analyzed participants'} prediction error\medit{s}. Here, a one-way ANOVA did reveal a significant difference in \medit{participants'} prediction error\medit{s between} the four primary \medit{experimental} conditions ($F\left(3,594\right) = 8.60$, $p < 0.001$). That said, the maximum pairwise difference in prediction error between the four primary experimental conditions is not large (\$4,000 or roughly 3\% of the average selling price, which was \$120,000).

\medit{In line with the findings from our first experiment, we found that for the apartments with typical configurations}, participants \medit{assigned to} the four primary \medit{experimental} conditions had\medit{, on average,} higher \medit{prediction} error\medit{s} than the model, but \medit{lower prediction errors} than participants \medit{assigned to} the baseline condition ($t\left(745\right) = 10.62$, $p < 0.001$). \medit{Again, we found} the \medit{opposite pattern} for apartment 11\medit{: participants assigned to the four primary experimental conditions had, on average, lower prediction errors than the model, but higher prediction errors than participants assigned to the baseline condition} ($t\left(745\right) = -6.41$, $p < 0.001$). \medit{We also found, in line with the findings from our first e}xperiment, \medit{that using a clear model} further \medit{hampered participants when making their own} predictions \medit{about} apartment 11\medit{'s selling price}: participants who were shown a clear model made \medit{less accurate} predictions\medit{ compared to participants} who were shown a black-box model ($F\left(1,594\right) = 7.16$, $p = 0.008$ for the main effect of the transparency of the model under a two-way ANOVA).

\medit{To summarize, the main findings from our second e}xperiment \medit{closely} replicate the \medit{findings from our first e}xperiment, suggesting that \medit{New York City's unusually high prices did not influence participants' behavior. In b}oth experiments\medit{, we found} that participants who \medit{were shown a clear, two-feature} model \medit{could} better simulate the model's predictions. However, they \medit{did} not follow the \medit{model's} predictions \medit{more closely when it was beneficial for them to do so. They were also less able} to detect \medit{and correct for} the model\medit{'s sizable mistakes on unusual data points}.


\section{Experiment 3: Weight of Advice}
\label{sec:exp3}
In our first two experiments, \medit{we did not find a significant
difference in the extent to which participants followed the
predictions of the clear, two-feature model when it was beneficial for them to do so compared to the predictions of
the black-box, eight-feature model, measured in terms of the absolute
difference between the model's prediction and the participant's own
prediction for each of the apartments with typical
configurations. Because this finding was contrary to our hypotheses,
we wondered whether using an alternative way to measure the extent to which people follow a model's predictions would yield a different finding.} In \medit{our} third experiment, we \medit{therefore used} \emph{weight of advice}\medit{---a} measure \medit{commonly} used in the literature on advice-taking~\cite{Y04,GM07,L17}.

\edit{
  Weight of advice quantifies the extent to which people update their beliefs (e.g., their own predictions made \emph{before} seeing a model's predictions) toward any advice they are given (e.g., the model's predictions). In the context of our first two experiments, each participant's weight of advice for each apartment is defined as $\frac{\lvert{u^{(2)}_a - u^{(1)}_a}\rvert}{\lvert{m - u^{(1)}_a}\rvert}$, where $m$ is the model's prediction of the apartment's selling price , $u^{(1)}_a$ is the participant's initial prediction of the apartment's selling price before seeing $m$, and $u^{(2)}_a$ is the participant's final prediction of the apartment's selling price after seeing $m$. Weight of advice is equal to 1 if the participant's final prediction matches the model's prediction and equal to 0.5 if the participant averages their initial prediction and the model's prediction.
}

\edit{ To understand the benefits of weight of advice, consider
  the scenario in which a participant's final prediction $u^{(2)}_a$
  is close to the model's prediction $m$. There are two reasons
  why this might happen.  On the one hand, it could be the case that
  the participant's initial prediction $u^{(1)}_a$ was far from $m$ and they made a significant update to their initial prediction
  after seeing the model's prediction. On the other hand, it could be the case
  that the participant's initial prediction $u^{(1)}_a$ was already close to $m$,
  so they did not update their prediction at all after
  seeing the model's prediction. The absolute difference between the model's
  prediction $m$ and the participant's final prediction $u^{(2)}_a$ does not
  distinguish between these two cases. In contrast, weight of advice
  does---i.e., it will be high in the first case and low in the second.
}

\edit{
We additionally used our third experiment to check whether participants' behavior would be different if they were told that the predictions were made by a ``human expert'' instead of a model. Previous studies have examined this question from different perspectives with differing results~\citep{OG+09,DSM15,dzindolet2002,dijkstra1998,dijkstra1999}.
Most closely related to our experiment the work of Logg~\citep{L17,logg2019algorithm}, which showed that when people have no information about the quality of the predictions they are shown, they follow the predictions that appear to come from a computational system more closely than those that appear to come from a person.
We were curious to see whether this would also be the the case when people were given a chance to assess the quality of the predictions before deciding how closely to follow them.
}

\edit{The details of our hypotheses for this experiment are provided in Appendix~\ref{appndx:experiment_3}.}

\subsection{Experimental design}

\edit{
  For this experiment, we returned to the original New York City prices and used the same four primary experimental conditions as in the first two experiments. However, we also added a new condition, in which participants saw exactly the same information as in the condition involving the black-box, eight feature model, but with the model labeled as ``Human Expert'' instead of ``Model.'' We did not include a baseline condition because the most natural baseline would have been to simply ask each participant for their own prediction of each apartment's selling price, which was already the first half of this experiment's testing phase, as described below.
}

\edit{
As before, we ran the experiment on Amazon Mechanical Turk using psiTurk. We excluded Turkers who had participated in our first two experiments, and recruited 1,000 new participants, all of whom satisfied the screening criteria from our first two experiments. However, when analyzing the data, we excluded the responses from one participant who reported technical difficulties with the experiment. We randomly assigned participants to the experimental conditions (\abr{clear-2}, $N=202$; \abr{clear-8}, $N=200$; \abr{bb-2}, $N=202$; \abr{bb-8}, $N=198$; and \abr{expert}, $N=197$). For this experiment, each participant received a flat payment of $\$1.50$.
}

\edit{
  We asked participants to predict the selling prices of the same apartments that we used in our first two experiments. However, we slightly modified the testing phase so that we could calculate weight of advice. In particular, each participant was asked for two predictions of each apartment's selling price: an initial prediction before being shown the model's prediction and a final prediction after being shown the model's prediction. To keep the length of the experiment reasonable, we did not ask participants to guess what the model would predict for each apartment's selling price.
}

\edit{
  We also designed the experiment so as to elicit each participant's initial predictions for all twelve apartments before showing them the model. This is because we ran a small experiment in which participants were first shown an apartment's configuration (i.e., feature values) and asked for their prediction of its selling price. They were then shown the model's prediction---and the model itself, whose internals were either clear or black box---and asked to update their prediction before moving on to the next apartment. We found that participants assigned to the condition involving the clear, two-feature model made initial predictions that were closer to the model's predictions---even though they had not seen the model's prediction when making their initial prediction---compared to participants assigned to the other primary experimental conditions ($t\left(239\right)=-3.42$, $p < 0.001$).} 
  \edit{We suspect that this is because the clear, two-feature model was easiest for participants to simulate. As a result, participants may have more easily internalized the model's coefficients when making their final prediction for an apartment and then used them to make their initial predictions for subsequent apartments. Although this kind of behavior is often beneficial, here it posed a threat to the validity of our experiment: for us to be able to compare participants' weight of advice between different experimental conditions, a participant's initial predictions should not be influenced by the condition to which they were assigned.}


  \begin{figure*}
  \captionsetup[subfigure]{aboveskip=-1pt,belowskip=-1pt}
\centering
	\begin{framed}
        \begin{subfigure}[t!]{\textwidth}
                \centering
                \imagebox{15mm}{\includegraphics[width=\linewidth]{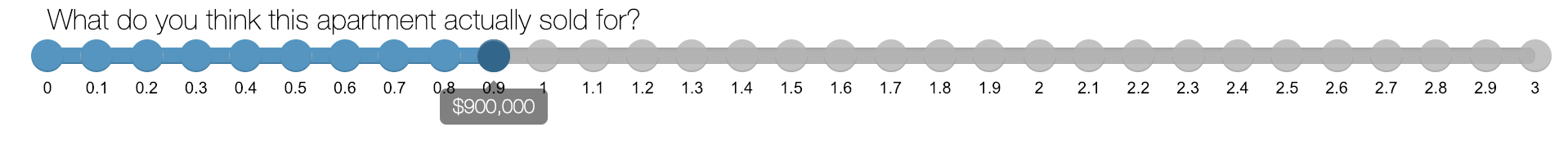}}
                \caption{First half: Participants were asked for their initial prediction of the apartment's selling price.}
                \label{fig:woa_step1}
                \Description[]{Image of a input slider with values from 0 to 3 million dollars under the question ``What do you think this apartment actually sold for?''}
        \end{subfigure}
	\end{framed}

        \begin{framed}
        \begin{subfigure}[t!]{\textwidth}
                \centering
                \imagebox{55mm}{\includegraphics[width=\linewidth]{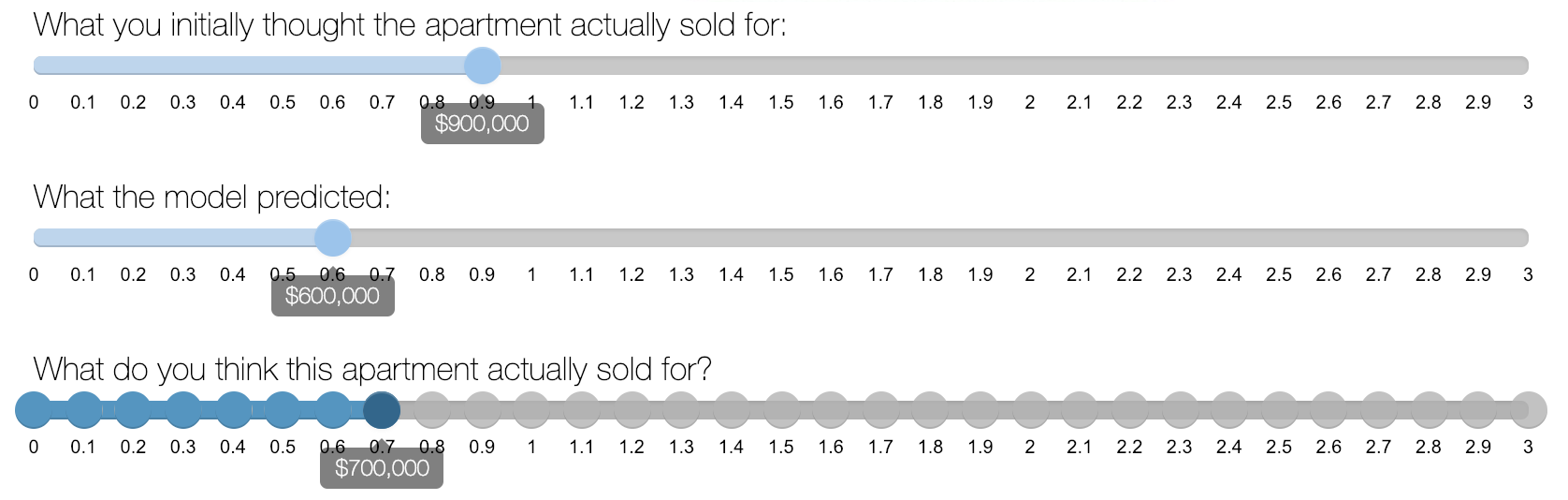}}
                \caption{Second half: Participants were shown the model's prediction and asked to update their prediction.}
                \label{fig:woa_step2}
                \Description[]{Image with three sliders ranging from 0 to 3 million (dollars). The top display slider reads ``What you initially thought the apartment sold for''. The middle display slider reads ``What the model predicted'' and the bottom input slider asks ``What do you think this apartment actually sold for?''}
        \end{subfigure}
        \end{framed}
        \caption{Part of the testing phase from our third experiment.}
  \end{figure*}

\edit{
  As in our first two experiments, participants were first shown detailed instructions (which, this time, intentionally did not include any information about the model or ``human expert''), before proceeding with the experiment, which consisted of two phases. In the (short) training phase, participants were shown three apartments in a random order. For each one, they were asked for their prediction of the apartment's selling price and shown the actual selling price. The testing phase consisted of two halves. In the first half, participants were shown another twelve apartments. The order of all twelve apartments was randomized. Participants were asked for their initial prediction of each apartment's selling price (see Figure~\ref{fig:woa_step1}). In the second half, participants were first introduced to the model or ``human expert'' before revisiting the twelve apartments (see Figure~\ref{fig:woa_step2}). The order of the first ten apartments was randomized, while the remaining two (apartments 11 and 12) always appeared last, as in the first two experiments. For each apartment, participants were first reminded of their initial prediction, then shown the model or expert's prediction, and only then asked to make their final prediction of the apartment's selling price. To simplify the experiment, we did not ask participants to state their confidence in either their or the model's predictions.
}

\subsection{Findings}

\medit{We briefly summarize our findings here and provide full details in Appendix~\ref{appndx:experiment_3}.} \medit{This e}xperiment confirmed \medit{our} findings from the first two experiments \medit{about the extent to which participants followed the predictions of the clear, two-feature model when it was beneficial for them to do so compared to the predictions of
the black-box, eight-feature model.} We again found \medit{no significant difference in how closely participants followed the predictions of the clear, two-feature model compared to the predictions of the black-box, eight-feature model---}this time \medit{measur}ed \medit{in terms of} weight of advice, as well as \medit{in terms} of the \medit{absolute
difference between} the \medit{model's} prediction \medit{and} the \medit{participant's final}
prediction for each of the apartments with typical
configurations.

We also found that participants \medit{followed} the predictions of \medit{the ``}human expert\medit{'' no more clos}ely \medit{than they} follow\medit{ed} the predictions of the black-box models. \medit{We suspect that the difference between this finding and those of Logg~\citep{L17,logg2019algorithm} is due to participants' increasing experience with the model or ``human expert'' over the course of our experiment.}

Finally, in contrast to the findings from \medit{our} first two experiments, we did \emph{not} find that participants \medit{assigned to} the conditions \medit{involving clear models} were less able to \medit{detect and} correct \medit{for the model's overly high} predictions for \medit{either apartment 11 or apartment 12}. This \medit{last finding} motivated our final experiment, which \medit{we describe in the next section.}


\section{Experiment 4: Outlier Focus and Detection of Mistakes}
\label{sec:exp4}
\medit{C}ontrary to \medit{our intuition when designing the first two experiments}, participants \medit{who were shown a} clear \medit{model} in those experiments were less \medit{able} to \medit{detect and correct for} the model\medit{'s sizable mistakes on} apartment\medit{s with unusual configurations} compared \medit{to participants assigned to} conditions \medit{involving black-box models (see} Figures~\ref{fig:exp1_q12_pred} and~\ref{fig:exp2_q12_pred}).
In our third experiment\medit{, in seeming contradiction, we found} no \medit{such} difference \medit{between the} conditions \medit{involving clear} model\medit{s and} the \medit{conditions involving black-box models}. In this section, we propose a \medit{possible} explanation for these findings and \medit{then} support it with a \medit{final} experiment. The explanation rests on two \medit{reasons}, \medit{which we outline below}.

\medit{First, in all three experiments, participants who were shown a clear model may have been overwhelmed by the amount of detail in front of them---i.e., they may have experienced information overload\footnote{We emphasize that we are referring to visual information overload that affects attention to items on a display~\citep{chun2000contextual}, not cognitive load in
working memory, which has also been shown to be related to interpretability~\citep{abdul2020cogam,lage2019human}.}~\citep{ackoff1967,jacoby1984,keller1987}---causing them to be less likely to notice the unusual apartment configurations when making their own predictions. We conjecture that this effect may have been less pronounced in our third experiment, though, because participants were asked for their initial predictions for all twelve apartments' selling prices before being introduced to the model. In turn, this may have meant that they paid greater attention to each each apartment's configuration---unusual or not.}

\medit{Second, in all three experiments, participants may have anchored on the prediction visible to them when making their own final prediction of an apartment's selling price~\citep{TK74,dietvorst2018overcoming}. However, the possible anchor values differed between the experiments: In the first two experiments, participants made their final prediction of each apartment's selling price while seeing their \emph{simulation of the model's prediction} (see Figure~\ref{fig:exp1_step3}). In contrast, in the third experiment, participants made their final prediction of each apartment's selling price while seeing their \emph{own initial prediction of the apartment's selling price} (see Figure~\ref{fig:woa_step2}).}

\medit{Furthermore, in the first two experiments, the anchor values differed between the experimental conditions because they were influenced by the model involved. Participants assigned to the condition involving the clear, two-feature model could better simulate the model compared to participants assigned to the other experimental conditions (see Figures~\ref{fig:exp1_simulation_error} and~\ref{fig:exp2_simulation_error}). However, if the model has overpriced an apartment, then better simulating it might cause participants to anchor on a selling price that is too high. On top of that, because clear models reveal more information, participants may have been even less likely to notice the unusual apartment configurations due to information overload. In contrast, participants assigned to the conditions involving black-box models were not able to simulate the model so well and, perhaps undistracted by what was in front of them, may have been} more likely to notice the unusual apartment configurations. Interestingly, participants \medit{assigned to} the conditions involving black-box models apparently (incorrectly) assumed \medit{that} the model would take the unusual \medit{apartment} configurations into account and \medit{therefore made lower guesses for} the model\medit{'s predictions. In other words, in the first two experiments}, participants \medit{assigned to} the \medit{conditions involving black-box models} could have had two \medit{things} working in their favor: they were \medit{less likely to be} overwhelmed by the \medit{amount of detail in front of them and they may have anchored on their lower guesses for the model's predictions.}

\begin{figure}[t]
  \centering
  \includegraphics[width=0.9\linewidth]{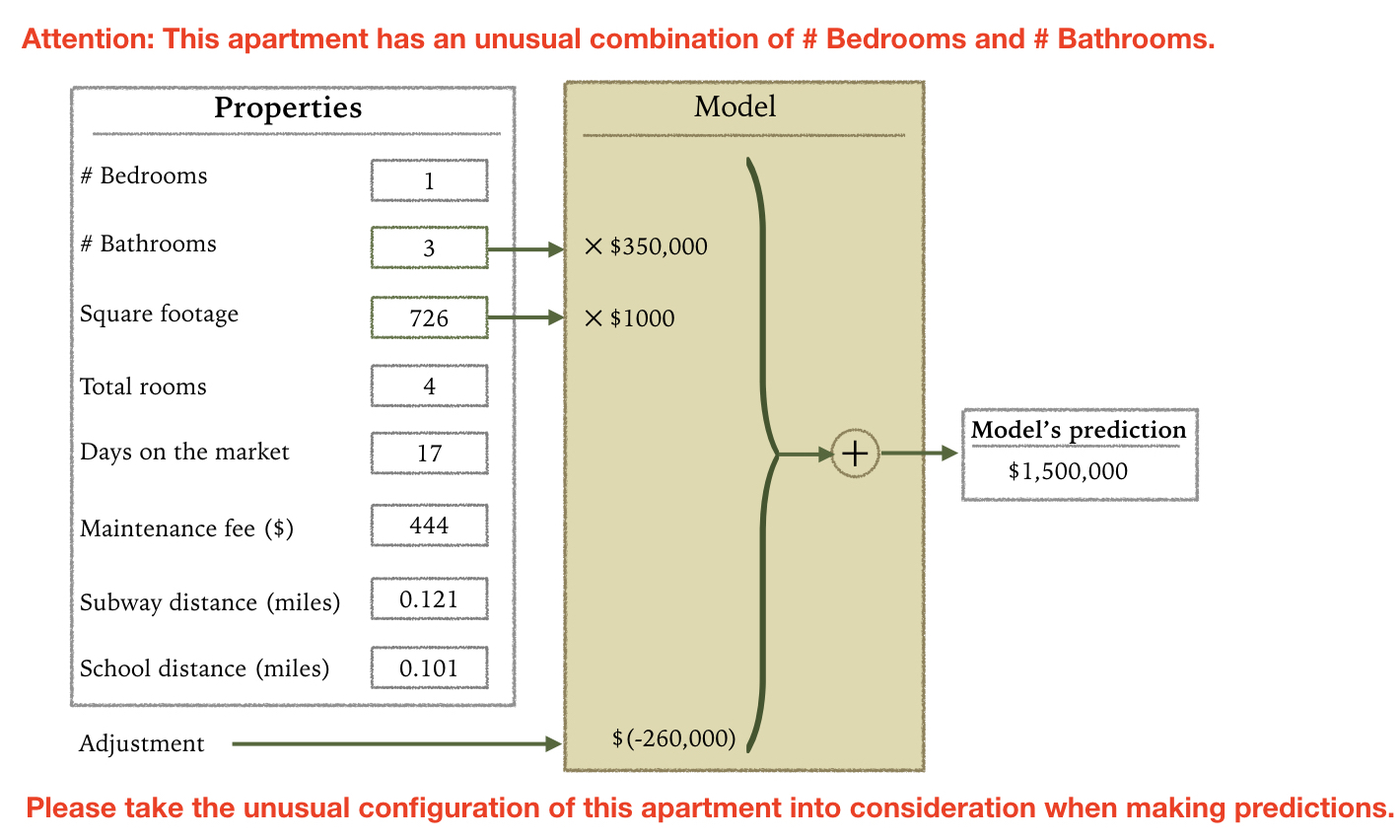}
\caption{Apartment 6 in the conditions involving an outlier focus message in our fourth experiment.}
\label{fig:exp4_apt6}
\Description[]{Image showing how apartment 6 was presented to the participants in the condition involving an outlier focus message. Above the image is the message (in red) ``Attention: This apartment has an unusual combination of # Bedrooms and # Bathrooms''. Beneath the message (in red) is stated ``Please take the unusual configuration of this apartment into consideration when making predictions.''}
\end{figure}

\medit{We designed our fourth experiment to test this possible explanation. As we describe below, this experiment removed the potential for anchoring and measured the effect of an ``outlier focus'' message highlighting the apartments with unusual configurations as possible outliers (see Figure~\ref{fig:exp4_apt6}). In our previous experiments, the number of features did not appear to have a strong effect on participants' abilities to detect and correct for the model's sizable mistakes, so we used only the two-feature linear regression model in this experiment.}

Before running th\medit{e} experiment, we posited and pre-registered three hypotheses\medit{, stated informally below}:\edit{\footnote{Pre-registered hypotheses for this experiment are available at \url{https://aspredicted.org/5xy8y.pdf}.}}


H11. \textbf{Outlier focus.} Participants \medit{that see an} outlier focus \medit{message and participants that don't see an outlier focus message will be differently able to detect and correct for the model's sizable mistakes on unusual data points.}

H12. \medit{\textbf{Transparency (clear vs. black box) and no outlier focus.} When they are not shown an outlier focus message, participants who are shown a clear model and participants who are shown a black-box model will be differently able to detect and correct for the model's sizable mistakes on unusual data points.}

H13. \medit{\textbf{Transparency (clear vs. black box) and outlier focus.} When they are shown an outlier focus message, participants who are shown a clear model and participants who are shown a black-box model will be differently able to detect and correct for the model's sizable mistakes on unusual data points.}

\subsection{Experimental design}
\label{subsec:exp4_design}
Similar to the \medit{design of our} first experiment, we asked \medit{participants} to predict \medit{the selling prices of} apartment\medit{s in New York City} with the help of a \medit{machine learning} model. We \medit{used} a $2 \times 2$ design:

\begin{itemize}[nosep]
\item Participants were randomly assigned to see \medit{either a clear} model \medit{(i.e., a  linear regression model with visible coefficients) or a black-box model.}
\item Participants were randomly assigned to \medit{either} see an \medit{outlier focus} message \medit{highlight}ing the \medit{apartments with} unusual configurations \medit{as possible outliers} or \medit{to} not \medit{see such a message}.
\end{itemize}

We again ran the experiment on Amazon Mechanical Turk \medit{using psiTurk}. We excluded \medit{Turkers} who had participated in our first three experiments, and recruited 800 new participants\medit{,} all of whom satisfied the \medit{screening} criteria from our first three experiments.
\medit{W}e randomly assigned \medit{participants} to the \medit{experimental} conditions (\abr{clear-focus}, $N=202$; \abr{clear-no-focus}, $N=195$; \abr{bb-focus}, $N=201$; and \abr{bb-no-focus}, $N=202$). Each participant received a flat payment of $\$1.00$.

\medit{To keep} the experiment \medit{short, we} limit\medit{ed} the training phase and the first \medit{portion} of the testing phase to only five of the original ten apartments \medit{previously used} in each phase.
We \medit{us}ed three apartments \medit{for} the \medit{seco}nd \medit{portion} of the testing phase: two synthetically generated apartments with unusual configurations, with an apartment with a \medit{typical} configuration (one\medit{ }bedroom, one\medit{ }bathroom, 788 square feet) in between.
The first synthetically generated apartment \medit{(``apartment 6'')} was apartment 12 from our previous experiments (\medit{a} one-bedroom, three-bathroom, 726\medit{-}square\medit{-foot }apartment\medit{).}
The second synthetically generated apartment \medit{(``apartment 8'')} had an even more unusual configuration (one bedroom, three bathrooms, 350 square feet\medit{) and, like apartment 6, was overpriced by} the model. The order of the two synthetically generated apartments was randomized, while the apartment with the \medit{typical} configuration (\medit{``}apartment 7\medit{''}) was always shown in the middle.

\medit{The} training phase \medit{was the same} as in the first two experiments \medit{(except with fewer apartments). In} the testing phase, participants were \medit{shown eight apartments, described above. For each apartment, participants were} first shown the model's prediction and asked to \medit{st}ate \medit{their} confiden\medit{ce in} that \medit{prediction}. They were \medit{then} asked \medit{for} their own prediction of the apartment's \medit{selling} price and to \medit{state their confidence} in this prediction. To remove \medit{the} potential \medit{for} anchoring, participants \medit{were not asked} to \medit{guess what} the model \medit{would predict for each apartment}.

\begin{figure*}[t]
  \captionsetup[subfigure]{aboveskip=-2pt,belowskip=-2pt}
  \centering
     \begin{subfigure}[b]{0.5\linewidth}
                \includegraphics[width=\linewidth]{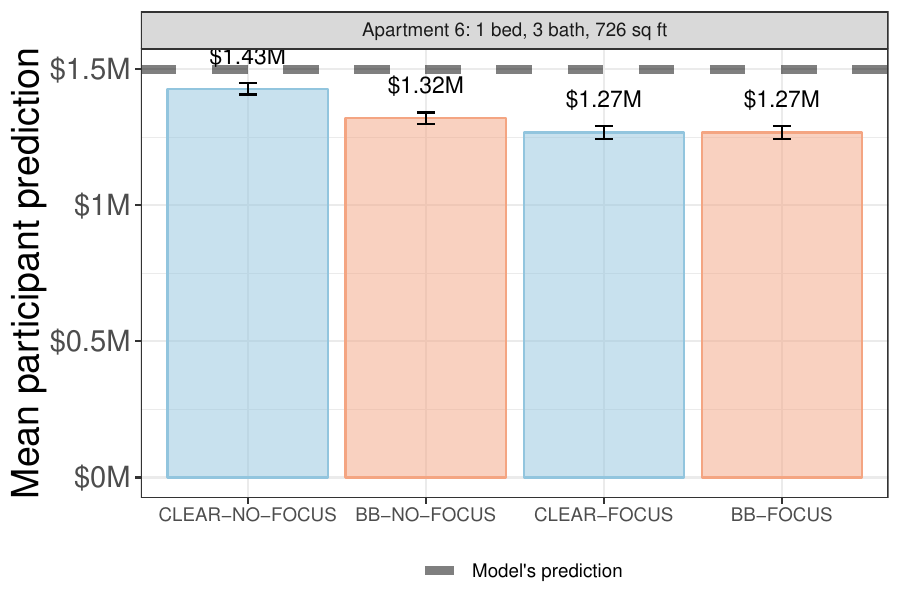}
                \caption{}
        \end{subfigure}%
        \begin{subfigure}[b]{0.5\linewidth}
		\includegraphics[width=\linewidth]{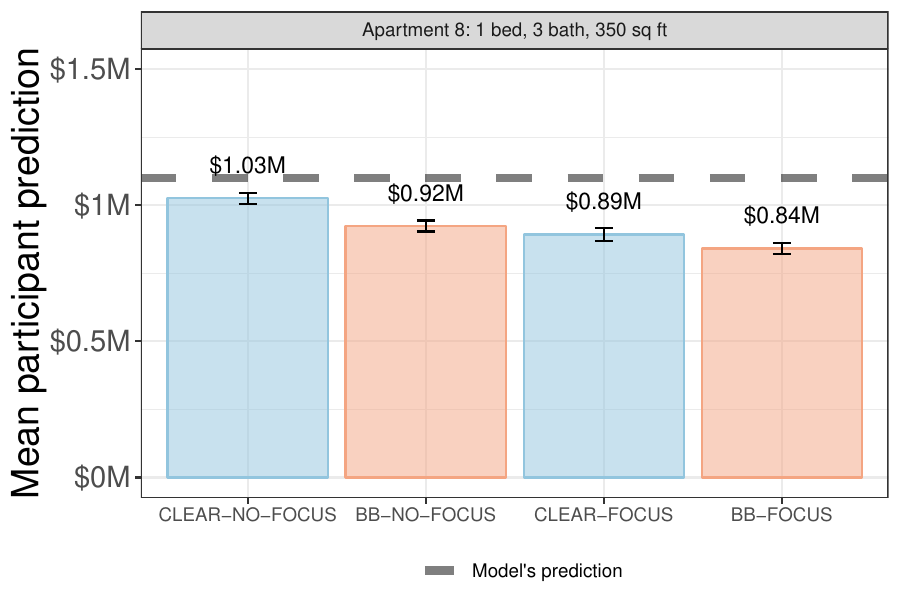}
                \caption{}
        \end{subfigure}%
		\caption{Results from our fourth experiment: participants' mean predictions of the selling prices for the apartments with unusual configurations: (a) apartment 6 and (b) apartment 8. Horizontal lines indicate the model's predictions and error bars indicate one standard error.}
              \label{fig:attention_check}
        \Description[]{Bar chart showing participants' mean predictions of the selling prices for the apartments with unusual configurations for all four experimental conditions. The  clear-no-focus condition is highest and closest to the model's prediction. The two focus conditions are lower than the black-box-no-focus condition, which is lower than the clear-no-focus condition. The same pattern holds for both apartment 6 and apartment 8.}
\end{figure*}

\subsection{Findings}
\label{subsec:exp4_results}
Figure~\ref{fig:attention_check} shows \medit{participants'} mean predictions of the \medit{selling} prices \medit{for the} apartment\medit{s with unusual configurations (i.e,} apartment \medit{6 and apartment 8)}. To test our hypotheses, we \medit{defined each} participant's \medit{deviation from the model's} prediction of each apartment's selling price \medit{to be} $u_a - m$, \medit{where $m$ is} the model's prediction \medit{and $u_a$ is the participant's prediction of the apartment's selling price. We used signed difference (rather than absolute difference, as in our first two experiments) because the goal of this experiment was to study participants' abilities to detect and correct for the model's mistakes. Using signed difference enabled us to more easily tell whether a participant's deviation from the model's prediction was in the right direction.}

H11. \textbf{\medit{O}utlier focus.} \medit{We found that p}articipants in \medit{conditions involving an} outlier focus \medit{message deviated} from the model's prediction\medit{s}, on average, more \medit{compared to} participants \medit{who did not see an outlier focus message} for both apartment 6 ($t\left(791\right)=-4.72$, $p < 0.001$) and apartment 8
($t\left(795\right)=-5.00$, $p < 0.001$). \medit{In other words, showing participants} an outlier focus message \medit{better enabled them to detect and correct for the model's sizable mistakes on the apartments with unusual configurations}.

H12. \textbf{\medit{T}ransparency \medit{(clear vs. black box) and no outlier focus}.} \medit{Participants assigned to} conditions \medit{involving} the clear \medit{model and no outlier focus message deviated} from the model's prediction\medit{s}, on average, less \medit{compared to participants assigned to conditions involving} the black-box \medit{model and no outlier focus message} for both apartment 6 ($t\left(393\right)=-3.65$, $p < 0.001$) and apartment 8 ($t\left(395\right)=-3.51$, $p < 0.001$). In other words, in \medit{line with} the \medit{findings from our first two experiments, without} an outlier focus message, participants \medit{who were shown} the clear \medit{model were less able to detect and correct for the model's sizable mistakes on the apartments with unusual configurations, compared to participants who were shown a black-box model.}

H13. \textbf{\medit{T}ransparency \medit{(clear vs. black box) and outlier focus}.} \medit{We found no significant
difference in participants' deviations from the model's predictions
between the condition involving the clear model and an outlier
focus message and the condition involving the black-box model and an
outlier focus message ($t\left(401\right) = -0.004$, $p = 0.996$
for apartment 6 and $t\left(394\right) = -1.64$, $p = 0.101$ for
apartment 8). In other words, with an outlier focus message,
participants who were shown the clear model were similarly able to
detect and correct for the model's sizable mistakes, compared to
participants who were shown a black-box model. This finding suggests
that an outlier focus message helps participants pay attention to
information that they might otherwise miss due to information overload.}

\medit{Taken together, these findings support our explanation for the difference between the findings from our first two experiments and the findings from our third. They also highlight an unintended disadvantage of using clear models---and offer a simple way to mitigate it.}

\medit{In light of this, we returned to the data from our first two experiments and conducted some additional post-hoc analyses. First, we analyzed participants' prediction errors. In our first experiment, although a one-way ANOVA did not reveal a significant difference in participants' prediction errors between the four primary experimental conditions (see Section~\ref{sec:exp1}), a visual inspection of our results (see Figure~\ref{fig:exp1_prediction_error}) indicates that participants assigned to the condition involving the clear, eight-feature model had, on average, higher prediction errors than participants assigned to the other primary experimental conditions. Indeed, this difference is statistically significant ($t\left(994\right) = 2.68$, $p = 0.007$). Of course, we note that with large sample sizes, statistical significance might not mean practical significance~\citep{vicente2000earth, meehl1990summaries}. Indeed, this seems to be the case with participants' prediction errors. For example, in Figure~\ref{fig:exp1_prediction_error}, the maximum pairwise difference in prediction error between the four primary experimental conditions is quite small---only about \$16,000 or roughly 1\% of the average selling price, which was \$1.2 million. In contrast, in Figure~\ref{fig:exp1_simulation_error}, the maximum pairwise difference in simulation error between the four primary experimental conditions is more substantial at \$129,000.}

\medit{Analyzing the data from our second experiment revealed a similar pattern. Here, a one-way ANOVA did reveal a small but significant difference in participants' prediction errors between the four primary experimental conditions (see Section~\ref{sec:exp2}). Again, a visual inspection of our results (see Figure~\ref{fig:exp2_prediction_error}) indicates that participants assigned to the condition involving the clear, eight-feature model had, on average, higher prediction errors than participants assigned to the other primary experimental conditions. Similar to our first experiment, this difference was significant ($t\left(594\right) = 4.78$, $p < 0.001$). Though again, we note that although these differences are statistically significant, they are not very large.
}

\medit{These findings motivated us to also investigate whether there were other differences between the condition involving the clear, eight-feature model and the other primary experimental conditions. For both the first and second experiment, we found that participants who were assigned to the condition involving the clear, eight-feature model were less good at simulating the model's predictions compared to participants assigned to the other primary experimental conditions ($t\left(994\right) = 7.96$, $p < 0.001$ for the first experiment, $t\left(594\right) = 7.23$, $p < 0.001$ for the second experiment; see Figures~\ref{fig:exp1_simulation_error} and~\ref{fig:exp2_simulation_error}) and that they were least likely to follow the model's predictions when it was beneficial for them to do so ($t\left(994\right) = 2.37$, $p = 0.018$ for the first experiment, $t\left(594\right) = 2.49$, $p = 0.012$ for the second experiment; see Figures~\ref{fig:exp1_dev_from_model} and~\ref{fig:exp2_dev_from_model}).}

\medit{To summarize, the findings from these additional post-hoc analyses of the data from our first two experiments lend even more support to our explanation for the difference between the findings from our first two experiments and the findings from our third.}


\section{Limitations}

\edit{One limitation of our work is that our experiments focused on one type of stakeholder (laypeople) using one type of model (linear regression) in one domain (real estate valuation). Future extensions to other types of stakeholders (e.g., data scientists, domain experts), other tasks (e.g., classification), other types of models (e.g, decision trees, rule lists, deep neural networks), and other domains (e.g., medical diagnosis, credit risk assessment, judicial sentencing and bail, hiring) may yield different findings.}

\edit{In our first three experiments, we constrained the two-feature model and the eight-feature to make the same predictions. Although there are some domains where this is possible~\citep{jung2020simple}, there are of course others---such as computer vision and natural language processing---where more complex, deep models tend to outperform simpler ones.
We did not experiment with such models because it would have created a confound, meaning that we would not have known whether any differences we observed were due to the presentation of the model, the model fidelity, or the very large number of features that complex, deep models typically use.
Although our experiments did not involve complex, deep models, our main findings still have important implications for these domains: absent other reasons for using clear models, scientific evidence about what aids decision making the most should carry more weight than common intuition about interpretability.
We also emphasize that, in our experiments, the conditions involving black-box models were designed to capture how people engage with models that could have arbitrarily complex internal structures, including, for instance, deep neural networks. Indeed, although readers of this paper know that we used linear regression models, participants in our experiments had no reason to believe that this was the case.
}

\edit{Even though our experiments were carefully designed and tightly
controlled, we cannot rule out the possibility that other aspects of the models influenced our findings. For example, participants might have found the particular features used in
the two-feature model (i.e., bathrooms and square feet) to be less
intuitive than other possible combinations of two features (e.g.,
bedrooms and bathrooms) or even three features (e.g., bedrooms,
bathrooms, and square feet). Also, participants who were shown the
two-feature model had access to more information than the model---a
scenario known in the decision-making literature as the ``broken leg
problem''}~\cite[p.~151]{dawes1989clinical}\edit{. For this reason, they may
have thought that the model was not relying on information that it
should have. Perhaps if they were told that using the remaining six
features would not improve the model's accuracy, they would have
viewed the two-feature model differently. Conversely, though, it could have
been the case that aspects of the eight-feature model led participants
to question it. For instance, the negative coefficient for total rooms
(which accounted for correlations between number of bedrooms and
number of bathrooms) might have been confusing or mistakenly viewed as
wrong, leading participants to follow the model's predictions less
closely than they would have done otherwise.}

\edit{Lastly, our experiments were run without process measures as dependent variables, which limited our ability to reflect on the cognitive and sensemaking processes that might have been at play. As one example, while we measure participants' ability to detect and correct for the model's sizable mistakes in terms of their deviation from the model on apartments with unusual configurations, we are unable to directly infer from these results whether participants understood why the model made these mistakes. Qualitative experiments (involving interviews, think aloud protocols, process-tracing measures, etc.), targeted at understanding \emph{why} people behave in the ways they do, may be useful for investigating cognitive and sensemaking aspects of interpretability. On top of that, our experiments were short and one shot. Deeper insight into sensemaking could be gained not only by collecting process measures but by doing so longitudinally.}

\section{Discussion and Conclusion}

\medit{Our experiments yielded some unexpected findings. First, we did not find a significant improvement in the extent to which participants followed the predictions of a clear model with few features compared to the predictions of a black-box model with more features. We also found that participants would have had lower prediction errors had they simply followed the model's predictions.}

\medit{Furthermore, we found that using a clear model hampered participants' abilities to detect when the model had made a sizable mistake, seemingly due to information overload caused by the amount of detail in front of them. When we investigated an outlier focus message, intended to counter information overload, we found that this behavior disappeared. Several findings from our post-hoc analyses are also consistent with the idea that too much information can be detrimental. In our first two experiments, in the condition in which participants were shown the \emph{most} information---i.e., the condition involving the clear, eight-feature model---participants were \emph{worst} at simulating the model's predictions, followed the model's predictions \emph{less}, and made \emph{less accurate} predictions of the apartments' selling prices compared to participants assigned to the other primary experimental conditions.}

\medit{These findings suggest new ways to present models to people.
When technically possible, it may be helpful to alert people when the data point in front of them may be an outlier. This could be achieved by training an auxiliary model to detect such data points.
In addition, it may be prudent to ask people for their own predictions before seeing the model's predictions or even the model itself.
Doing so could encourage people to inspect each data point carefully, making them more likely to notice any unusual feature values.} \edit{Indeed, this idea is supported by recent research, which found that eliciting predictions and presenting feedback is beneficial for people's memory and comprehension of data points~\citep{kim2017explaining}}. \medit{Lastly, despite the potential benefits of clear models, it may be detrimental to expose model internals by default, as doing so might cause people to experience information overload.
Instead, model internals could be hidden until the person using the model requests to see them.
Testing these suggestions empirically would be a natural direction for future research.}

\medit{We emphasize that none of this is to say that the number of features or the transparency of the model should be ignored. Instead, our findings underscore the point that there are many possible goals when developing interpretable models, and that testing, not intuition, should be used to assess whether these goals have been met~\citep{wang2019designing, liao2020questioning}.}

\medit{Although we found that two factors commonly thought to make machine learning models more interpretable often have negligible effects on people's behavior and, in some cases, even have detrimental effects}, \medit{there is still a long list of reasons why clear models with few features may be desirable.
First, in some domains, transparency may play an important role in people's willingness to use a model on ethical grounds.
For instance, if a model is used to aid judicial decision making, policy makers may demand transparency so as to be assured that the model does not rely on disallowed information, like race, or proxies for disallowed information.
Second, access to model internals permits types of debugging or analyses that would otherwise be difficult. In fact, we leveraged this aspect of our linear regression models to generate some of the unusual apartment configurations used in our experiments, since we could easily see that the models would place an unreasonably high value on additional bathrooms when other feature values were held constant.
Third, in scenarios where it is desirable to have a model that is easy to simulate, our findings suggest people can better simulate the predictions of clear models with few features.
Fourth, although we did not investigate the field adoption of machine learning models, it might be the case that people are more likely to use simpler models than more complex ones because they find them more appealing~\citep{jung2020simple}.} \medit{Given that we did not find a large difference in participants' prediction errors between the primary experimental condition in our first two experiments, if people are more willing to use simpler models, there could be substantial benefits in terms of accuracy.}

\medit{Given the widespread and increasing use of machine learning models, it is likely that people will make more and more decisions in collaboration with models. As this happens, it is also likely that there will be an increased demand for models that are interpretable. We hope that our work reinforces the importance of testing over intuition when developing interpretable models---i.e., what is or is not interpretable must be defined by people's behavior.}

\bibliographystyle{ACM-Reference-Format}
\bibliography{references}
\clearpage


\section*{Appendices}
\begin{appendices}
  \section{Scenarios Where Users Have Access to More Information Than Models}
  \label{appndx:decision_aid_table}
  \begin{longtable}[]{@{}llll@{}}
\toprule
\begin{minipage}[b]{0.20\columnwidth}\raggedright
\textbf{~\\
Domain}\strut
\end{minipage} & \begin{minipage}[b]{0.25\columnwidth}\raggedright
\textbf{Information the model uses}\strut
\end{minipage} & \begin{minipage}[b]{0.35\columnwidth}\raggedright
\textbf{Side information a user has\\ that the model does
not}\strut
\end{minipage} & \begin{minipage}[b]{0.15\columnwidth}\raggedright
\textbf{Citation}\strut
\end{minipage}\tabularnewline
\midrule
\endhead
\begin{minipage}[t]{0.20\columnwidth}\raggedright
Malignancy risk in mammography\strut
\end{minipage} & \begin{minipage}[t]{0.25\columnwidth}\raggedright
Age tumor density, 5 binary variables describing tumor shape\strut
\end{minipage} & \begin{minipage}[t]{0.35\columnwidth}\raggedright
Full mammogram image, full medical records, clinical interview (habits,
family history, etc), plus 10 binary variables not in interpretable
model\strut
\end{minipage} & \begin{minipage}[t]{0.15\columnwidth}\raggedright
\cite{wang2015}\strut
\end{minipage}\tabularnewline
\begin{minipage}[t]{0.20\columnwidth}\raggedright
Hospital readmission risk\strut
\end{minipage} & \begin{minipage}[t]{0.25\columnwidth}\raggedright
7 binary features (bed sores, mood problems)\strut
\end{minipage} & \begin{minipage}[t]{0.35\columnwidth}\raggedright
Full medical records, plus 23 other features not in interpretable model
(``chronic pain'', ``feels unsafe'', etc.)\strut
\end{minipage} & \begin{minipage}[t]{0.15\columnwidth}\raggedright
\cite{wang2015}\strut
\end{minipage}\tabularnewline
\begin{minipage}[t]{0.20\columnwidth}\raggedright
Wildfire risk\strut
\end{minipage} & \begin{minipage}[t]{0.25\columnwidth}\raggedright
29 binary features, mostly covering terrain type and temperature\strut
\end{minipage} & \begin{minipage}[t]{0.35\columnwidth}\raggedright
Experience of past fires. local knowledge: history, hazardous
industries, previous arson, etc. In addition, information not used by
model: Continuous values of 9 variables, Any value of 4 variables, 8
dichotomous variables\strut
\end{minipage} & \begin{minipage}[t]{0.15\columnwidth}\raggedright
\cite{amatulli2006}\strut
\end{minipage}\tabularnewline
\begin{minipage}[t]{0.20\columnwidth}\raggedright
Bail decisions\strut
\end{minipage} & \begin{minipage}[t]{0.25\columnwidth}\raggedright
Statistical information available to judges at time of inquiry except
disallowed ones\strut
\end{minipage} & \begin{minipage}[t]{0.35\columnwidth}\raggedright
Information that is not used by the model because it is not allowed:
Race, ethnicity, gender. Physical appearance of the defendant (e.g.,
``tattoos''), answers to questions, apparent remorse, etc..\strut
\end{minipage} & \begin{minipage}[t]{0.15\columnwidth}\raggedright
\cite{kleinberg2018human}\strut
\end{minipage}\tabularnewline
\begin{minipage}[t]{0.20\columnwidth}\raggedright
Pre-trial release decisions\strut
\end{minipage} & \begin{minipage}[t]{0.25\columnwidth}\raggedright
7 binary variables covering age and past failures to appear\strut
\end{minipage} & \begin{minipage}[t]{0.35\columnwidth}\raggedright
Physical appearance of the defendant, answers to questions, apparent
remorse, etc. Information not used by the interpretable model: 49
features describing the charges, 13 characteristics of the
defendant.\strut
\end{minipage} & \begin{minipage}[t]{0.15\columnwidth}\raggedright
\cite{jung2020simple}\strut
\end{minipage}\tabularnewline
\begin{minipage}[t]{0.20\columnwidth}\raggedright
Success of early-stage ventures\strut
\end{minipage} & \begin{minipage}[t]{0.25\columnwidth}\raggedright
21 trinary features of companies\strut
\end{minipage} & \begin{minipage}[t]{0.35\columnwidth}\raggedright
Industry knowledge and experience, subjective assessments that lead to
trinary scores, and 16 trinary features not used by the interpretable
model\strut
\end{minipage} & \begin{minipage}[t]{0.15\columnwidth}\raggedright
\cite{aastebro2006}\strut
\end{minipage}\tabularnewline
\begin{minipage}[t]{0.20\columnwidth}\raggedright
Housing price prediction\strut
\end{minipage} & \begin{minipage}[t]{0.25\columnwidth}\raggedright
3 features: number of rooms, \% lower-income citizens, student-teacher
ratio\strut
\end{minipage} & \begin{minipage}[t]{0.35\columnwidth}\raggedright
Physical walkthrough of property, neighborhood knowledge. Continuous
values of the 3 features used by the model, 10 continuous variables not
used by the model.\strut
\end{minipage} & \begin{minipage}[t]{0.15\columnwidth}\raggedright
\cite{kim2007}\strut
\end{minipage}\tabularnewline
\begin{minipage}[t]{0.20\columnwidth}\raggedright
Baseball player salary prediction\strut
\end{minipage} & \begin{minipage}[t]{0.25\columnwidth}\raggedright
3 variables: number of years in major leagues, career hits, hits in
previous year\strut
\end{minipage} & \begin{minipage}[t]{0.35\columnwidth}\raggedright
Personal experience watching players. 19 variables not included in the
interpretable model.\strut
\end{minipage} & \begin{minipage}[t]{0.15\columnwidth}\raggedright
\cite{kim2007}\strut
\end{minipage}\tabularnewline
\begin{minipage}[t]{0.20\columnwidth}\raggedright
Sleep apnea screening\strut
\end{minipage} & \begin{minipage}[t]{0.25\columnwidth}\raggedright
5 binary features\strut
\end{minipage} & \begin{minipage}[t]{0.25\columnwidth}\raggedright
Medical records, patient interview, 28 binary features not included in
the interpretable model.\strut
\end{minipage} & \begin{minipage}[t]{0.15\columnwidth}\raggedright
\cite{UR16}\strut
\end{minipage}\tabularnewline
\begin{minipage}[t]{0.20\columnwidth}\raggedright
Classification of high- and low-risk heart attack patients\strut
\end{minipage} & \begin{minipage}[t]{0.25\columnwidth}\raggedright
3 binary features\strut
\end{minipage} & \begin{minipage}[t]{0.35\columnwidth}\raggedright
Patient interview. Continuous measures of these 3 features. 16 features
collected at intake but not included in the interpretable model.\strut
\end{minipage} & \begin{minipage}[t]{0.15\columnwidth}\raggedright
\cite{gigerenzer1999}; \cite{breiman1984}\strut
\end{minipage}\tabularnewline
\begin{minipage}[t]{0.20\columnwidth}\raggedright
Prediction of non-viable pregnancies\strut
\end{minipage} & \begin{minipage}[t]{0.25\columnwidth}\raggedright
6 features, each cut into 2 to 5 bins: maternal
age, bleeding score, gestational age, gestational sac
diameter, yolk sac diameter, fetal heart beat.\strut
\end{minipage} & \begin{minipage}[t]{0.35\columnwidth}\raggedright
Continuous values on all variables. Any other information in medical
records, patient interview, etc.~\strut
\end{minipage} & \begin{minipage}[t]{0.15\columnwidth}\raggedright
\cite{van2012}\strut
\end{minipage}\tabularnewline
\bottomrule
\caption{Examples of decision aids whose users have access to more information than the models do.}
\label{tab:decision_aid_table}
\end{longtable}
\clearpage

  \section{Instructions from the \abr{clear-2} Condition in Experiment 1}
  \label{appndx:instructions_exp1}
  
The following instructions were shown to participants assigned to the \abr{clear-2} condition in our first experiment on Mechanical Turk. The instructions for other conditions and experiments were adapted from these instructions with minimal changes.
\setboolean{@twoside}{false}

\includepdf[pages=-, scale=0.7, pagecommand={}, frame=true]{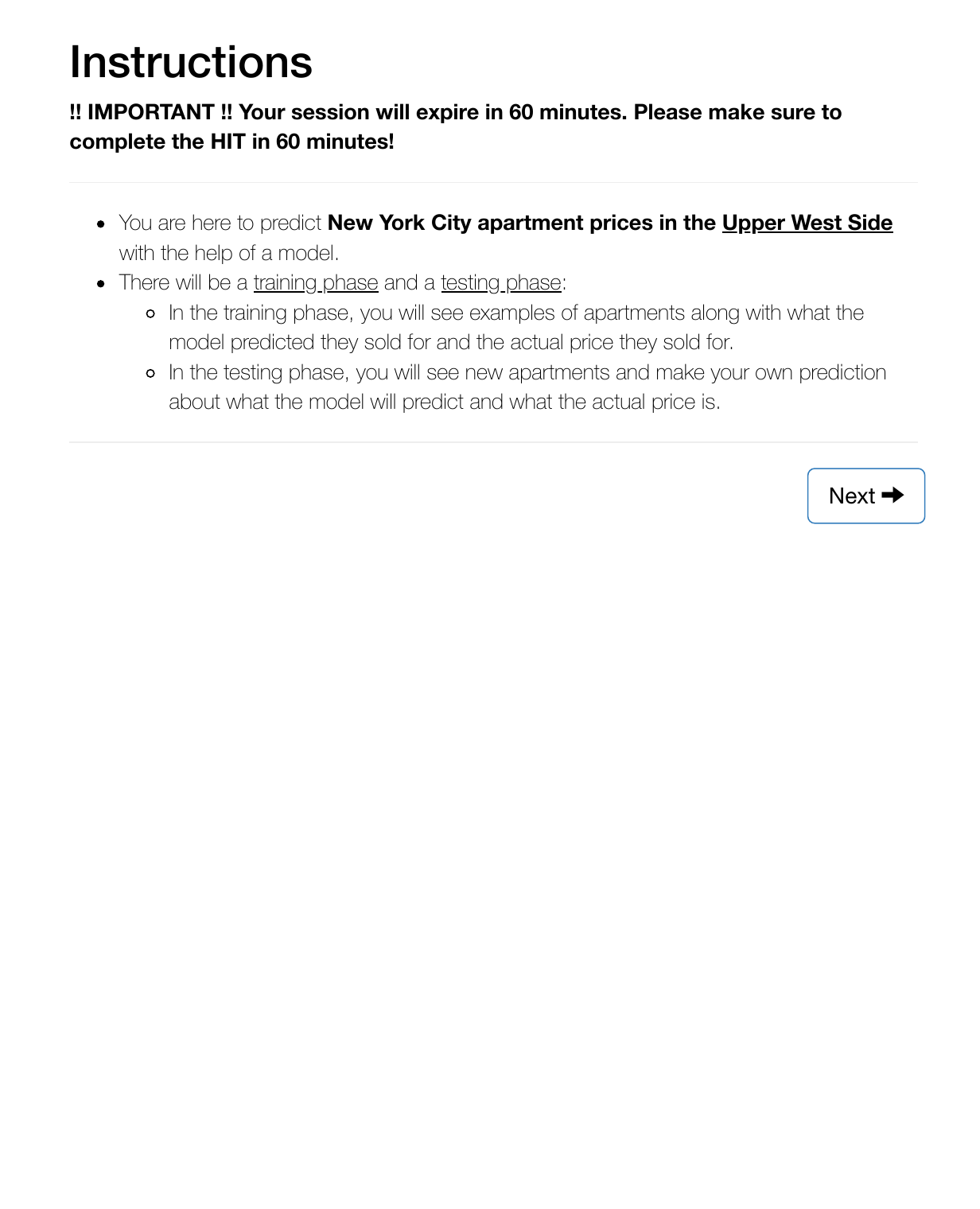}


\clearpage

  \section{Apartment Selection Details}
  \label{appndx:apartment_selection_details}
  We used the following procedure to construct a set of ten apartments which are representative in terms of the models' prediction errors ($m-a$). First we selected all apartments for which the rounded predictions of the two- and eight-feature models agreed. Then we randomly sampled 5,000 sets of ten such apartments, computed the errors the model made on each apartment, and sorted them within each set to obtain the largest error, second largest error, and so on. We then computed the average largest error across all 5,000 sets and rounded it to the nearest \$100K. We repeated this for the second through tenth largest errors. This resulted in the following ten average error values: -\$500K, -\$300K, -\$200K, -\$200K, -\$100K, \$0, \$0, \$100K, \$100K, \$300K. 

For each of these ten error values, we randomly selected two apartments (for which the difference between the rounded model prediction and the rounded actual price matched the error value) and we randomly assigned one to the training and one to the testing phase. To ensure participants would see a good variety of apartment configurations---defined as the combination of number of bedrooms and number of bathrooms---this process was repeated until neither the training nor the testing set contained more than three apartments with the same configuration. Tables~\ref{tab:apt_configs_training} and~\ref{tab:apt_configs_testing} show the configurations of apartments that were used during the training and testing phase of each of our experiments, respectively. Tables~\ref{tab:model_errors_train} and~\ref{tab:model_errors_test} show the predictions and errors of the two- and eight-feature models on each of the apartments.
\begin{table*}[h]
\begin{center}
\footnotesize
 \begin{tabular}{c c c c c c c c c}
 \toprule
 \pbox{10cm}{Apartment \\ ID} & \multicolumn{8}{c}{Apartment configurations} \\
 \hline
 & Bedrooms & Bathrooms & \pbox{2cm} {Square \\ footage} & \pbox{2cm}{Total \\ rooms} & \pbox{2cm}{Days on \\the market} & \pbox{2cm}{Maintenance \\ fee} &  \pbox{2cm}{Distance from \\ the subway \\ (miles)} & \pbox{2cm}{Distance from \\ a school \\ (miles)}\\ 
 \hline\hline
1&1&1&750&3&51&947&0.179&0.104 \\
\hline
2&1&1&550&3&90&409&0.122&0.278 \\
\hline
3&2&1&800&4&36&1160&0.218&0.365 \\
\hline
4&2&1&850&4&30&1720&0.105&0.153 \\
\hline
5&1&1&550&3&135&442&0.231&0.124 \\
\hline
6&0&1&540&2.5&72&332&0.064&0.271 \\
\hline
7&3&2&1990&6&213&1280&0.183&0.329 \\
\hline
8&2&1&1150&4&37&1500&0.129&0.351 \\
\hline
9&0&1&540&2.5&59&331&0.064&0.271 \\
\hline
10&2&2&1300&5&39&1110&0.110&0.250 \\
\hline
\end{tabular}
\caption{Configuration of the apartments used in experiments 1, 2, and 3 during the training phase. In Experiment 4, apartments 4, 5, 6, 8, and 10 were used.}
\label{tab:apt_configs_training}
\end{center}
\end{table*}

\begin{table*}[t!]
\begin{center}
 \begin{tabular}{c c c c | c c c}
 \toprule
 Apartment \\ ID & \multicolumn{3}{c}{two-feature model} & \multicolumn{3}{c}{eight-feature model} \\
 \hline
 & prediction & error & error fraction & prediction & error & error fraction\\ 
 \hline\hline
1&840,000&-9,000&0.011&768,930&-80,070&0.094\\
2&640,000&-10,000&0.015&632,010&-17,990&0.028\\
3&890,000&241,000&0.371&893,500&244,500&0.377\\
4&940,000&115,000&0.139&850,600&25,600&0.031\\
5&640,000&-184,000&0.223&614,880&-209,120&0.254\\
6&630,000&175,000&0.385&550,080&95,080&0.209\\
7&2,430,000&-470,000&0.162&2,417,800&-482,200&0.166\\
8&1,240,000&90,000&0.078&1,195,600&45,600&0.04\\
9&630,000&-165,000&0.208&552,790&-242,210&0.305\\
10&1,740,000&-260,000&0.13&1,701,100&-298,900&0.149\\
\hline
\hline
\end{tabular}
\caption{Prediction, prediction error (i.e., $m-a$), and prediction error fraction (i.e., $(m-a)/a$) of the two- and eight-feature models on the apartments used in the training phase of our experiments. }
\label{tab:model_errors_train}
\end{center}
\end{table*}

\begin{table*}[t!]
\begin{center}
\footnotesize
 \begin{tabular}{c c c c c c c c c}
 \toprule
 \pbox{10cm}{Apartment \\ ID} & \multicolumn{8}{c}{Apartment configurations} \\
 \hline
 & Bedrooms & Bathrooms & \pbox{2cm} {Square \\ footage} & \pbox{2cm}{Total \\ rooms} & \pbox{2cm}{Days on \\the market} & \pbox{2cm}{Maintenance \\ fee} &  \pbox{2cm}{Distance from \\ the subway \\ (miles)} & \pbox{2cm}{Distance from \\ a school \\ (miles)}\\ 
 \hline\hline
1&1&1&925&3&80&954&0.173&0.312 \\
\hline
2&2&1&1080&5&39&846&0.207&0.212 \\
\hline
3&3&2&1530&5&15&1550&0.226&0.251 \\
\hline
4&2&2&1140&4.5&93&863&0.122&0.278 \\
\hline
5&1&1&540&3&11&437&0.202&0.199 \\
\hline
6&0&1&540&2.5&74&341&0.122&0.278 \\
\hline
7&2&1&1240&4.5&32&1370&0.081&0.262 \\
\hline
8&2&2&1240&4.5&14&906&0.178&0.225 \\
\hline
9&2&1&1250&5&23&1480&0.089&0.281 \\
\hline
10&1&1&532&2.5&20&388&0.122&0.278 \\
\hline
11&1&2&750&3&225&825&0.159&0.144 \\
\hline
12&1&3&726&4&17&444&0.121&0.101 \\
\hline
13&1&1&788&3.5&51&473&0.122&0.278 \\
\hline
14&1&3&350&4&13&430&0.221&0.131 \\
\hline
\hline
\end{tabular}
\caption{Configuration of the apartments used in our experiments during the testing phase. Apartments 1--12 were used in experiments 1, 2, and 3. In Experiment 4, apartments 1, 6, 8, 9, 10, 12 (``Apartment 6'' in Experiment 4), 13 (``Apartment 7'' in Experiment 4), and 14 (``Apartment 8'' in Experiment 4) were used. Apartments 12 and 14 were synthetically generated.}
\label{tab:apt_configs_testing}
\end{center}
\end{table*}

\begin{table*}[t!]
\begin{center}
 \begin{tabular}{c c c c | c c c}
 \toprule
 Apartment \\ID & \multicolumn{3}{c}{two-feature model} & \multicolumn{3}{c}{eight-feature model} \\
 \hline
 & prediction & error & error fraction & prediction & error & error fraction\\ 
 \hline\hline
1& 1,015,000&90,000&0.097&957,560&32,560&0.035 \\
2& 1,170,000&-80,000&0.064&1,166,040&-83,960&0.067 \\
3& 1,970,000&-560,000&0.221&1,989,200&-540,800&0.214 \\
4& 1,580,000&-170,000&0.097&1,573,970&-176,030&0.100 \\
5& 630,000&74,000&0.133&634,830&78,830&0.142 \\
6& 630,000&-145,000&0.187&555,190&-219,810&0.284 \\
7& 1,330,000&331,000&0.331&1,274,700&275,700&0.276 \\
8& 1,680,000&-20,000&0.012&1,685,340&-14,660&0.009 \\
9& 1,340,000&-210,000&0.135&1,264,600&-285,400&0.184 \\
10& 622,000&97,000&0.185&642,820&117,820&0.224 \\
11& 1,190,000&541,000&0.834&1,099,550&450,550&0.694 \\
12& 1,516,000& \NA & \NA &1,475,960& \NA & \NA \\
13& 878,000&-292,000&0.25&858,270&-311,730&0.266 \\
14& 1,140,000& \NA & \NA &1,115,300& \NA & \NA \\
\hline
\hline
\end{tabular}
\caption{Prediction, prediction error (i.e., $m-a$), and prediction error fraction (i.e., $(m-a)/a$) of the two- and eight-feature models on the apartments used in the testing phase of our experiments. }
\label{tab:model_errors_test}
\end{center}
\end{table*}
\clearpage

 \section{Experiment 3 Hypotheses and Findings}
  \label{appndx:experiment_3}
We pre-registered four hypotheses:\edit{\footnote{Pre-registered hypotheses this experiment are available at \url{https://aspredicted.org/795du.pdf
}.}}
\begin{itemize}[nosep]
\item[H7.] \textbf{Deviation.} Participants' predictions will deviate less from the predictions of a clear model with a small number of features than the predictions of a black-box model with a large number of features.
\item[H8.] \textbf{Weight of advice.} Weight of advice will be higher for participants who see a clear model with a small number of features than for those who see a black-box model with a large number of features.
\item[H9.] \textbf{Humans vs. machines.} Participants' deviation and weight of advice measures will differ depending on whether the predictions come from a black-box model with a large number of features or a human expert.\looseness=-1
\item[H10.] \textbf{Detection of mistakes.} Participants in different conditions will exhibit varying abilities to correct the model's inaccurate predictions on unusual examples.
\end{itemize}

The first two hypotheses are variations on H2 from our first experiment, while the last hypothesis is identical to H3.
\subsection{Results}
\begin{figure*}[t]
  \captionsetup[subfigure]{aboveskip=-2pt,belowskip=-2pt}
  \centering
  \textbf{\fontfamily{phv}\selectfont Experiment 3: Weight of advice}\par\medskip

        \begin{subfigure}[b]{0.45\textwidth}
                \includegraphics[width=\linewidth]{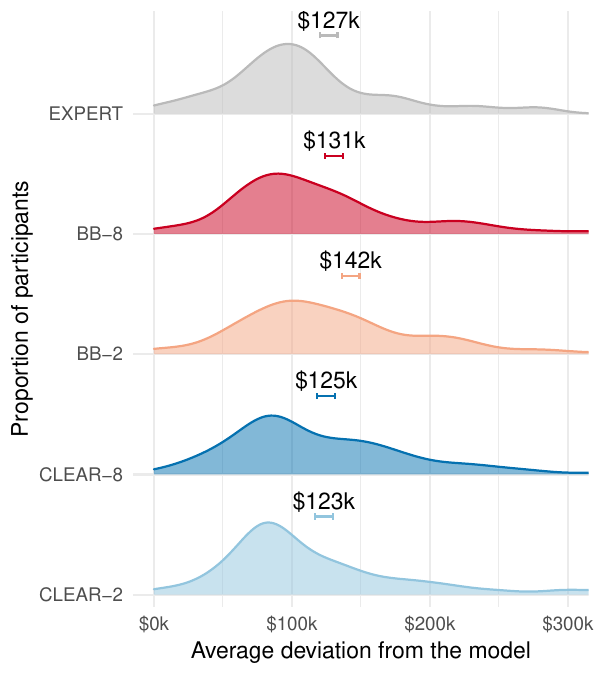}
               \caption{}
                \label{fig:exp3_dev_from_model}
                \Description[]{Distributions of deviations in the four main conditions and a condition labelled ``Expert''. Distributions are generally similar in shape and their means range from \$123,000 (clear-2) to \$142,000 (black-box-2). The expert condition has a mean of \$127,000}
        \end{subfigure} %
        \begin{subfigure}[b]{0.45\textwidth}
                \includegraphics[width=\linewidth]{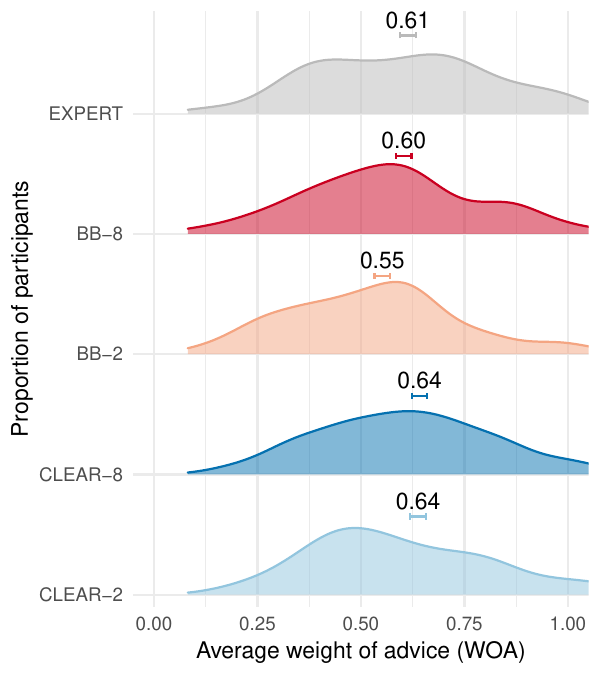}
                \caption{}
                \label{fig:exp3_woa}
                \Description[]{Distributions of weight of advice measures in the four main conditions and a condition labelled ``expert''. Distributions are similar in shape and their means range from .55 (black-box-2) to .64 (clear-2 and clear-8). The expert condition has a mean of .61. The black-box-2 and black-box-8 conditions have thinner right tails.}
        \end{subfigure}
\caption{Results from Experiment 3: density plots for (a) mean deviation of participants' predictions from the model's prediction and (b) mean weight of advice. Numbers in each subplot indicate average values over all participants in the corresponding condition and error bars indicate one standard error.}
\label{fig:exp3_main_results}
\end{figure*}

H7. \textbf{Deviation.} In line with the findings from the first two experiments, there was no significant difference in participants' deviation from the model between \abr{clear-2} and \abr{bb-8} ($t\left(798\right) = -0.87$ , $p = 0.384$, see Figure~\ref{fig:exp3_dev_from_model}).

H8. \textbf{Weight of advice.} Weight of advice is not well defined when a participant's initial prediction matches the model's prediction (i.e., $u_1 = m$). For each condition, we therefore calculated the mean weight of advice over all participant--apartment pairs for which the participant's initial prediction did not match the model's prediction, which can be viewed as calculating the mean conditional on there being a difference between the participant's and the model's predictions.  Between conditions, we found no significant difference in the fraction of times that participants' initial predictions matched the model's predictions. In line with the findings for deviation in the first two experiments, there was no significant difference in participants' weight of advice between the \abr{clear-2} and \abr{bb-8} conditions ($t\left(819\right) = 1.27$, $p = 0.205$, see Figure~\ref{fig:exp3_woa}).

H9. \textbf{Humans vs. machines.} The hypothesis that people would deviate less from machine predictions was not supported as there was not a significant difference in participants' deviation from the model ($t\left(994\right) = 0.45$ , $p = 0.655$) or in their weight of advice ($t\left(1005\right) = -0.38$ , $p = 0.704$) between the \abr{bb-8} and \abr{expert} conditions. We expect that the difference between our results and those in \cite{L17,logg2019algorithm} is due to participants getting more experience with the model (or expert) and its predictions over the course of twelve apartments in our experiment.

H10. \textbf{Detection of mistakes.} Participants in the clear conditions were no less able to correct inaccurate predictions ($t\left(798\right) = -0.96$, $p = 0.337$ and $t\left(798\right) = -0.19$, $p = 0.847$ for the contrast of \abr{clear-2} and \abr{clear-8} with \abr{bb-2} and \abr{bb-8} for apartments 11 and 12, respectively). We investigate this further in Experiment 4 (Section~\ref{sec:exp4}).\looseness=-1

  \section{Full Distributions of Participants' Predictions}
  \label{appndx:distributions}
\begin{figure*}[h]
  \centering
  \textbf{\fontfamily{phv}\selectfont Experiment 1: New York City prices (training phase)}\par\medskip
                \includegraphics[width=\linewidth]{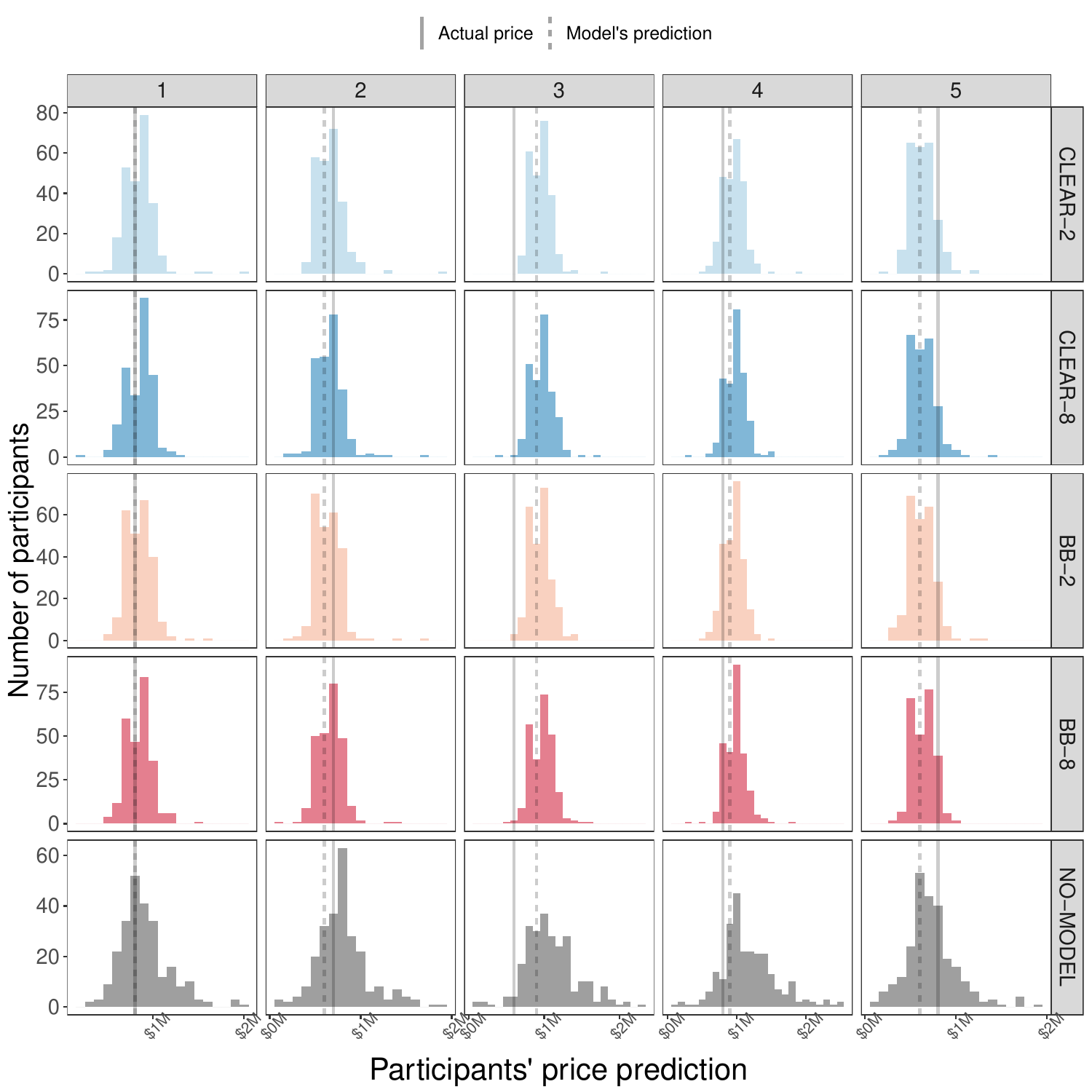}
				\label{fig:exp1_dist_training_final_prediction_1}
                \Description[]{Matrix of histograms of predicted prices in which the columns are apartment numbers and the rows are the names of the four main conditions and the no-model condition. Solid and dashed lines in each plot indicate the actual selling price and model's prediction. The four model condition histograms are rather similar within apartment. The no-model distributions are higher in variance.}
                \caption{Distribution of participants' predictions of prices of apartments 1--5 in the training phase in Experiment 1.}
 \end{figure*}
\clearpage
\begin{figure*}[t!]
  \centering
  \textbf{\fontfamily{phv}\selectfont Experiment 1: New York City prices (training phase)}\par\medskip
                \includegraphics[width=\linewidth]{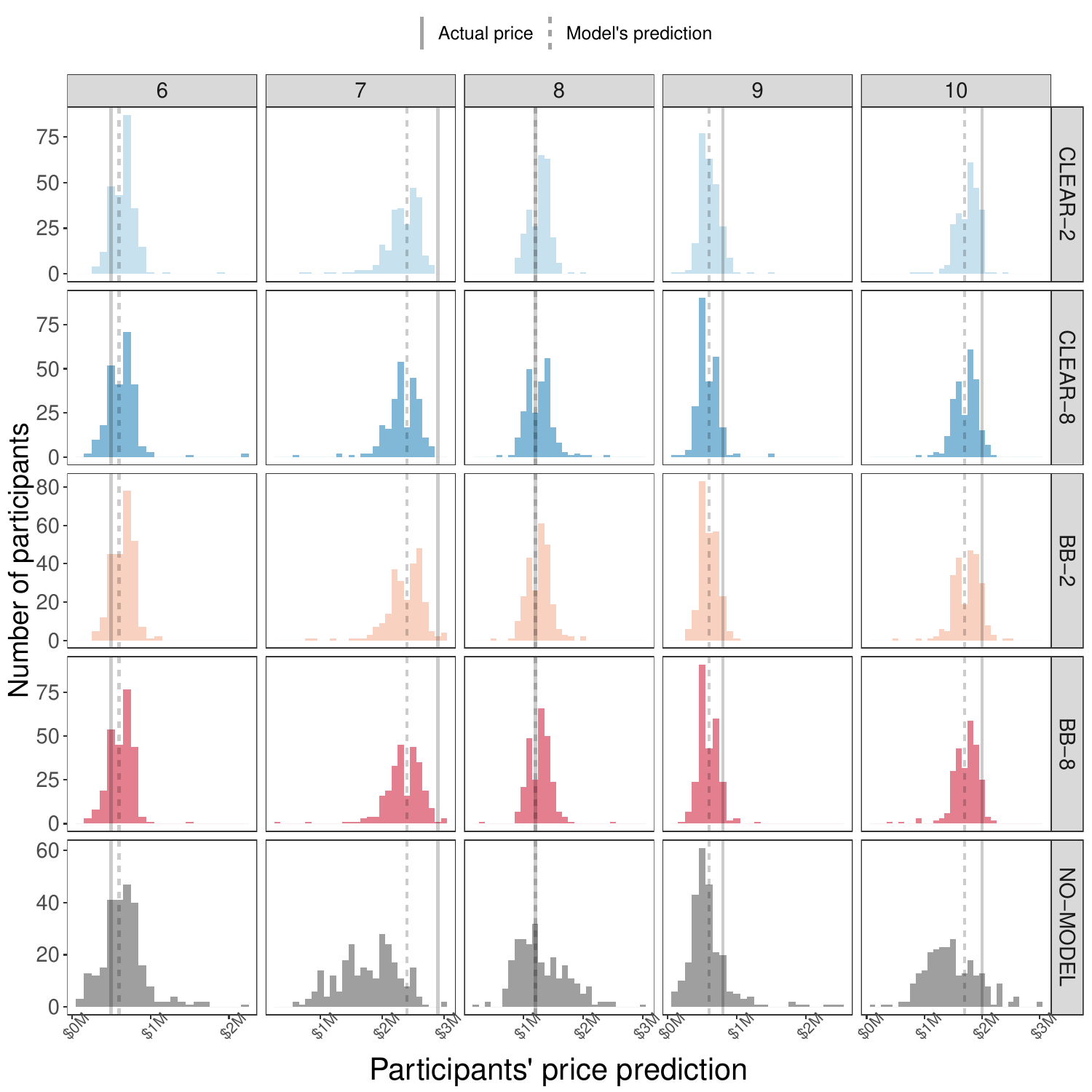}
                \label{fig:exp1_dist_training_final_prediction_2}
                \Description[]{Matrix of histograms of predicted prices in which the columns are apartment numbers and the rows are the names of the four main conditions and the no-model condition. Solid and dashed lines in each plot indicate the actual selling price and model's prediction. The four model condition histograms are rather similar within apartment. The no-model distributions are higher in variance.}
                \caption{Distribution of participants' predictions of prices of apartments 6--10 in the training phase in Experiment 1.}
\end{figure*}
\clearpage
\begin{figure*}[t!]
  \centering
  \textbf{\fontfamily{phv}\selectfont Experiment 1: New York City prices (testing phase)}\par\medskip
                \includegraphics[width=\linewidth]{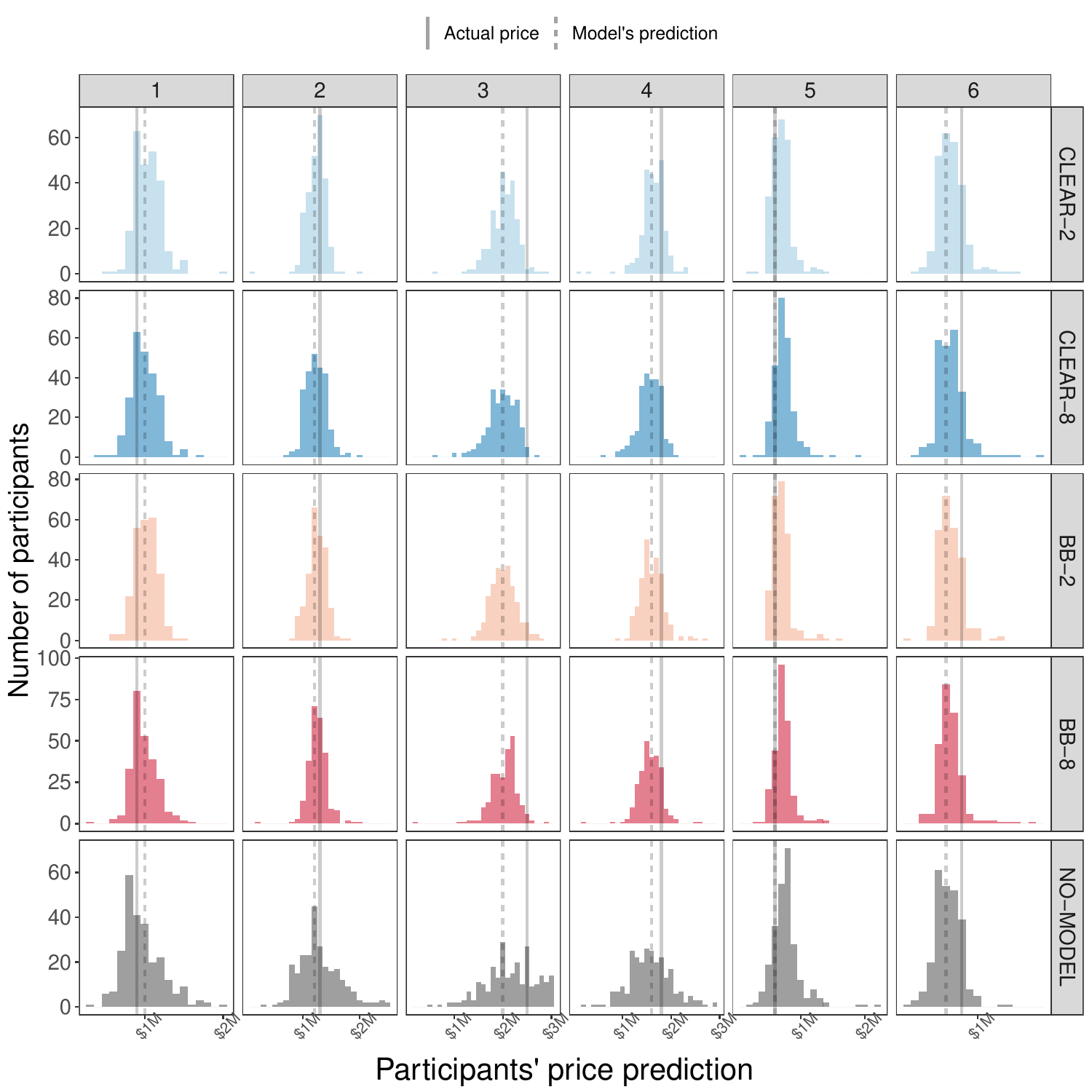}
				\label{fig:exp1_final_pred_dist_1}
                \Description[]{Matrix of histograms of predicted prices in which the columns are apartment numbers and the rows are the names of the four main conditions and the no-model condition. Solid and dashed lines in each plot indicate the actual selling price and model's prediction. The four model condition histograms are rather similar within apartment. The no-model distributions are higher in variance.}
                \caption{Distribution of participants' predictions of prices of apartments 1--6 in the testing phase in Experiment 1.}
 \end{figure*}
\clearpage

\begin{figure*}[t!]
  \centering
  \textbf{\fontfamily{phv}\selectfont Experiment 1: New York City prices (testing phase)}\par\medskip
                \includegraphics[width=\linewidth]{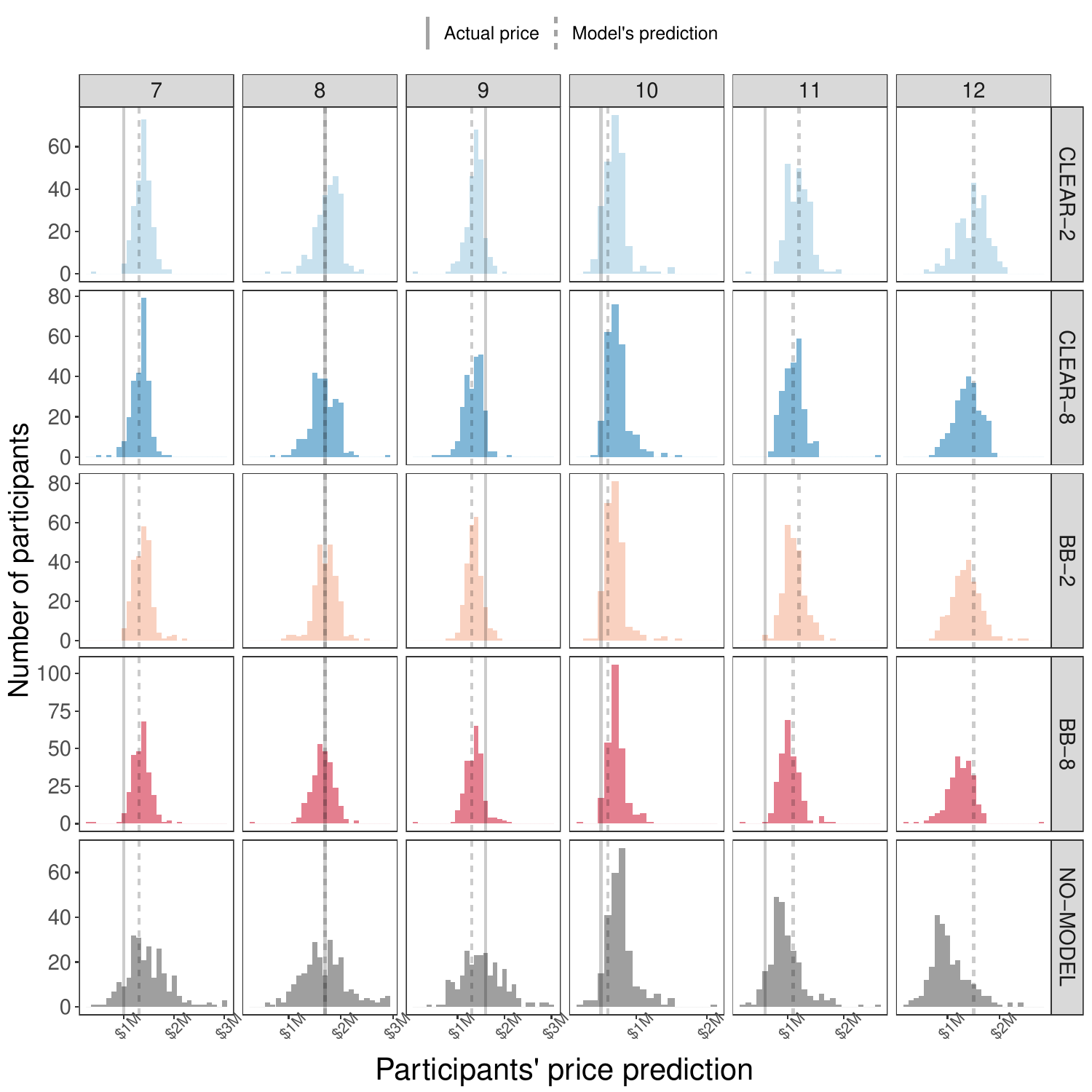}
                \label{fig:exp1_final_pred_dist_2}
                \Description[]{Matrix of histograms of predicted prices in which the columns are apartment numbers and the rows are the names of the four main conditions and the no-model condition. Solid and dashed lines in each plot indicate the actual selling price and model's prediction. The four model condition histograms are rather similar within apartment. The no-model distributions are higher in variance.}
        \caption{Distribution of participants' predictions of prices of apartments 7--12 in the testing phase in Experiment 1.}
\end{figure*}
\clearpage
\begin{figure*}[t!]
  \centering
  \textbf{\fontfamily{phv}\selectfont Experiment 2: Representative U.S. prices (training phase)}\par\medskip
                \includegraphics[width=\linewidth]{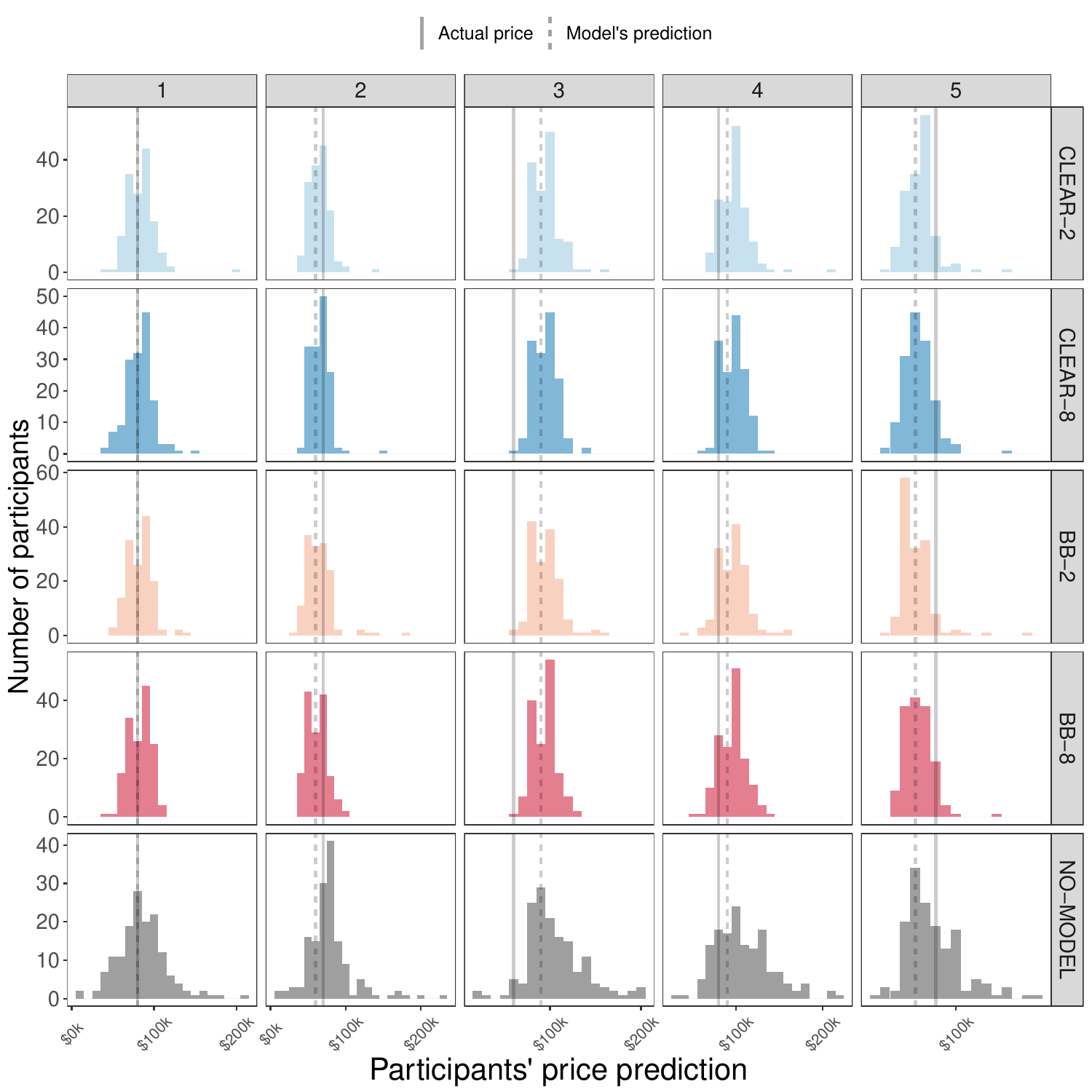}
				\label{fig:exp2_dist_training_final_prediction_1}
                \Description[]{Matrix of histograms of predicted prices in which the columns are apartment numbers and the rows are the names of the four main conditions and the no-model condition. Solid and dashed lines in each plot indicate the actual selling price and model's prediction. The four model condition histograms are rather similar within apartment. The no-model distributions are higher in variance.}
                \caption{Distribution of participants' predictions of prices of apartments 1--5 in the training phase in Experiment 2.}
\end{figure*}
\clearpage

\begin{figure*}[t!]
  \centering
  \textbf{\fontfamily{phv}\selectfont Experiment 2: Representative U.S. prices (training phase)}\par\medskip
                \includegraphics[width=\linewidth]{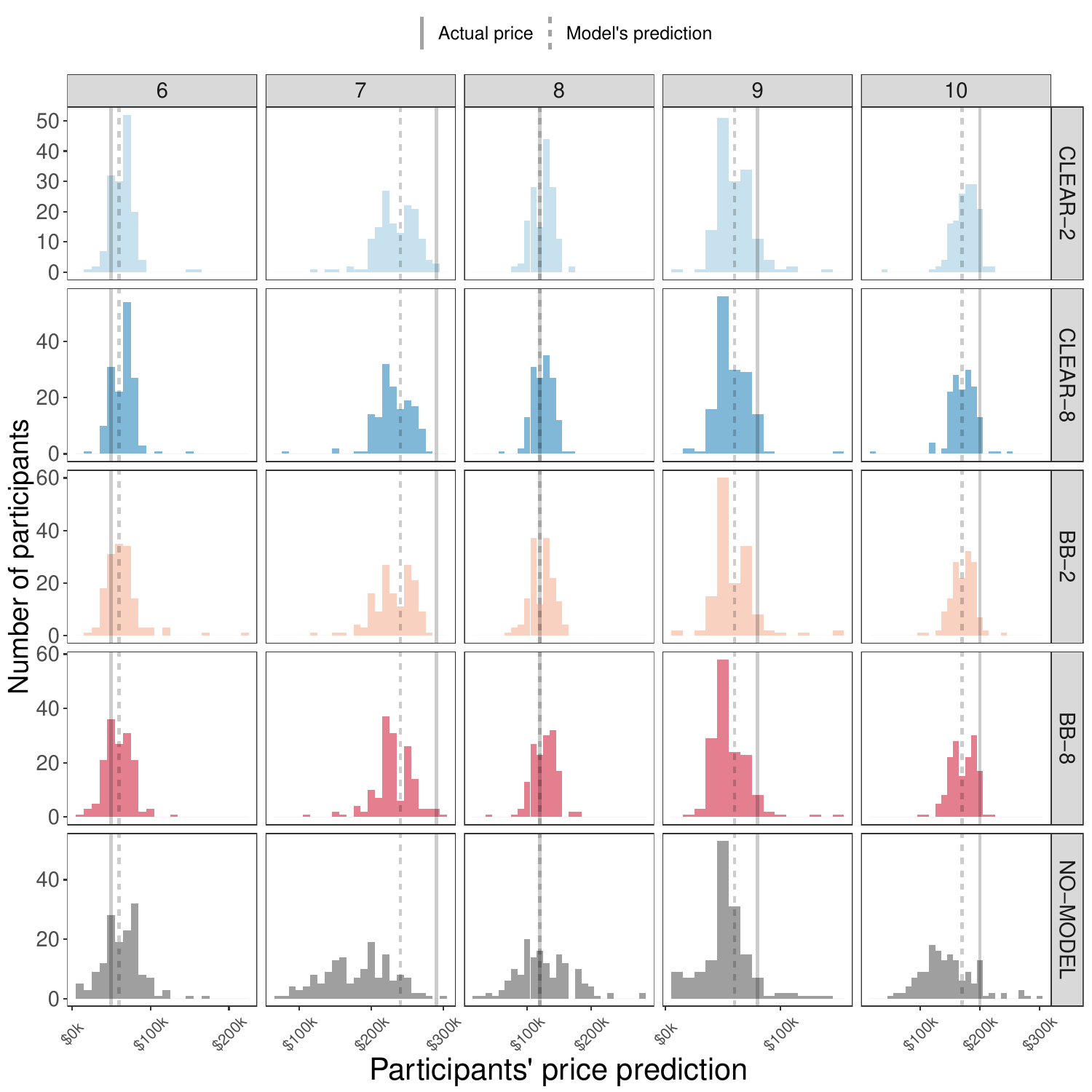}
                \label{fig:exp2_dist_training_final_prediction_2}
                 \Description[]{Matrix of histograms of predicted prices in which the columns are apartment numbers and the rows are the names of the four main conditions and the no-model condition. Solid and dashed lines in each plot indicate the actual selling price and model's prediction. The four model condition histograms are rather similar within apartment. The no-model distributions are higher in variance.}
                \caption{Distribution of participants' predictions of prices of apartments 6--10 in the training phase in Experiment 2.}
\end{figure*}
\clearpage

\begin{figure*}[t!]
  \centering
  \textbf{\fontfamily{phv}\selectfont Experiment 2: Representative U.S. prices (testing phase)}\par\medskip
                \includegraphics[width=\linewidth]{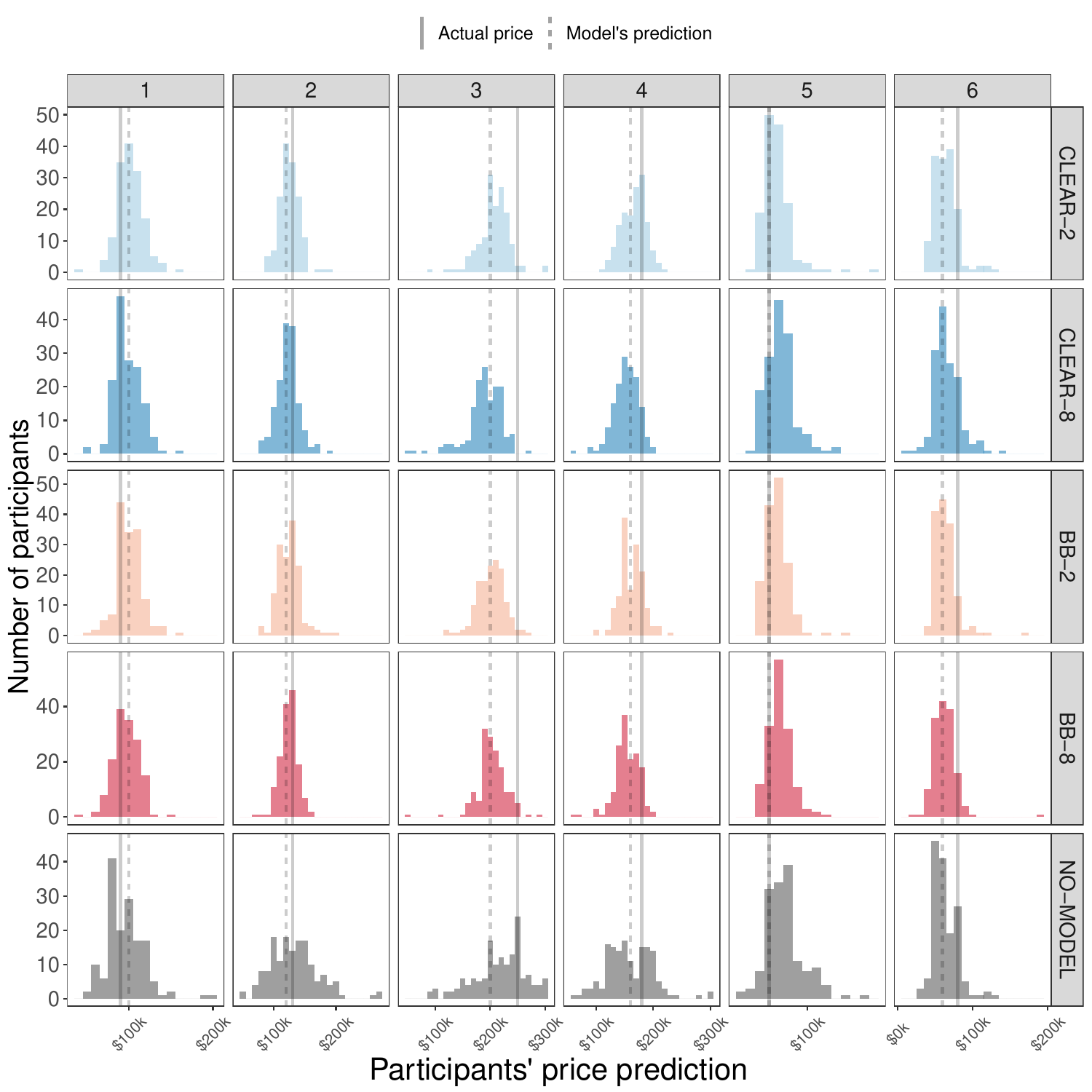}
                \label{fig:exp2_final_pred_dist_1}
                 \Description[]{Matrix of histograms of predicted prices in which the columns are apartment numbers and the rows are the names of the four main conditions and the no-model condition. Solid and dashed lines in each plot indicate the actual selling price and model's prediction. The four model condition histograms are rather similar within apartment. The no-model distributions are higher in variance.}
                \caption{Distribution of participants' predictions of prices of apartments 1--6 in the testing phase in Experiment 2.}
                
\end{figure*}
\clearpage

\begin{figure*}[t!]
  \centering
  \textbf{\fontfamily{phv}\selectfont Experiment 2: Representative U.S. prices (testing phase)}\par\medskip 
                \includegraphics[width=\linewidth]{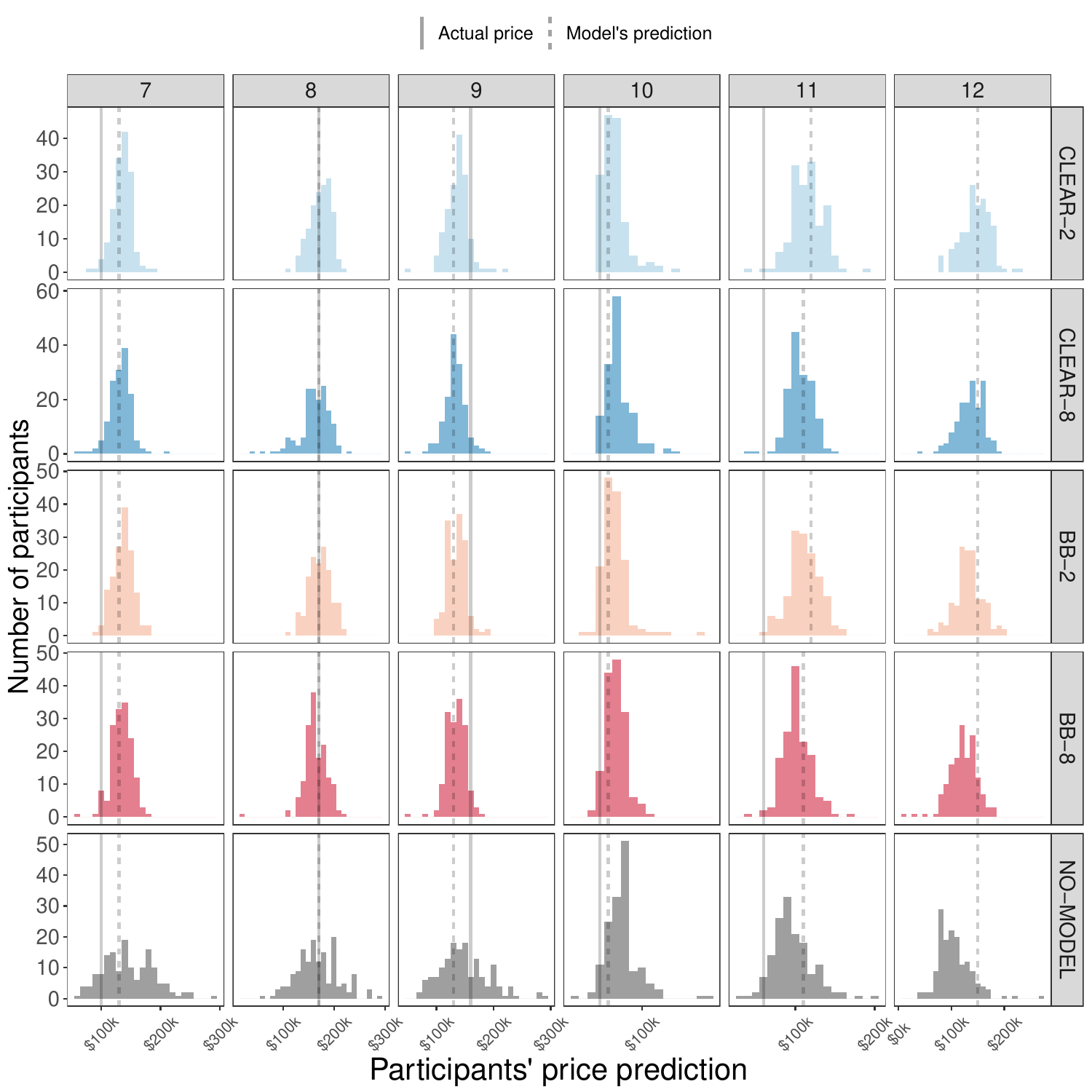}
                \label{fig:exp2_final_pred_dist_2}
                 \Description[]{Matrix of histograms of predicted prices in which the columns are apartment numbers and the rows are the names of the four main conditions and the no-model condition. Solid and dashed lines in each plot indicate the actual selling price and model's prediction. The four model condition histograms are rather similar within apartment. The no-model distributions are higher in variance.}
        \caption{Distribution of participants' predictions of prices of apartments 7--12 in the testing phase in Experiment 2.}
\end{figure*}
\clearpage
\begin{figure*}[t!]
  \centering
  \textbf{\fontfamily{phv}\selectfont Experiment 3: Weight of advice (testing phase)}\par\medskip
                \includegraphics[width=\linewidth]{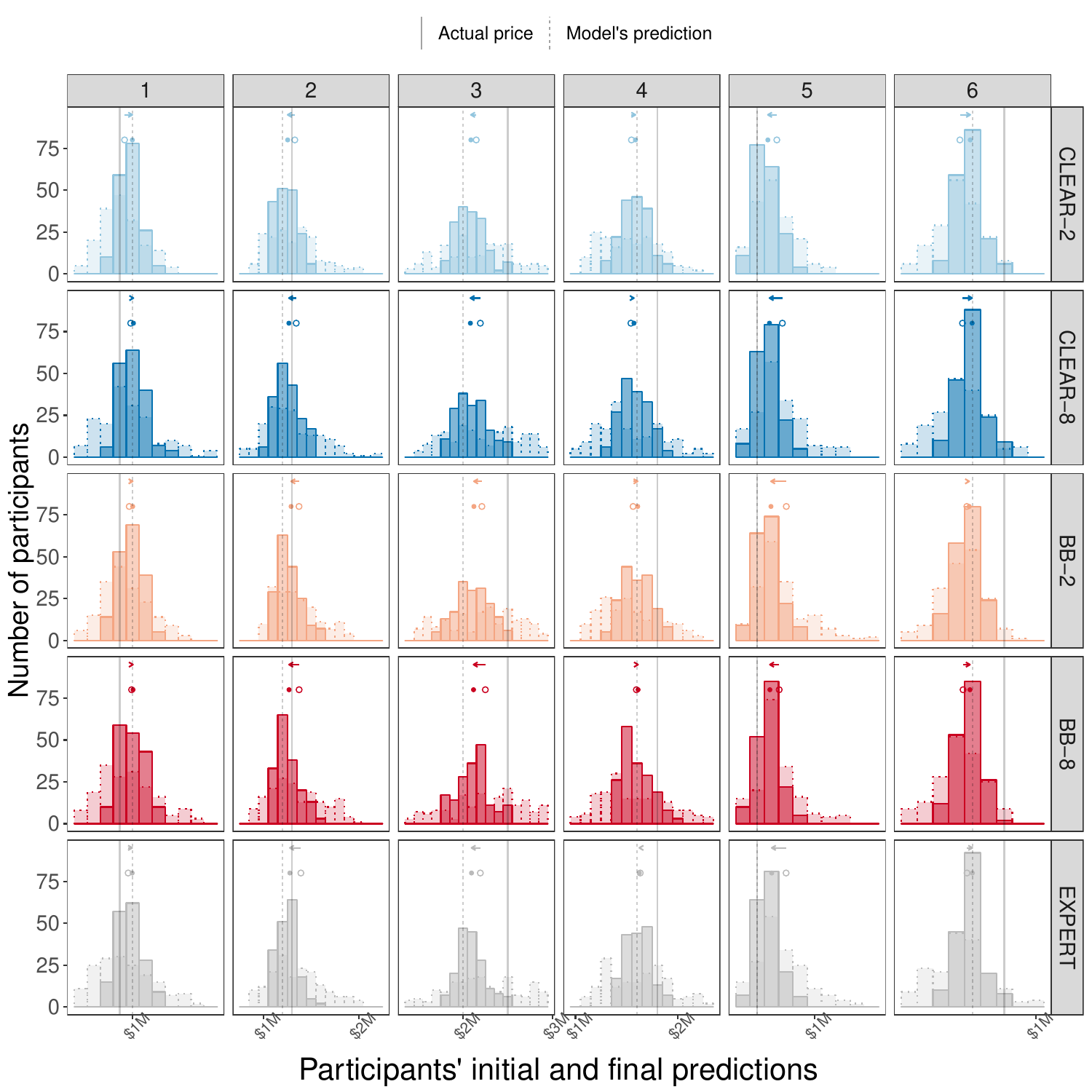}
        \caption{Distribution of participants' initial (before seeing the model's prediction, dotted) predictions and their final (after seeing the model's prediction, solid) prediction of prices of apartments 1--6 in Experiment 3. Points show the mean initial and final predictions and the arrow indicates the shift in the mean predictions.}
                \label{fig:exp3_dist_init_final_1}
                \Description[]{Matrix of histograms of price predictions in which the columns are apartment numbers and the rows are the names of the four main conditions and the ``Expert'' condition. Each plot in the matrix contains two overlapping histograms, based on initial and final predictions. Dashed and solid lines in each plot indicate the actual prices and model's predictions. All conditions' histograms are rather similar within apartment and time (initial and final). The final distributions are lower in variance.}
\end{figure*}
\clearpage

\begin{figure*}[t!]
  \centering
  \textbf{\fontfamily{phv}\selectfont Experiment 3: Weight of advice (testing phase)}\par\medskip
                \includegraphics[width=\linewidth]{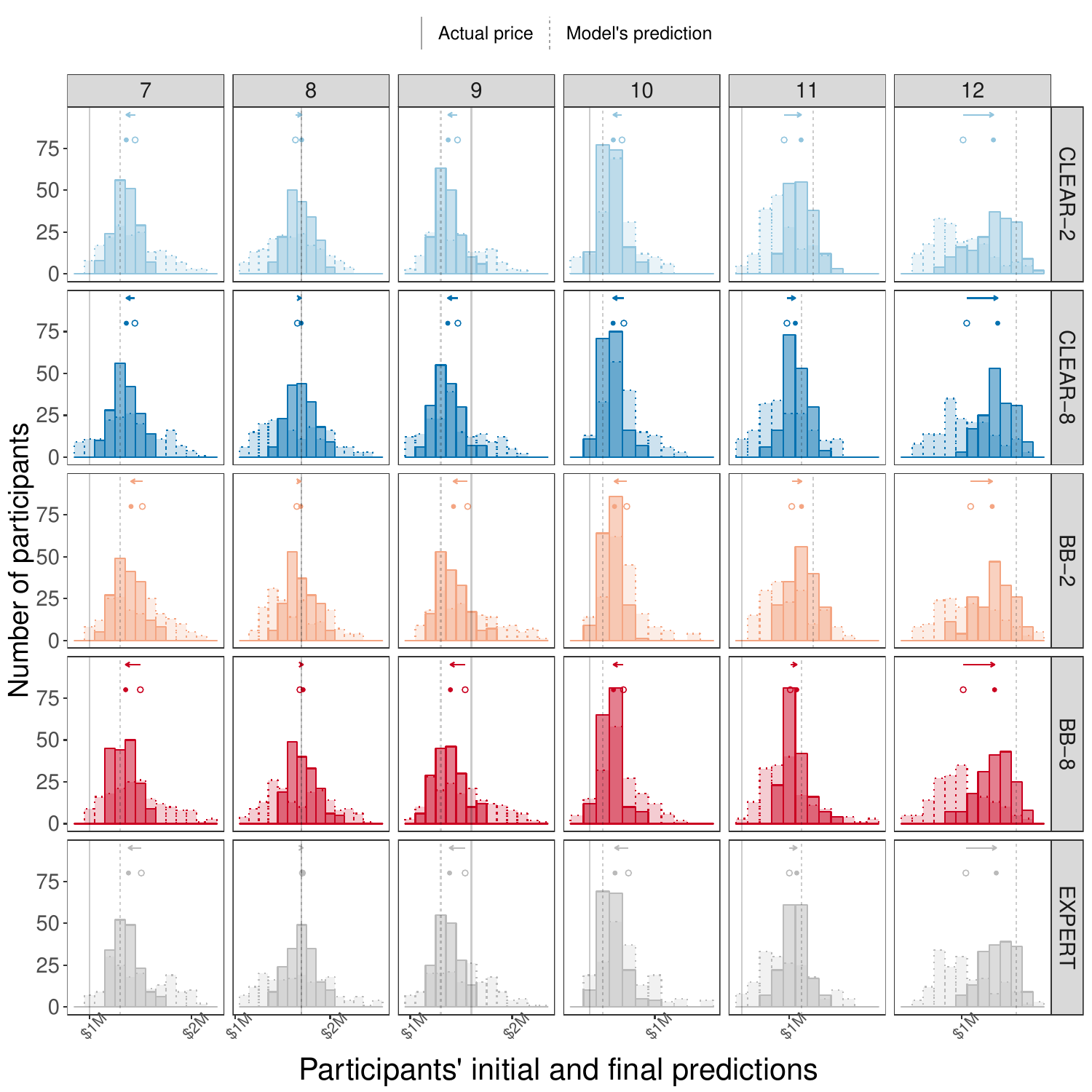}
        \caption{Distribution of participants' initial (before seeing the model's prediction, dotted) predictions and their final (after seeing the model's prediction, solid) prediction of prices of apartments 7--12 in Experiment 3. Points show the mean initial and final predictions and the arrow indicates the shift in the mean predictions.}
                \label{fig:exp3_dist_init_final_2}
                \Description[]{Matrix of histograms of price predictions in which the columns are apartment numbers and the rows are the names of the four main conditions and the ``expert'' condition. Each plot in the matrix contains two overlapping histograms, based on initial and final predictions. Dashed and solid lines in each plot indicate the actual prices and model's predictions. All conditions' histograms are rather similar within apartment and time (initial and final). The final distributions are lower in variance.}
\end{figure*}
\clearpage
\begin{figure*}[t!]
  \centering
  \textbf{\fontfamily{phv}\selectfont Experiment 4: Outlier focus (training phase)}\par\medskip
                \includegraphics[width=\linewidth]{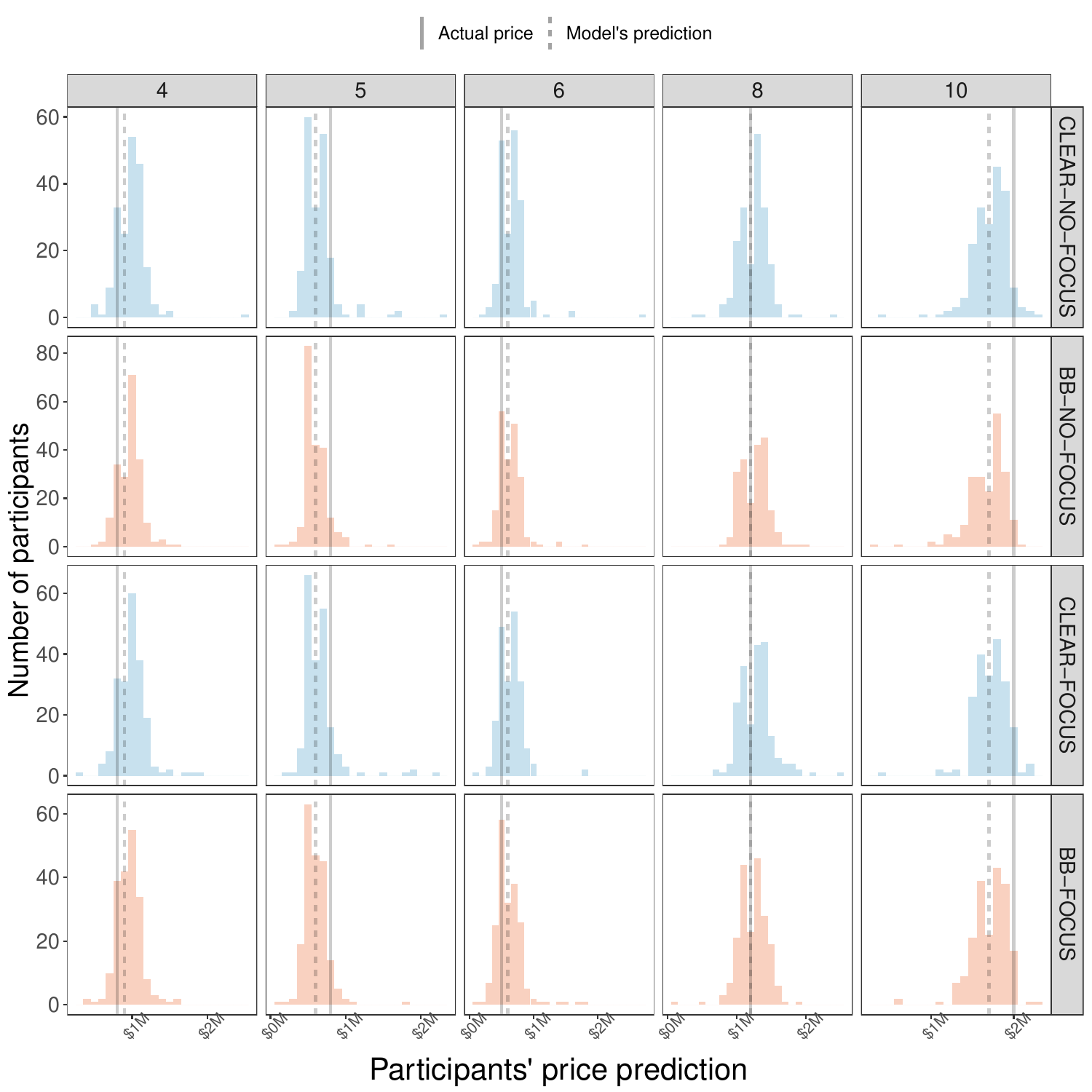}
                \label{fig:exp4_dist_training_final_prediction}
                \Description[]{Matrix of histograms of price predictions in which the columns are apartment numbers and the rows are the names of the four main experimental conditions (clear-no-focus, black-box-no-focus, clear-focus, black-box-focus). Dashed and solid lines in each plot indicate the model's prediction and actual selling price. All conditions' histograms are rather similar within apartment.}
                \caption{Distribution of participants' predictions of prices of apartments in the training phase in Experiment 4.}
\end{figure*}
\clearpage

\begin{figure*}[t!]
  \centering
  \textbf{\fontfamily{phv}\selectfont Experiment 4: Outlier focus (testing phase)}\par\medskip
                \includegraphics[width=\linewidth]{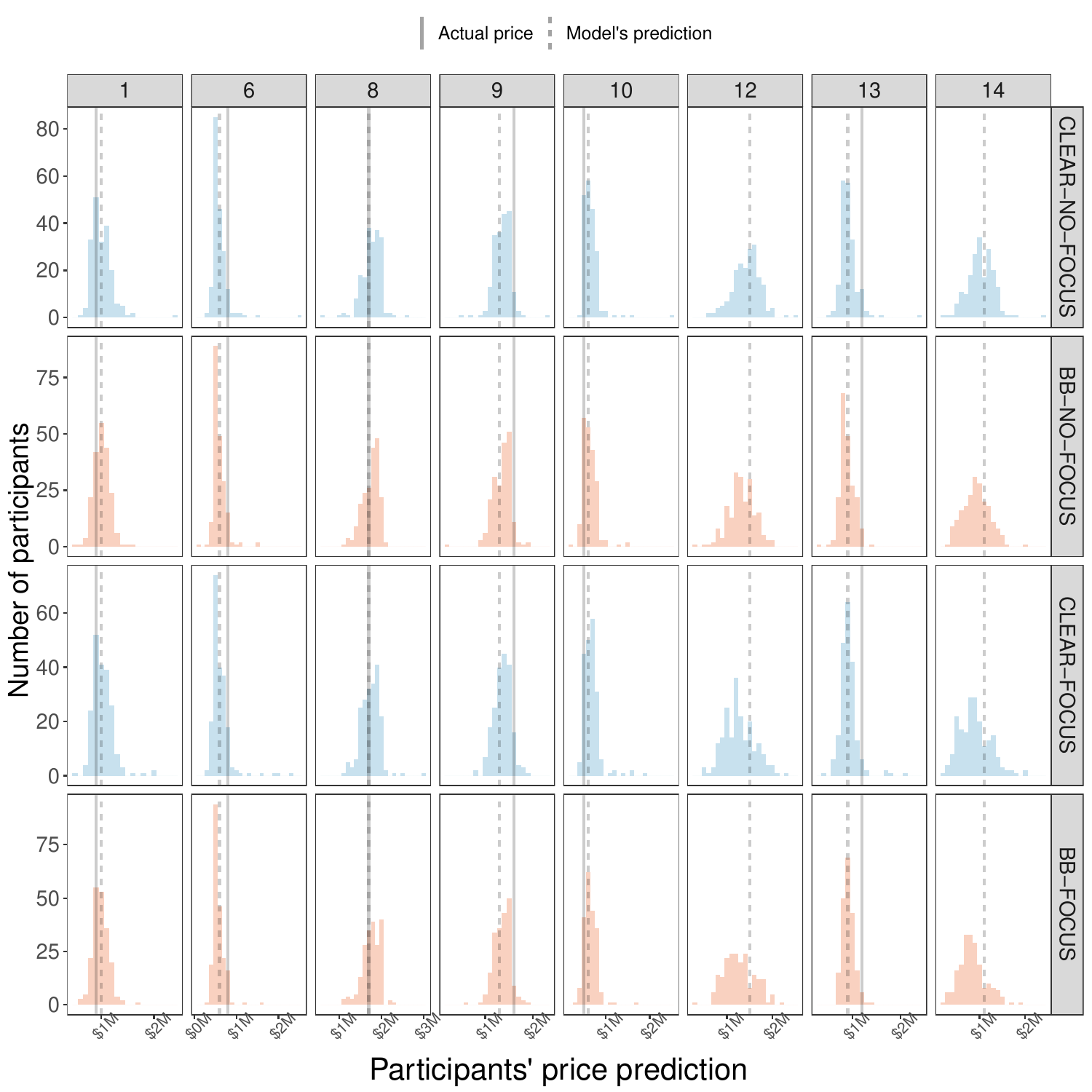}
                \label{fig:exp4_final_pred_dist}
                \Description[]{Matrix of histograms of price predictions in which the columns are apartment numbers and the rows are the names of the four main experimental conditions (clear-no-focus, black-box-no-focus, clear-focus, black-box-focus). Dashed and solid lines in each plot indicate the model's prediction and actual selling price. All conditions' histograms are rather similar within apartment though the distributions for the two unusual apartments differ in their central tendency by condition, as described in the text.}
                \caption{Distribution of participants' predictions of prices of apartments in the testing phase in Experiment 4.}
                
\end{figure*}
\clearpage

  \section{ANOVA Tables}
  \label{appndx:anova_tables}
  \subsection{Experiment 1: Predicting Prices}

\begin{table}[H]
\centering
\begin{tabular}{lrrrrrr}
  \hline
 & Sum Sq & Mean Sq & NumDF & DenDF & F value & Pr($>$F) \\ 
  \hline
transparency & 0.52 & 0.52 & 1.00 & 994.00 & 12.57 & 0.0004 \\ 
  num\_features & 4.90 & 4.90 & 1.00 & 994.00 & 119.54 & 0.0000 \\ 
  transparency:num\_features & 1.70 & 1.70 & 1.00 & 994.00 & 41.48 & 0.0000 \\ 
   \hline
\end{tabular}
\caption{Results from two-way ANOVA on the simulation error in Experiment 1.}
\label{tab:anova_exp1_simerr}
\end{table}

\begin{table}[H]
\centering
\begin{tabular}{lrrrrrr}
  \hline
 & Sum Sq & Mean Sq & NumDF & DenDF & F value & Pr($>$F) \\ 
  \hline
transparency & 0.10 & 0.10 & 1.00 & 994.00 & 5.83 & 0.0159 \\ 
  num\_features & 0.04 & 0.04 & 1.00 & 994.00 & 2.15 & 0.1427 \\ 
  transparency:num\_features & 0.00 & 0.00 & 1.00 & 994.00 & 0.06 & 0.8143 \\ 
   \hline
\end{tabular}
\caption{Results from two-way ANOVA on the deviation between the model's prediction and participants' prediction of the price in Experiment 1.}
\label{tab:anova_exp1_deviation}
\end{table}


\begin{table}[H]
\centering
\begin{tabular}{lrrrrr}
  \hline
 & Df & Sum Sq & Mean Sq & F value & Pr($>$F) \\ 
  \hline
transparency & 1 & 0.02 & 0.02 & 1.38 & 0.2405 \\ 
  num\_features & 1 & 0.03 & 0.03 & 1.70 & 0.1920 \\ 
  transparency:num\_features & 1 & 0.00 & 0.00 & 0.00 & 0.9509 \\ 
  Residuals & 994 & 17.01 & 0.02 &  &  \\ 
   \hline
   \end{tabular}
\caption{Results from two-way ANOVA on the deviation between the model's prediction and participants' prediction of the price for apartment 11 in Experiment 1.}
\label{tab:anova_exp1_q11}
\end{table}
\begin{table}[H]
\centering
\begin{tabular}{lrrrrr}
  \hline
 & Df & Sum Sq & Mean Sq & F value & Pr($>$F) \\ 
  \hline
transparency & 1 & 0.34 & 0.34 & 8.81 & 0.0031 \\ 
  num\_features & 1 & 0.04 & 0.04 & 1.13 & 0.2882 \\ 
  transparency:num\_features & 1 & 0.13 & 0.13 & 3.32 & 0.0687 \\ 
  Residuals & 994 & 38.07 & 0.04 &  &  \\ 
   \hline
\end{tabular}
\caption{Results from two-way ANOVA on the deviation between the model's prediction and participants' prediction of the price for apartment 12 in Experiment 1.}
\label{tab:anova_exp1_q12}
\end{table}

%

\subsection{Experiment 2: Scaled-down prices}

\begin{table}[H]
\centering
\begin{tabular}{lrrrrrr}
  \hline
 & Sum Sq & Mean Sq & NumDF & DenDF & F value & Pr($>$F) \\ 
  \hline
transparency & 0.003 & 0.003 & 1.00 & 594.00 & 7.54 & 0.0062 \\ 
  num\_features & 0.032 & 0.032 & 1.00 & 594.00 & 75.45 & 0.0000 \\ 
  transparency:num\_features & 0.018 & 0.018 & 1.00 & 594.00 & 43.14 & 0.0000 \\ 
   \hline
\end{tabular}
\caption{Results from two-way ANOVA on the simulation error in Experiment 2.}
\label{tab:anova_exp2_simerr}
\end{table}

\begin{table}[H]
\centering
\begin{tabular}{lrrrrrr}
  \hline
 & Sum Sq & Mean Sq & NumDF & DenDF & F value & Pr($>$F) \\ 
  \hline
transparency & 0.001 & 0.001 & 1.00 & 594.00 & 3.33 & 0.0685 \\ 
  num\_features & 0.000 & 0.000 & 1.00 & 594.00 & 1.29 & 0.2556 \\ 
  transparency:num\_features & 0.000 & 0.000 & 1.00 & 594.00 & 1.82 & 0.1775 \\ 
   \hline
\end{tabular}
\caption{Results from two-way ANOVA on the deviation between the model's prediction and participants' prediction of the price in Experiment 2.}
\label{tab:anova_exp2_deviation}
\end{table}

\begin{table}[H]
\centering
\begin{tabular}{lrrrrr}
  \hline
 & Df & Sum Sq & Mean Sq & F value & Pr($>$F) \\ 
  \hline
transparency & 1.0 & 0.001 & 0.001 & 3.29 & 0.0702 \\ 
  num\_features & 1.0 & 0.001 & 0.001 & 4.51 & 0.0340 \\ 
  transparency:num\_features & 1.0 & 0.000 & 0.000 & 1.20 & 0.2731 \\ 
  Residuals & 594.0 & 0.092 & 0.000 &  &  \\ 
   \hline
   \end{tabular}
\caption{Results from two-way ANOVA on the deviation between the model's prediction and participants' prediction of the price for apartment 11 in Experiment 2.}
\label{tab:anova_exp2_q11}
\end{table}
\begin{table}[H]
\centering
\begin{tabular}{lrrrrr}
  \hline
 & Df & Sum Sq & Mean Sq & F value & Pr($>$F) \\ 
  \hline
transparency & 1.0 & 0.007 & 0.007 & 17.53 & 0.0000 \\ 
  num\_features & 1.0 & 0.002 & 0.002 & 4.05 & 0.0446 \\ 
  transparency:num\_features & 1.0 & 0.001 & 0.001 & 2.31 & 0.1291 \\ 
  Residuals & 594.0 & 0.229 & 0.000 &  &  \\ 
   \hline
\end{tabular}
\caption{Results from two-way ANOVA on the deviation between the model's prediction and participants' prediction of the price for apartment 12 in Experiment 2.}
\label{tab:anova_exp2_q12}
\end{table}

%
\subsection{Experiment 3: Weight of Advice}
\begin{table}[H]
\centering
\begin{tabular}{lrrrrrr}
  \hline
 & Sum Sq & Mean Sq & NumDF & DenDF & F value & Pr($>$F) \\ 
  \hline
transparency & 0.07 & 0.07 & 1.00 & 798.00 & 4.20 & 0.0409 \\ 
  num\_features & 0.01 & 0.01 & 1.00 & 798.00 & 0.66 & 0.4151 \\ 
  transparency:num\_features & 0.02 & 0.02 & 1.00 & 798.00 & 1.20 & 0.2731 \\ 
   \hline
\end{tabular}
\caption{Results from two-way ANOVA on the deviation between the model's prediction and participants' prediction of the price in the four primary conditions in Experiment 3.}
\label{tab:anova_exp3_deviation}
\end{table}

\begin{table}[H]
\centering
\begin{tabular}{lrrrrrr}
  \hline
 & Sum Sq & Mean Sq & NumDF & DenDF & F value & Pr($>$F) \\ 
  \hline
condition & 0.09 & 0.02 & 4.00 & 994.00 & 1.45 & 0.2147 \\ 
   \hline
\end{tabular}
\caption{Results from one-way ANOVA on the deviation between the model's prediction and participants' prediction of the price in all conditions (including the ``human expert'' condition) in Experiment 3.}
\label{tab:anova_exp3_deviation_bb8_vs_expert}
\end{table}

\begin{table}[H]
\centering
\begin{tabular}{lrrrrrr}
  \hline
 & Sum Sq & Mean Sq & NumDF & DenDF & F value & Pr($>$F) \\ 
  \hline
transparency & 2.14 & 2.14 & 1.00 & 817.77 & 10.47 & 0.0013 \\ 
  num\_features & 0.42 & 0.42 & 1.00 & 817.77 & 2.07 & 0.1509 \\ 
  transparency:num\_features & 0.32 & 0.32 & 1.00 & 817.77 & 1.57 & 0.2109 \\ 
   \hline
\end{tabular}
\caption{Results from two-way ANOVA on the weight of advice in the four primary conditions in Experiment 3.}
\label{tab:anova_exp3_woa}
\end{table}

\begin{table}[H]
\centering
\begin{tabular}{lrrrrrr}
  \hline
 & Sum Sq & Mean Sq & NumDF & DenDF & F value & Pr($>$F) \\ 
  \hline
condition & 2.92 & 0.73 & 4.00 & 1013.65 & 3.77 & 0.0048 \\ 
   \hline
\end{tabular}
\caption{Results from one-way ANOVA on weight of advice in all conditions (including the ``human expert'' condition) in Experiment 3.}
\label{tab:anova_exp3_deviation_bb8_vs_expert}
\end{table}

\begin{table}[H]
\centering
\begin{tabular}{lrrrrr}
  \hline
 & Df & Sum Sq & Mean Sq & F value & Pr($>$F) \\ 
  \hline
transparency & 1 & 0.01 & 0.01 & 0.92 & 0.3380 \\ 
  num\_features & 1 & 0.15 & 0.15 & 10.87 & 0.0010 \\ 
  transparency:num\_features & 1 & 0.03 & 0.03 & 2.08 & 0.1497 \\ 
  Residuals & 798 & 10.86 & 0.01 &  &  \\ 
   \hline
\end{tabular}
\caption{Results from two-way ANOVA on the deviation between the model's prediction and participants' prediction of the price for apartment 11 in Experiment 3.}
\label{tab:anova_exp3_q11}
\end{table}

\begin{table}[H]
\centering
\begin{tabular}{lrrrrr}
  \hline
 & Df & Sum Sq & Mean Sq & F value & Pr($>$F) \\ 
  \hline
transparency & 1 & 0.00 & 0.00 & 0.04 & 0.8439 \\ 
  num\_features & 1 & 0.38 & 0.38 & 10.26 & 0.0014 \\ 
  transparency:num\_features & 1 & 0.01 & 0.01 & 0.35 & 0.5516 \\ 
  Residuals & 798 & 29.32 & 0.04 &  &  \\ 
   \hline
\end{tabular}
\caption{Results from two-way ANOVA on the deviation between the model's prediction and participants' prediction of the price for apartment 12 in Experiment 3.}
\label{tab:anova_exp3_q12}
\end{table}
\clearpage

\end{appendices}

\end{document}